\documentclass[letterpaper]{article} %
\usepackage{aaai24}  %
\usepackage{times}  %
\usepackage{helvet}  %
\usepackage{courier}  %
\usepackage[hyphens]{url}  %
\usepackage{graphicx} %
\def\UrlFont{\rm}  %
\usepackage{natbib}  %
\usepackage{caption} %
\usepackage{algorithm}
\usepackage{algorithmic}
\usepackage{amsmath}
\usepackage{amssymb}
\usepackage{mathtools}
\usepackage{amsthm}

\usepackage{latexsym}
\usepackage{microtype}
\usepackage{times,latexsym}
\usepackage{url}
\usepackage{microtype}
\usepackage{graphicx}
\usepackage{multirow}
\usepackage{multicol}
\usepackage{tabularx}
\usepackage{booktabs}
\usepackage{subcaption}
\usepackage{kantlipsum}
\usepackage{hhline}
\usepackage{longtable}
\usepackage{tikz}
\usepackage{url}
\usepackage[toc,page,header]{appendix}
\usepackage{titletoc}
\usepackage{etoolbox}
\usepackage[inline]{enumitem}
\usepackage{rotating}
\usepackage{amsmath,amsthm,mathtools,amsfonts,amssymb}
\usepackage{array}
\usepackage{amsthm}
\usepackage{amssymb}
\usepackage{amsmath}
\usepackage{bbold}
\theoremstyle{plain}

\theoremstyle{definition}

\theoremstyle{remark}

\newcommand{\Drond}{\mathcal D}

\newcommand{\Prond}{\mathcal P}

\newcommand{\Rrond}{\mathcal R}

\newcommand{\Xrond}{\mathcal X}
\newcommand{\Yrond}{\mathcal Y}

\newcommand{\xbold}{\mathbf{x}}

\newcommand{\zbold}{\mathbf{z}}
\newcommand{\pbold}{\mathbf{p}}

\newcommand{\vertiii}[1]{{\left\vert\kern-0.25ex\left\vert\kern-0.25ex\left\vert #1 
    \right\vert\kern-0.25ex\right\vert\kern-0.25ex\right\vert}}

\newcommand{\norm}[1]{\left|\left|#1\right|\right|}

\newcommand{\card}[1]{\left|#1\right|}

\newcommand{\set}[1]{\left\{ #1\right\}}

\renewcommand{\leq}{\leqslant}

\usepackage{newfloat}
\usepackage{listings}
\DeclareCaptionStyle{ruled}{labelfont=normalfont,labelsep=colon,strut=off} %
\floatstyle{ruled}
\newfloat{listing}{tb}{lst}{}
\floatname{listing}{Listing}

\def\AUROC{\texttt{AUROC} $\uparrow$}
\def\AUPRIN{\texttt{AUPR-IN}}
\def\AUPROUT{\texttt{AUPR-OUT}}
\def\BENCHMARK{\texttt{LOFTER}}
\def\FPR{\texttt{FPR} $\downarrow$ }
\def\ERR{\texttt{Err}}

\def\LOF{\texttt{LOF}}
\def\BENCHMARK{\texttt{MILTOOD-C}}

\theoremstyle{plain}

\newtheorem{remark}{Remark}

\usepackage[capitalize,noabbrev]{cleveref}

\theoremstyle{plain}

\theoremstyle{definition}

\theoremstyle{remark}

\title{Unsupervised Layer-wise Score Aggregation  for Textual OOD Detection}

\author{ {\bf Maxime \textsc{Darrin}$^{1, 2, 3, 4}$} \quad {\bf Guillaume \textsc{Staerman}$^{4, 6, 7}$} \quad {Eduardo DC \textsc{Gomez}$^{4, 5, 6, 7}$} \quad {\bf Jackie~CK~\textsc{Cheung}$^{2, 3, 8}$} \quad {\bf Pablo \textsc{Piantanida}$^{1, 2, 4, 6}$} {\bf Pierre \textsc{Colombo}$^{4, 7, 10, 11}$}\\
}

\affiliations{
 { $^{1}$International Laboratory on Learning Systems, $^{2}$MILA - Quebec AI Institute, $^{3}$McGill University} \\ {  $^{4}$Université Paris-Saclay, $^{5}$ Laboratoire signaux et systèmes, $^{6}$CNRS, $^7$CentraleSupélec} \\ {$^8$ Canada CIFAR AI Chair, Mila, $^9$ INRIA, CEA, Paris, $^{10}$ Equal, Paris, $^{11}$ MICS} 

}

\usepackage{bibentry}

\begin{document}

\maketitle

\begin{abstract}
Out-of-distribution (OOD) detection is a rapidly growing field due to new robustness and security requirements driven by an increased number of AI-based systems. Existing OOD textual detectors often rely on anomaly scores (\textit{e.g.}, Mahalanobis distance) computed on the embedding output of the last layer of the encoder. In this work, we observe that OOD detection performance varies greatly depending on the task and layer output. More importantly, we show that the usual choice (the last layer) is rarely the best one for OOD detection and that far better results can be achieved, provided that an oracle selects the best layer. We propose a data-driven, unsupervised method to leverage this observation to combine layer-wise anomaly scores. In addition, we extend classical textual OOD benchmarks by including classification tasks with a more significant number of classes (up to 150), which reflects more realistic settings. On this augmented benchmark, we show that the proposed post-aggregation methods achieve robust and consistent results comparable to using the best layer according to an oracle while removing manual feature selection altogether. 
\end{abstract}

\section{Introduction}

With the increasing deployment of ML tools and systems, the issue of their safety and robustness is becoming more and more critical. Out-of-distribution robustness and detection have emerged as an important research direction~\cite{ood_review, liu2020energybased, winkens2020contrastive, hendrycks2016baseline, serra2019input, hendrycks2021many, mcallister2019robustness}. These OOD samples can cause the deployed AI system to fail as neural models rely heavily on previously seen concepts or patterns~\cite{jakubovitz2019generalization} and tend to struggle with anomalous samples~\cite{berend2020cats, bulusu2020anomalous} or new concepts. These failures can affect user confidence, or even rule out the adoption of AI in critical applications.

Distinguishing OOD samples (OUT) from in-distribution (IN) samples is a challenge when working on complex data structures (\textit{e.g.}, text or image) due to their high dimensionality. Although OOD detection has attracted much attention in computer vision \cite{huangdensity,wangout,fang2022is}, few studies focused on textual data. Furthermore, distortion and perturbation methods for sensitivity analysis used in computer vision are not suitable due to the discrete nature of text~\cite{lee2022gradient, schwinn2021identifying}.

A fruitful line of research \cite{lee2018simple, odin, esmaeilpour2022zero, xu2020deep, huang2020feature} focuses on adding simple filtering methods on top of pre-trained models without requiring retraining the model. They include plug-in detectors that rely on softmax-based- or hidden-layer-based- confidence scores~\cite{lee2018simple, odin, esmaeilpour2022zero, xu2020deep, huang2020feature}. Softmax-based detectors~\cite{liu2020energybased, pearce2021understanding, techapanurak2019hyperparameter} rely on the predicted probabilities to decide whether a sample is OOD. In contrast, hidden-layer-based scores (\textit{e.g.}, cosine similarity, data-depth \cite{colombobeyond}, or Mahalanobis distance \cite{lee2018simple}) rely on input embedding of the model encoder. In computer vision and more recently in natural language processing, these methods arbitrarily rely on either the embedding generated by the last layer of encoder \cite{podolskiy2021revisiting} or on the logits~\cite{wang2022vim, khalid2022rodd} to compute anomaly scores. While Softmax-based detectors can be applied in black-box scenarios, where one can only access the model's output, they have a very narrow view of the model's behaviour. In contrast, hidden-layer-based methods enable one to get deeper insights. They tend to yield better performance at the cost of memory and compute overhead.

\begin{figure*}[!ht]
    \centering
    \begin{subfigure}[b]{0.48\textwidth}
        \includegraphics[width=\textwidth]{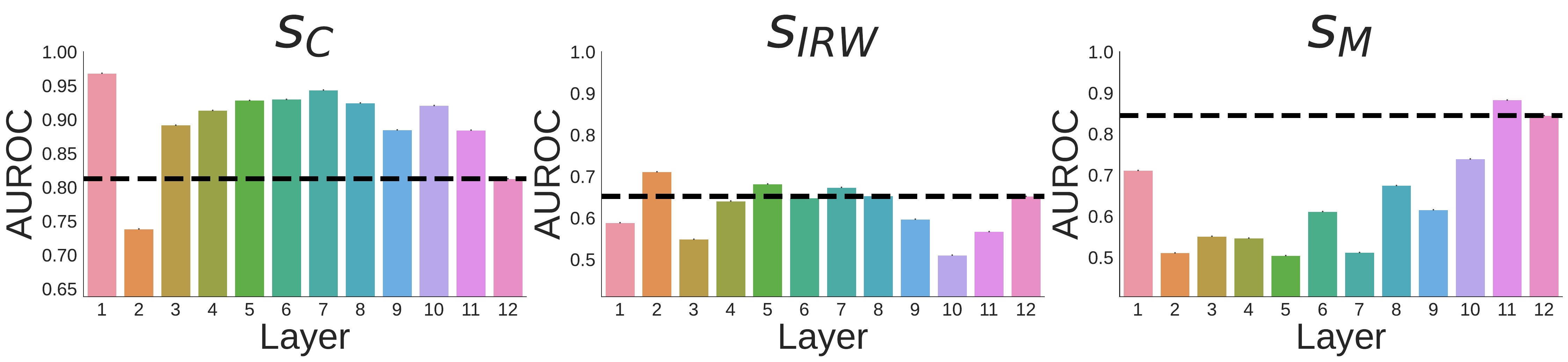}
        \caption{Emotion}
    \end{subfigure}    \begin{subfigure}[b]{0.48\textwidth}
        \includegraphics[width=\textwidth]{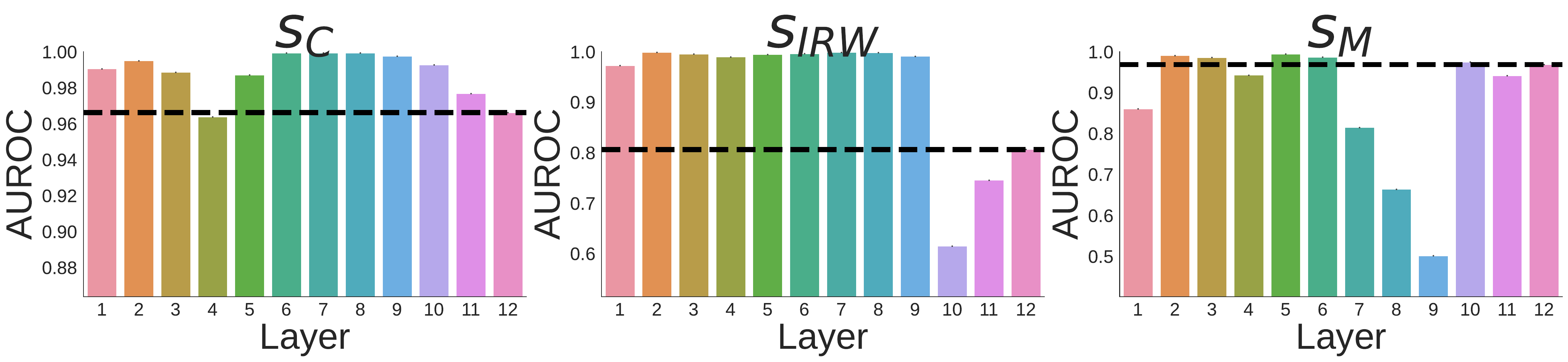}
        \caption{20-newsgroup}
    \end{subfigure}
    
    \begin{subfigure}[b]{0.48\textwidth}
        \includegraphics[width=\textwidth]{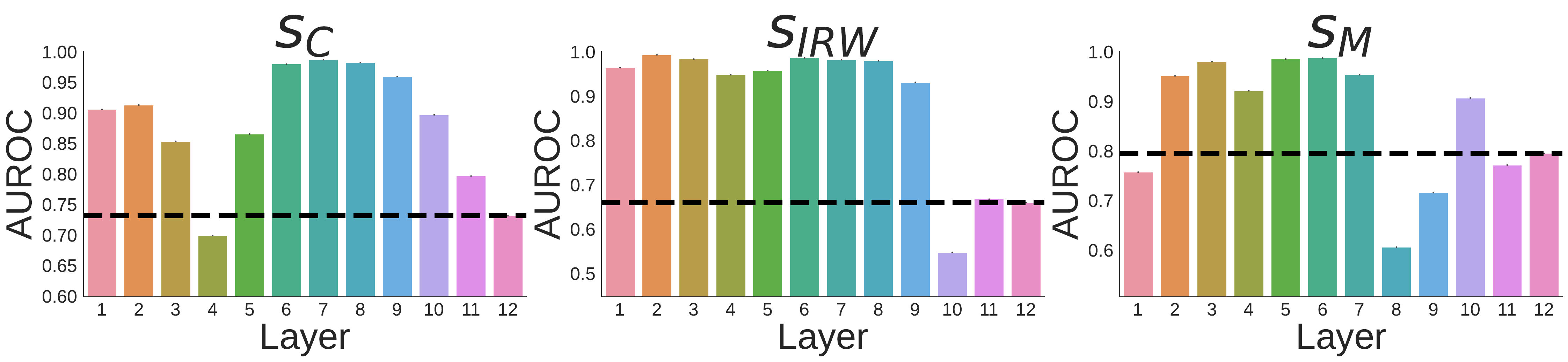}
        \caption{IMDB}
    \end{subfigure}\begin{subfigure}[b]{0.48\textwidth}
        \includegraphics[width=\textwidth]{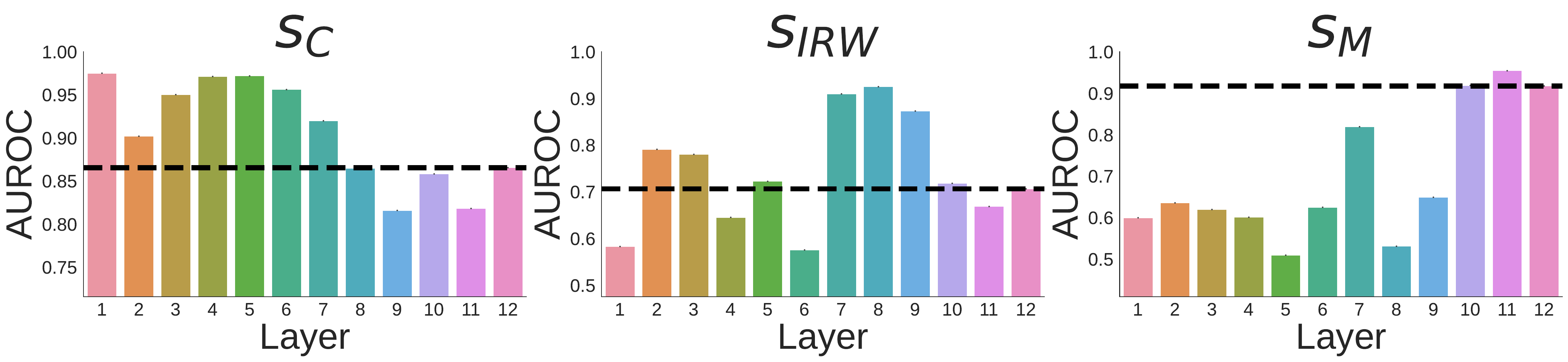}
        \caption{Trec}
    \end{subfigure}
    \caption{OOD detection performance in terms of {\AUROC} for each features-based OOD score (Mahalanobis distance ($s_M$), Maximum cosine similarity ($s_C$) and IRW ($s_{IRW}$)) computed at each layer of the encoder for different OOD datasets for a model fine-tuned on SST2. We observe that the performance of each metric on each layer varies significantly with the OOD task and that OOD detection based on the last layer (dark dotted line) rarely yields the best results.}
    \label{fig:perfs_per_layer}
\end{figure*}

We argue that the choice of the penultimate layers (\textit{i.e.}, the last layer, or the logits) ignores the multi-layer nature of the encoder and should be questioned. We give evidence that these representations are (i) not always the best choices (see Fig.~\ref{fig:perfs_per_layer}) and (ii) that leveraging information from all layers can be beneficial. We introduce a data-driven procedure to exploit the information extracted from existing OOD scores across all the different layers of the encoder.

\textbf{Our contribution can be summarized as follows:}

\noindent1.\textbf{We introduce a new paradigm.} Previous methods rely on a manual selection of the layer to be used, which ignores the information in the other layers of the encoder. We propose an automatic approach to aggregate information from all hidden layers without human (supervised) intervention. \textit{Our method does not require access to OOD samples and harnesses information available in all model layers by leveraging principled anomaly detection tools.}

\noindent2.\textbf{We conduct extensive experiments on our newly proposed benchmark}: We introduce {\BENCHMARK} A \textbf{\underline{M}}ultI \textbf{\underline{L}}ingual \textbf{\underline{T}}ext \textbf{\underline{OOD}} detection benchmark for \textbf{\underline{C}}lassification tasks. {\BENCHMARK} alleviates two main limitations of previous works: (i) contrary to previous work that relies on datasets involving a limited number of classes (up to $5$), {\BENCHMARK} includes datasets with a higher number of classes (up to $150$ classes); (ii) {\BENCHMARK} goes beyond the English-centric setting and includes French, Spanish, and German datasets. Our experiments involve four models and over $186$ pairs of IN and OUT datasets, which show that our new aggregation procedures achieve high performance. At the same time, previous methods tend to suffer a drop in performance in these more realistic scenarios.

\section{OOD detection for text classification}

\subsection{Background and notations}\label{sec:background}
We adopt a text classification setting and rely on the encoder section of a model. Let $\Omega$ be a vocabulary and $\Omega^*$ its Kleene closure\footnote{The Kleene closure corresponds to sequences of arbitrary size written with words in $\Omega$. Formally: $\Omega^* = \overset{\infty}{\underset{i = 0}{\bigcup}} \Omega^i$.}.  We consider $(X, Y)$ a random variable with values in $\Xrond \times \Yrond$ such that $\Xrond \subseteq \Omega^*$ is the textual input space, and $P_{XY}$ is its joint probability distribution. The set $\Yrond = \set{1, \ldots, C}$ represents the classes of a classification task and $\Prond(\Yrond) =  \set{\pbold \in [0, 1]^{\card{\Yrond} }: \sum_{i=1}^{\card{\Yrond}}
p_i
= 1} $ the probability simplex over the classes. It is assumed that we have access to a training set $\mathcal{D}_N = \{(\mathbf{x}_i,y_i) \}_{i=1}^N $ composed of independent and identically distributed (i.i.d) realizations of $P_{XY}$. The Out-Of-Distribution (OOD) detection problem consists of deciding whether a new, previously unseen sample comes (or not) from the IN distribution $P_{XY}$. The goal is to build a binary function $g:\mathcal{X} \rightarrow \{0,1\}$ based on the thresholding of an anomaly score $s: \mathcal{X} \rightarrow \mathbb{R}_+$ that separates IN samples from OOD samples. Namely, 
for a threshold $\gamma \in \mathbb{R}_+$,
we have: 
$$
g(\mathbf{x} , \gamma)~=~\left\{\begin{array}{ll}
1 & \text {if } s\left(\mathbf{x}\right) > \gamma, \\
0 & \text {if } s\left(\mathbf{x}\right) \leq \gamma.
\end{array}\right.
$$ 

\subsection{Building an OOD detector}\label{subsec:exist}
We assume that we have given a  classifier $f_\theta : \mathcal{X} \rightarrow \Prond(\mathcal{Y}$):
\begin{equation}\label{eq:deep_net}
    f_\theta= \operatorname{softmax} \circ~ h \circ~ f^\theta_L \circ f^\theta_{L-1} \circ \dots \circ f^\theta_1,
\end{equation} with $L > 1$ layers\footnote{For the sake of brevity, we omit the parameters $\theta$ in the following.}, where $f_{\ell}: \mathbb{R}^{d_{\ell-1}} \rightarrow \mathbb{R}^{d_{\ell}}$ is the $\ell$-th
layer of the encoder with $d_{\ell}$ being the dimension of the latent space after the $\ell$-th layer ($d_0 = d$). It is worth noting that in the case of transformers \cite{vaswani2017attention}, all latent spaces have the same dimension. Finally, $h$ represents the logit function of the classifier. 

To compute the anomaly score $s$ from $f_\theta$, OOD approaches rely on the hidden representations of the (multilayer) encoder. For $\xbold \in \Xrond$ an input sequence, we denote $\zbold_{\ell}=(f_{\ell} \circ \dots \circ f_1)(\mathbf{x})$ its latent representation at layer $\ell$. The latent representation obtained after the $\ell$-th layer of the training set is denoted as $\Drond_N^\ell=\{ (\zbold_{\ell, i}, y_i)\}_{i=1}^N$. Furthermore, we denote by $\Drond_N^{\ell,y}$  the restriction of $\Drond_N^\ell$ to the samples with label $y$, i.e., $\Drond_N^{\ell, y} = \{ (\zbold_{\ell, i}, y_i) \in \Drond^\ell_N : y_i = y \}$ with $N_{y}= | \Drond_N^{\ell,y}|$ indicates the cardinal of this set.

Feature-based OOD detectors usually rely on three key elements:
\begin{enumerate}[wide, topsep=0pt, labelwidth=0pt, labelindent=0pt]

    \item[(i)] \textbf{Selecting features}: the layer $\ell$ whose representation is considered to be the input of the anomaly score.
    \item[(ii)] \textbf{A notion of an anomaly (or novelty) score} built on the mapping $\Drond_N^\ell$ of the training set on the chosen feature space. We can build such a score $s(\cdot,\Drond_N^\ell)$ defined on $\mathbb{R}^d \times \left(\mathbb{R}^d\right)^N$ for any notion of abnormality. 

    \item[(iii)] \textbf{Setting a threshold} to build the final decision function.
\end{enumerate}

\begin{remark}
\textbf{Choice of the threshold.} 
    To select $\gamma$, we follow previous work \cite{colombobeyond, picot2023adversarial} by choosing a number of training samples (\textit{i.e.}, ``outliers") the detector can wrongfully detect. A classical choice is to set this proportion to $95\%$. 
\end{remark}

\subsection{Popular Anomaly Scores}\label{sec:popular_anomaly}
In what follows, we present three common anomaly scores for step (ii) of the previously mentioned procedure.

\noindent\textbf{Mahalanobis distance.} Authors of \citet{lee2018simple} (see also \cite{podolskiy2021revisiting}) propose to compute the Mahalanobis distance on the abstract representations of each layer and each class.  Precisely, this distance is given by:\begin{equation*}
s_{M}(\zbold_\ell, \Drond_N^{\ell,y})= \left(\zbold_\ell-\mu_{\ell,y}\right)^{\top} \Sigma_{\ell,y}^{-1}\left(\zbold_\ell-\mu_{\ell,y}\right)
\end{equation*} on each layer $\ell$ and each class $y$ where  $\mu_{\ell,y}$ and $\Sigma_{\ell,y}$ are the estimated class-conditional mean and covariance matrix computed on $\Drond_N^{\ell,y}$, respectively. The final score from \citet{lee2018simple} is obtained by choosing the minimum of these scores over the classes on the penultimate encoder layer.

\noindent \textbf{Integrated Rank-Weighted depth.}  \citet{colombobeyond} propose to leverage the Integrated Rank-Weighted (IRW) depth \citep{ramsay2019integrated,AIIRW}. Similar to the Mahalanobis distance, the IRW data depth measures the centrality/distance of a point to a point cloud. For the $\ell$-th layer, an approximation of the  IRW depth can be defined as:

\begin{align*}
    s_{\text{IRW}}(\zbold_\ell, \Drond_N^{\ell,y})= \frac{1}{n_{\mathrm{proj}}} \sum_{k=1}^{n_{\mathrm{proj}}} \min &\Bigg \{ \frac{1}{n}\sum_{i=1}^{N_{y}} \mathbb{1} \left\{ g_{k,i}(\zbold_\ell) \leq 0 \right\},  \\&  \frac{1}{n}\sum_{i=1}^{N_y} \mathbb{1} \left\{g_{k,i}(\zbold_\ell) > 0 \right\} \Bigg \},
\end{align*}
where $ g_{k,i}(\zbold_\ell)= \langle u_k,\zbold_{\ell, i}-\zbold_\ell \rangle$, $u_k\in \mathbb{S}^{d-1}, \zbold_{\ell, i} \in \Drond_N^{\ell,y}$ where $\mathbb{S}^{d-1}=\{x\in \mathbb{R}^d \; : \; ||x||=1 \}$ is the unit hypersphere and $n_{\mathrm{proj}}$ is the number of directions sampled on the sphere. 

\noindent\textbf{Cosine similarity.} \citet{zhou2021contrastive} propose to compute the maximum cosine similarity  between the embedded sample  $\zbold_\ell$ and the training set $\Drond_N^\ell$ at layer $\ell$:
\begin{equation*}
    s_{C}(\zbold_\ell, \Drond_N^\ell)= - \underset{\zbold_{\ell,i} \in \Drond_N^\ell}{\max}\; \;~\frac{ \langle \zbold_\ell , \zbold_{\ell, i} \rangle}{\norm{\zbold_\ell} \norm{\zbold_{\ell, i}}}, 
\end{equation*} %
where $\langle \cdot, \cdot \rangle$ and $\norm{\cdot}$ denote the Euclidean inner product and norm, respectively. They also choose the penultimate layer. It is worth noting they do not rely on a per-class decision.

\subsection{Related works and limitations}\label{subsec:all}

We claim that the choice of layer is crucial in textual OOD detection; we report in Fig.~\ref{fig:perfs_per_layer} the OOD performance of popular detectors described in Sec.~\ref{sec:popular_anomaly}, applied at each layer of the encoder. We observe a high variability across different layers. The last layer is rarely the best-performing layer, and there is room for improvement if we could choose the best possible layer or gather useful information from all of them. This observation is consistent with the literature, as neural networks are known to extract different information and construct different abstractions at each layer~\cite{ilin2017abstraction,KOZMA2018203, features_extraction}.

\noindent \textbf{Works relying on a manually selected layer.} The choice of layer for step (i) in Sec. \ref{subsec:exist} is not usually a question. Most work arbitrarily relies on the logits~\cite{odin,liu2020energybased} or the last layer of the encoder~\cite{contrastive_training_ood,podolskiy2021revisiting,Sun2022OutofdistributionDW, ren2019likelihoodratios, gomes2022igeood,ood_review, hendrycks2016baseline, wang2022vim}. We argue that these choices are unjustified and that previous work gives up on important information in the other layers.

\noindent \textbf{Works that feature layer aggregation.} Besides~\citeauthor{colombobeyond}, which proposes to aggregate the features instead of aggregating the OOD scores as we do, we are unaware of any other work suggesting layer aggregation for textual OOD detection. There, however, exist several adjacent works in computer vision. \citeauthor{abdelzad2019detecting} proposes a method which starts by finding the best layer for OOD detection for a validation set and uses that layer afterwards. Still, there is no adaptation for different datasets. Other works such as \citeauthor{lin2021mood, Ballas_2022} focus on training models with specific architectures (for computer vision) to leverage early-stage exit or different information from different layers to detect OOD samples or improve generalization on OOD samples. In addition, these works rely on image-specific properties that do not translate well in text (lossy compression for example). More closely related work are~\citeauthor{wang2022layer, lee2018simple}. The first proposes to compute OOD scores at each layer (1-SVM), and they select the layer that yields the highest confidence in terms of margin. However, their work features a single OOD score and a single aggregation method in computer vision. In contrast, we propose a more general framework and systematically evaluate different OOD scores and aggregation procedures. The second proposes to learn a linear combination of Mahalanobis scores but does not explore OOD score aggregation or layer selection systematically. In addition, both of these works have been proposed in computer vision and show that aggregation does not lead to significant improvements. This dramatically contrasts with our findings in textual models, consistent with the literature~\cite {picot2023adversarial, raghuram}.

We propose to compute standard OOD scores on each layer of the encoder (and not only on the logits or the representation generated by the last layer) and to aggregate this score in an unsupervised fashion to select and combine the most relevant following the task at hand.

\section{Leveraging information from all layers}
In this section, we describe our aggregation methods that use the information available in the different layers of the encoder.

\subsection{Problem Statement}

For an input $\xbold \in \mathcal{X}$ and a training dataset $\mathcal{D}_N$, we obtain their set of embedding representation sets: $\{\zbold_\ell\}_{\ell=1}^L$ and $\{\mathcal{D}^{\ell}_N\}_{\ell=1}^{L}$, respectively. Given an anomaly score function $s: \; \mathbb{R}^d \times \left(\mathbb{R}^d \right)^N  \rightarrow \mathbb{R}$ (e.g.,  those described in Sec. \ref{sec:popular_anomaly}), we define the OOD score set of an input $\xbold$ as $\mathcal{S}_s(\xbold; \mathcal{D}_N) = \{\{s(\zbold_\ell; \mathcal{D}^{\ell, y}_N) \}_{\ell=1}^L \}_{y=1}^{C} \in \mathbb{R}^{L\times C}$. Similarly, it is possible to obtain a reference set of  $\mathcal{R}(\mathcal{D}_N) = \{ \mathcal{S}_s(\xbold; \mathcal{D}_N), \; \forall \; (\xbold,y) \in  \mathcal{D}_N \}$ from the training data\footnote{When using the cosine similarity, which does not rely on a per-class decision, $\mathcal{S}_s(\xbold; \mathcal{D}_N)$ is reduced to $\{s(\zbold_\ell; \mathcal{D}^{\ell}_N) \}_{\ell=1}^L$.}.  In what follows, we aim to answer the following question.

\textit{Can we leverage all the information available in $\mathcal{S}_s(\xbold; \mathcal{D}_N)$  and/or $\mathcal{R}(\mathcal{D}_N)$ to build an OOD detector?}

\subsection{Proposed Framework}

Our framework aims at comparing the set of scores of a sample to the sets of scores of a reference relying on principled anomaly detection algorithms. 

The goal of this work is to propose a data-driven aggregation method of OOD scores\footnote{We do not assume that we have access to OOD samples as they are often not available.}, $\operatorname{Agg}$. $\operatorname{Agg}$ is defined as:
\begin{align*}
     \operatorname{Agg} : \quad \mathbb{R}^{L\times C} \times \left(\mathbb{R}^{L\times C}\right)^N  & \rightarrow \mathbb{R} \\
      (\mathcal{S}_s(\xbold; \mathcal{D}_N),\mathcal{R}(\mathcal{D}_N)) &  \rightarrow \operatorname{Agg}(\mathcal{S}_s, \mathcal{R}), 
\end{align*}
where  $\xbold$ denotes the input sample. 

\noindent \textbf{Intuition.} This framework allows us to consider the whole trace of a sample through the model. This formulation has two main advantages: it avoids manual layer selection and enables us to leverage information from all the encoder layers.

We propose two families of approaches: (i) one solely relies on the score set $\mathcal{S}_s(\xbold; \mathcal{D}_N)$ (corresponding to a no-reference scenario and denoted as $\operatorname{Agg}_{\emptyset}$) and (ii) the second one (named reference scenario) leverages the reference set $\mathcal{R}(\mathcal{D}_N)$.

\begin{remark}
    It is worth noting that our framework through  $\operatorname{Agg}_{{\emptyset}set}$ or $\operatorname{Agg}$ naturally includes previous approaches \cite{lee2018simple,zhou2021contrastive,colombobeyond}. For example, the detector of \citet{lee2018simple} can be obtained by defining  $\operatorname{Agg}_{{\emptyset}}$ as the minimum of the penultimate line of the matrix $\mathcal{S}_s(\xbold, \mathcal{D}_N)$.
    \label{rem:captures_previous_methods}
\end{remark}

\subsection{Detailed Aggregation Procedures}

\begin{figure*}
    \centering
\includegraphics[width=0.75\textwidth, trim={1.5cm 9cm 1cm 0cm}]{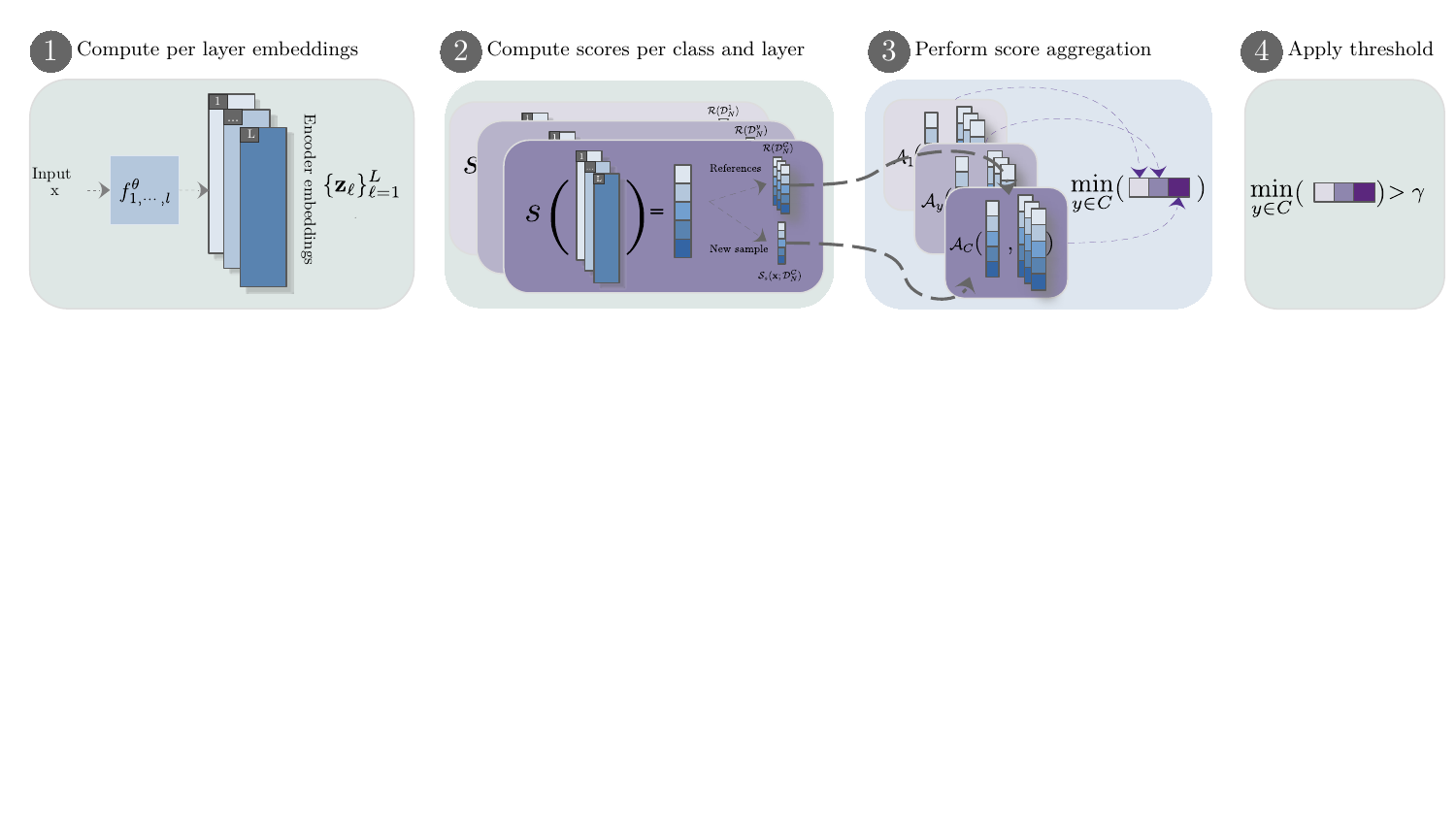}
    \caption{Schema of our aggregation procedure. (1) We extract the embeddings at each layer of the encoder for every sample. (2) We compute the per-class scores for a reference set and the new sample to be evaluated for each layer embedding. (3) We aggregate the scores over every layer to get an aggregated per-class score before taking the min score over the classes. (4) Finally, we apply the threshold on this minimum.}
    \label{fig:schema}
\end{figure*}

\textbf{Intuition.} Our framework through $\operatorname{Agg}$ and $\operatorname{Agg}_{\emptyset}$ requires two types of operations to extract a single score from  $\mathcal{S}_s(\xbold, \mathcal{D}_N)$ and $\mathcal{R}_s( \mathcal{D}_N)$: one aggregation operation over the layers and one aggregation operation over the classes, where necessary.

\textbf{Our framework in a nutshell.} We assume we are given an anomaly score, $s$, that we want to enhance by leveraging all the layers of the encoder.  For a given input $\xbold$, our framework follows $4$ steps (see Fig.~\ref{fig:schema} for a depiction of the procedure):

\begin{enumerate}[wide, topsep=0pt, labelwidth=0pt, labelindent=0pt, itemsep=0pt, parsep=0pt]
    \item Compute the embeddings $\{\zbold_l\}_{l=1}^L$ for $\xbold$ and every element of $\Drond_N$.
    \item Form $\mathcal{S}_s(\xbold; \mathcal{D}_N)$ and $\mathcal{R}(\mathcal{D}_N)$ using the score $s$.
    \item Perform $\operatorname{Agg}_{\emptyset}$ or $\operatorname{Agg}$:
    \begin{enumerate}
        \item (\emph{per layer}) Aggregate score information over the layers to obtain a vector composed of $C$ scores. 
        \item (\emph{per class})  Take the minimum value of this vector. 
    \end{enumerate}
    \item Apply a threshold $\gamma$ on that value.
\end{enumerate}

Step (3.b). is inspired by the OOD literature \citep{lee2018simple,colombobeyond}. It relies on the observation that if the input sample is IN-distribution, it is expected to have at least one low score in the class vector, whereas an OOD sample should only have high scores equivalent to a high minimum score.

\subsubsection{No-reference scenario ($\operatorname{Agg}_{\emptyset}$)} In the no-reference scenario, we have access to a limited amount of information. We
thus propose to rely on simple statistics to aggregate the OOD scores available in $\mathcal{S}_s(\xbold; \mathcal{D}_N)$ to compute step (3). of the proposed procedure. Precisely, we use the \textit{average}, the minimum (\textit{min}), the median (\textit{med}), and \textit{coordinate} (see Remark~\ref{rem:captures_previous_methods}) operators on the column of the matrix $\mathcal{S}_s(\xbold; \mathcal{D}_N)$.

\subsubsection{Data-driven scenario ($\operatorname{Agg}$)}
In the data-driven scenario, $\operatorname{Agg}$ also has access to the set of reference OOD scores (\textit{i.e.}, $\mathcal{R}_s(\mathcal{D}_N)$) for the given OOD score $s$. The goal, then, is to compare the score set $\mathcal{S}_s(\xbold; \mathcal{D}_N)$ of the input with this reference set $\mathcal{R}_s(\mathcal{D}_N)$ to obtain a score vector of size $C$. \textit{In the following, we propose an original solution for the layer operation.} 

For the \emph{per layer} operation we rely on an anomaly detection algorithm for each class $\mathcal{A}_{y}$ defined as:
\begin{align}
    \mathcal{A}_{y}: \mathbb{R}^L \times (\mathbb{R}^L)^{N_y}  & \rightarrow \mathbb{R} \nonumber\\
\mathbf{s}_{y} \times \mathcal{R}_{y}& \mapsto  \mathcal{A}_{y}(\mathbf{s}_{y}, \mathcal{R}_{y}),
\end{align}  
where $\mathbf{s}_{y}=\{s(\zbold_\ell; \mathcal{D}^{\ell, y}_N) \}_{\ell=1}^L$ and $\mathcal{R}_{y}=\mathcal{R}(\mathcal{D}_N^{y}) $.

\begin{remark}
     $\mathcal{A}_{y}$ is trained on the reference set $\mathcal{R}_{y}$ for each class and \textbf{thus does not involve any OOD samples}. The score returned for a vector $\mathbf{s}_{y}$ is the prediction score associated with the trained algorithm.
\end{remark}
\begin{remark}
We define a per-class decision for $\operatorname{Agg}$ since it has been shown to be significantly more effective than global scores~\cite{huang2021mos}. It is the approach chosen by most \textit{state-of-the-art-methods}. We have validated this approach by conducting extensive experiments. We refer the reader to  Sec.~\ref{sec:additional_algorithms} for further discussion.
\end{remark}
We propose several popular anomaly detection algorithms. First, we offer to reuse common OOD scores ($s_M$, $s_C$, $s_{\text{IRW}}$) as aggregation methods(see Sec.~\ref{sec:additional_algorithms} for more details): they are now trained on the reference set of sets of OOD scores $\Rrond_s(\Drond_N)$ and provide a notion of anomaly for the trace of a sample through the model. We also compute the median and average score as natural baselines for the aggregation setting. In addition, we propose more elaborate anomaly detection algorithms such as Isolation Forest (\texttt{IF}) \cite{LiuTZ08} and the Local Outlier Factor (\texttt{LOF}) \cite{LOF}. Below, we briefly recall the general insights of each of these algorithms. It is important to emphasize that our framework can accommodate any anomaly detection algorithms (further details are given in Sec.~\ref{sec:additional_algorithms}, and in Sec.~\ref{sec:computational_costs}, we provide a discussion about the computational overhead of the different methods).

\noindent\textit{Local Outlier Factor.} This method compares a sample's density with its neighbours' density. Any sample with a lower density than its neighbours is regarded as an outlier.

\noindent\textit{Isolation Forest.} This popular algorithm is built on the idea that anomalous instances should be easily distinguished from normal instances. It leads to a score that measures the complexity of separating a sample from others based on the number of necessary decision trees required to isolate a data point. It is computationally efficient, benefits from stable hyper-parameters, and is well suited to the unsupervised setting.

\subsection{Comparison to Baseline Methods}

\noindent Current State-of-the-art methods for OOD detection on textual data have been recently provided in \citet{colombobeyond} (\texttt{PW}). They aggregate the hidden layers using Power means~\cite{hardy1952inequalities,ruckle2018concatenated} and then apply an OOD score on this aggregated representation. They achieved previous SOTA performance by coupling it with the IRW depth and proposed a comparison with Mahalanobis and Cosine versions. We reproduce these results as it is a natural baseline for aggregation algorithms.

\noindent\textbf{Last Layer.} Considering that the model's last layer or logits should output the most abstract representation of an input, it has been the primary focus of attention for OOD detection. It is a natural choice for any architecture or model and therefore removes the hurdle of selecting features for different tasks and architectures. For this heuristic, we obtain OOD scores using the Mahanalobis distance (as in \cite{lee2018simple}), the IRW score (as in \cite{colombobeyond}), and the cosine similarity (as in \cite{contrastive_training_ood}).

\noindent\textbf{Additional methods.} It is common on OOD detection methods to report the Maximum Softmax Prediction (MSP) \cite{hendrycks2016baseline} as well as the Energy Score ($E$) \cite{liu2020energybased}.

\section{{\BENCHMARK}: A more realistic benchmark}
In this section, we highlight the limitations of existing benchmarks and introduce our own: {\BENCHMARK} A \textbf{\underline{M}}ultI \textbf{\underline{L}}ingual \textbf{\underline{T}}ext \textbf{\underline{OOD}} for classification tasks.

\subsection{Limitation of Existing Benchmarks}

\noindent \textbf{Number of classes.} Text classification benchmarks for OOD detection often consist of sentiment analysis tasks involving a small number of classes~\cite{fang2015sentiment,kharde2016sentiment}. Those tasks with a larger number of classes have been mostly ignored in previous OOD detection benchmarks~\cite {colombobeyond,li2021k,zhou2021contrastive}. However, real-world problems do involve vastly multi-class classification tasks~\cite{Casanueva2020}. Previous work in computer vision found that these problems require newer and carefully tuned methods to enable OOD detection in this more realistic setting~\cite {imagenet, tiny-imagenet}.

\noindent \textbf{Monolingual datasets.} Most methods have been tested on architectures tailored for the English language \cite{colombobeyond,li2021k,arora2021types}. With inclusivity and diversity in mind \cite{ruder2022statemultilingualai,van2022writing}, it is necessary to assess the performance of old and new OOD detection methods on a variety of languages~\cite{srinivasan2021predicting, de2020s, baheti2021just, zhang2022mdia}.

\subsection{A more realistic benchmark}

We now present {\BENCHMARK}, which addresses the aforementioned limitations. It consists of more than $25$ datasets involving up to $150$ classes and $4$ languages. 

\noindent \textbf{Dataset selection.}  We gathered a large and diverse benchmark encompassing many shift typologies, tasks, and languages. It covers $27$ datasets in $4$ different languages (\textit{i.e.}, English, German, Spanish, and French) and classifications tasks involving $2$ to $150$ classes. Following standard protocol \cite{hendrycks2020pretrained}, we train a classifier for each in-distribution dataset (\texttt{IN-DS}) while the OOD dataset (\texttt{OUT-DS}) is coming from a different dataset. We provide a comprehensive list of the $180$ pairs we considered in Sec.~\ref{sec:ood_pairs}. It is an order of magnitude larger than recent concurrent work from~\cite{colombobeyond}.

\noindent \textit{English benchmark.} We relied on the benchmark proposed by \citet{zhou2021contrastive, hendrycks2020pretrained}. It features three types of \texttt{IN-DS}: sentiment analysis (\textit{i.e.}, SST2 \cite{socher-etal-2013-recursive}, IMDB \cite{maas-etal-2011-learning}), topic classification (\textit{i.e.}, 20Newsgroup \cite{joachims1996probabilistic}) and question answering (\textit{i.e.}, TREC-10 and TREC-50 \cite{li-roth-2002-learning}). We also included the Massive~\cite{fitzgerald2022massive} dataset and the Banking~\cite{Casanueva2020} for a larger number of classes and NLI datasets (\textit{i.e.}, RTE \cite{Burger:05,Hickl:06} and MNLI \cite{williams-etal-2018-broad}) following~\citet{colombobeyond}. To go one step further in terms of number of classes, we considered HINT3\cite{hint3} and clink~\cite{clink}. We form IN and OOD pairs between the aforementioned tasks.

\noindent \textit{Beyond English-centric tasks.}\footnote{We did not work on language changes because they were easily detected with all the methods considered. Instead, we focus on intra-language drifts.} For language-specific datasets, we added the same tasks as for English when available and extended it with language-specific datasets such as the PAWS-S datasets~\cite{pawsx2019emnlp}, film reviews in French and Spanish~\cite{allocinedataset}. For French and German, we also added the Swiss judgments datasets~\cite{niklaus2022empirical}. Finally, we added different tweet classification tasks for each language (English, German, Spanish and French)~\cite{zotova-etal-2020-multilingual, barbieri-etal-2022-xlm}.

\noindent\textbf{Model selection.} To ensure that our results are consistent not only across tasks and shifts, but also across model architectures, we train classifiers based on $6$ different Transformer \cite{vaswani2017attention} decoders: BERT~\cite{devlin2018bert} (base, large and multilingual versions), DISTILBERT~\cite{sanh2019distilbert} and RoBERTa~\cite{liu2019roberta} (base and large versions) fine-tuned on each task.

\noindent \textbf{Evaluation metrics.} The OOD detection problem is a binary classification problem where the positive class is OUT. We follow concurrent work \cite{colombobeyond, darrin2022rainproof} and evaluate our detector using threshold-free metrics such as {\AUROC}, {\AUPRIN}/{\AUPROUT} and threshold based metrics such as {\FPR} at $95\%$ and {\ERR}. For sake of brevity, we report detailed definitions of the metrics in Sec.~\ref{sec:perfs_metrics_app}.

\section{Experimental Results}

\subsection{Quantifying Aggregation Gains}

\noindent\textbf{Overall results.} \textbf{Data driven aggregation methods (\textit{i.e.}, with reference) consistently outperform any other baselines or tested methods by a significant margin} (see Table~\ref{tab:average_per_aggr} and Table~\ref{tab:overall_average_app}) on our extensive {\BENCHMARK} benchmark. According to our experiments, the best combination of hidden feature-based OOD score and aggregation function is to use the Maximum cosine similarity as the underlying OOD score and to aggregate these scores using the IRW data depth ($s_\text{IRW}$). A first time to get the abnormality of the representations of the input and a second time to assess the abnormality of the set of layer-wise scores through the model. It reaches an average {\AUROC} of $0.99$ and a {\FPR} of $0.02$. It is a gain of more than $6\%$ compared to the previous \textit{state-of-the-art} methods in terms of {\AUROC} and more than $90\%$ in {\FPR}.

\noindent \textbf{Most versatile aggregation method.} While the $s_C$ and $s_{\text{IRW}}$ used as aggregation methods achieve excellent performance when paired with $s_C$ as the underlying OOD score, they fail to aggregate as well other underlying scores. Whereas the isolation forest algorithm is a more versatile and consistent data-driven aggregation method: it yields performance gain for every underlying OOD score.

\noindent \textbf{Performance of common baselines.} We show that, on average, using the last layer or the logits as features to perform OOD detection leads to poorer results than almost every other method. \textbf{It is interesting to point out that this is not the case in computer vision~\cite{ood_review, raghuram, picot2023adversarial, lee2018simple}. This finding further motivates the development of OOD detection methods tailored for text.}

    \begin{table}
    \centering
    \resizebox{0.45\textwidth}{!}{
\begin{tabular}{llrrrrrrrrrr}
\toprule
 & $\mathcal{A}$ & \multicolumn{2}{c}{\AUROC} & \multicolumn{2}{c}{\FPR} & \multicolumn{2}{c}{\ERR} \\
\midrule
\multirow[c]{5}{*}{$\operatorname{Agg}$} & $s_M$ & 0.88 & \color[HTML]{A0A1A3} ±0.16 & 0.32 & \color[HTML]{A0A1A3} ±0.35 & 0.16 & \color[HTML]{A0A1A3} ±0.19 \\
 & $s_{C}$ & 0.91 & \color[HTML]{A0A1A3} ±0.16 & \underline{\bfseries 0.21} & \color[HTML]{A0A1A3} ±0.35 & 0.13 & \color[HTML]{A0A1A3} ±0.21 \\
 & $s_{IRW}$ & 0.87 & \color[HTML]{A0A1A3} ±0.18 & 0.30 & \color[HTML]{A0A1A3} ±0.39 & 0.16 & \color[HTML]{A0A1A3} ±0.21 \\
 & \texttt{IF} & \underline{\bfseries 0.92} & \color[HTML]{A0A1A3} ±0.13 & \bfseries \underline{0.21} & \color[HTML]{A0A1A3} ±0.31 & \underline{\bfseries 0.10} & \color[HTML]{A0A1A3} ±0.15 \\
 & \texttt{LOF} & 0.87 & \color[HTML]{A0A1A3} ±0.15 & 0.37 & \color[HTML]{A0A1A3} ±0.36 & 0.19 & \color[HTML]{A0A1A3} ±0.22 \\
\cline{1-8}
\multirow[c]{3}{*}{$\operatorname{Agg}_{\emptyset}$} & Mean & \underline{0.84} & \color[HTML]{A0A1A3} ±0.18 & \underline{0.43} & \color[HTML]{A0A1A3} ±0.40 & 0.24 & \color[HTML]{A0A1A3} ±0.25 \\
 & Med. & 0.83 & \color[HTML]{A0A1A3} ±0.17 & 0.46 & \color[HTML]{A0A1A3} ±0.39 & 0.25 & \color[HTML]{A0A1A3} ±0.25 \\
 & \texttt{PW} & 0.82 & \color[HTML]{A0A1A3} ±0.17 & 0.48 & \color[HTML]{A0A1A3} ±0.39 & \underline{0.23} & \color[HTML]{A0A1A3} ±0.23 \\
\cline{1-8}
\multirow[c]{4}{*}{Bas.} & $E$ & 0.83 & \color[HTML]{A0A1A3} ±0.18 & \underline{0.39} & \color[HTML]{A0A1A3} ±0.30 & \underline{0.19} & \color[HTML]{A0A1A3} ±0.18 \\
 & L. L. & \underline{0.84} & \color[HTML]{A0A1A3} ±0.17 & 0.42 & \color[HTML]{A0A1A3} ±0.37 & 0.20 & \color[HTML]{A0A1A3} ±0.22 \\
 & Logs. & 0.75 & \color[HTML]{A0A1A3} ±0.16 & 0.60 & \color[HTML]{A0A1A3} ±0.34 & 0.30 & \color[HTML]{A0A1A3} ±0.25 \\
 & MSP & 0.83 & \color[HTML]{A0A1A3} ±0.17 & 0.39 & \color[HTML]{A0A1A3} ±0.28 & 0.19 & \color[HTML]{A0A1A3} ±0.18 \\
\cline{1-8}
\bottomrule
\end{tabular}

}
\caption{Overall average performance of each aggregation method for all architectures, tasks, and OOD scores. The best method overall is highlighted in bold and the best methods per setting are underlined.}
\label{tab:average_per_aggr}
\end{table}

\begin{table} 
\resizebox{0.45\textwidth}{!}{
\begin{tabular}{lllrrrr}
\toprule
 &  & & \multicolumn{2}{c}{\AUROC} & \multicolumn{2}{c}{\FPR} \\
 & Ours & Agg. &  &  &  &  \\
\midrule
$E$ & Bas. & $E$ & \underline{0.83} & \color[HTML]{A0A1A3} ±0.18 & \underline{0.39} & \color[HTML]{A0A1A3} ±0.31 \\
\cline{1-7} \cline{2-7}
\multirow[c]{10}{*}{$s_M$} & \multirow[c]{5}{*}{$\operatorname{Agg}$} & $s_M$ & 0.90 & \color[HTML]{A0A1A3} ±0.14 & 0.27 & \color[HTML]{A0A1A3} ±0.33 \\
 &  & $s_{C}$ & 0.88 & \color[HTML]{A0A1A3} ±0.17 & 0.32 & \color[HTML]{A0A1A3} ±0.40 \\
 &  & $s_{IRW}$ & 0.81 & \color[HTML]{A0A1A3} ±0.20 & 0.44 & \color[HTML]{A0A1A3} ±0.42 \\
 &  & \texttt{IF} & \underline{0.94} & \color[HTML]{A0A1A3} ±0.10 & \underline{0.19} & \color[HTML]{A0A1A3} ±0.25 \\
 &  & \texttt{LOF} & 0.87 & \color[HTML]{A0A1A3} ±0.15 & 0.39 & \color[HTML]{A0A1A3} ±0.37 \\
\cline{2-7}
 & \multirow[c]{3}{*}{$\operatorname{Agg}_{\emptyset}$} & Mean & 0.74 & \color[HTML]{A0A1A3} ±0.18 & \underline{0.59} & \color[HTML]{A0A1A3} ±0.39 \\
 &  & Median & 0.75 & \color[HTML]{A0A1A3} ±0.17 & 0.61 & \color[HTML]{A0A1A3} ±0.35 \\
 &  & \texttt{PW} & \underline{0.80} & \color[HTML]{A0A1A3} ±0.17 & 0.61 & \color[HTML]{A0A1A3} ±0.38 \\
\cline{2-7}
 & \multirow[c]{2}{*}{Bas.} & Last layer & \underline{0.92} & \color[HTML]{A0A1A3} ±0.11 & \underline{0.25} & \color[HTML]{A0A1A3} ±0.31 \\
 &  & Logits & 0.71 & \color[HTML]{A0A1A3} ±0.14 & 0.65 & \color[HTML]{A0A1A3} ±0.27 \\
\cline{1-7} \cline{2-7}
\multirow[c]{10}{*}{$s_{C}$} & \multirow[c]{5}{*}{$\operatorname{Agg}$} & $s_M$ & 0.93 & \color[HTML]{A0A1A3} ±0.11 & 0.20 & \color[HTML]{A0A1A3} ±0.27 \\
 &  & $s_{C}$ & 0.98 & \color[HTML]{A0A1A3} ±0.10 & 0.04 & \color[HTML]{A0A1A3} ±0.19 \\
 &  & $s_{IRW}$ & \underline{\bfseries 0.99} & \color[HTML]{A0A1A3} ±0.07 & \underline{\bfseries 0.02} & \color[HTML]{A0A1A3} ±0.15 \\
 &  & \texttt{IF} & 0.94 & \color[HTML]{A0A1A3} ±0.14 & 0.12 & \color[HTML]{A0A1A3} ±0.29 \\
 &  & \texttt{LOF} & 0.93 & \color[HTML]{A0A1A3} ±0.11 & 0.20 & \color[HTML]{A0A1A3} ±0.26 \\
\cline{2-7}
 & \multirow[c]{3}{*}{$\operatorname{Agg}_{\emptyset}$} & Mean & \underline{0.93} & \color[HTML]{A0A1A3} ±0.12 & 0.25 & \color[HTML]{A0A1A3} ±0.33 \\
 &  & Median & 0.92 & \color[HTML]{A0A1A3} ±0.12 & 0.27 & \color[HTML]{A0A1A3} ±0.34 \\
 &  & \texttt{PW} & 0.93 & \color[HTML]{A0A1A3} ±0.11 & \underline{0.19} & \color[HTML]{A0A1A3} ±0.27 \\
\cline{2-7}
 & \multirow[c]{2}{*}{Bas.} & Last layer & \underline{0.92} & \color[HTML]{A0A1A3} ±0.11 & \underline{0.22} & \color[HTML]{A0A1A3} ±0.26 \\
 &  & Logits & 0.81 & \color[HTML]{A0A1A3} ±0.17 & 0.52 & \color[HTML]{A0A1A3} ±0.42 \\
\cline{1-7} \cline{2-7}
\multirow[c]{10}{*}{$s_{\text{IRW}}$} & \multirow[c]{5}{*}{$\operatorname{Agg}$} & $s_M$ & 0.81 & \color[HTML]{A0A1A3} ±0.18 & 0.50 & \color[HTML]{A0A1A3} ±0.38 \\
 &  & $s_{C}$ & \underline{0.89} & \color[HTML]{A0A1A3} ±0.17 & \underline{0.28} & \color[HTML]{A0A1A3} ±0.36 \\
 &  & $s_{IRW}$ & 0.82 & \color[HTML]{A0A1A3} ±0.19 & 0.43 & \color[HTML]{A0A1A3} ±0.39 \\
 &  & \texttt{IF} & 0.89 & \color[HTML]{A0A1A3} ±0.15 & 0.34 & \color[HTML]{A0A1A3} ±0.36 \\
 &  & \texttt{LOF} & 0.82 & \color[HTML]{A0A1A3} ±0.15 & 0.54 & \color[HTML]{A0A1A3} ±0.35 \\
\cline{2-7}
 & \multirow[c]{3}{*}{$\operatorname{Agg}_{\emptyset}$} & Mean & \underline{0.84} & \color[HTML]{A0A1A3} ±0.18 & \underline{0.47} & \color[HTML]{A0A1A3} ±0.40 \\
 &  & Median & 0.82 & \color[HTML]{A0A1A3} ±0.18 & 0.50 & \color[HTML]{A0A1A3} ±0.39 \\
 &  & \texttt{PW} & 0.74 & \color[HTML]{A0A1A3} ±0.17 & 0.64 & \color[HTML]{A0A1A3} ±0.34 \\
\cline{2-7}
 & \multirow[c]{2}{*}{Bas.} & Last layer & 0.66 & \color[HTML]{A0A1A3} ±0.14 & 0.79 & \color[HTML]{A0A1A3} ±0.21 \\
 &  & Logits & \underline{0.73} & \color[HTML]{A0A1A3} ±0.16 & \underline{0.64} & \color[HTML]{A0A1A3} ±0.28 \\
\cline{1-7} \cline{2-7}
MSP & Bas. & MSP & \underline{0.83} & \color[HTML]{A0A1A3} ±0.17 & \underline{0.39} & \color[HTML]{A0A1A3} ±0.28 \\
\cline{1-7} \cline{2-7}
\bottomrule
\end{tabular}

}

\caption{Average performance of each considered metric over all the OOD pairs and model architectures in terms of {\AUROC}, {\ERR}, and {\FPR}. For each common OOD score, we report the results obtained using every aggregation method or choice of features to consider. The best method overall is highlighted in bold and the best methods per underlying metric and setting are underlined.}
\label{tab:overall_average_app}
\vspace{-0.5cm}
\end{table}

\noindent \textbf{Impact of data-driven aggregation.} In almost all scenarios, aggregating the score using a data-driven anomaly detection method leads to a significant gain in performance compared to baseline methods. This supports our claim that useful information is scattered across the layers currently ignored by most methods. \textbf{We show that this information can be retrieved and effectively leveraged to improve OOD detection.}

\subsection{Post Aggregation Is More Stable Across Task, Language, Model Architecture}
\begin{figure}
    \centering
    \includegraphics[width=0.44\textwidth, trim={0 0.5cm 0cm 0cm}]{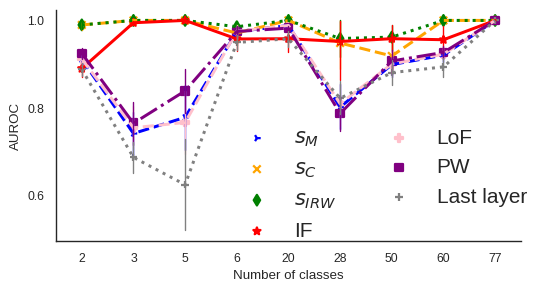}
    \caption{Average performance of OOD detectors in terms of {\AUROC} for tasks involving different numbers of classes.}
    \label{fig:per_nclasses_perfs}
\end{figure}

Most OOD scores have been crafted and finetuned for specific settings. In the case of NLP, they have usually been validated only on datasets involving a small number of classes or on English tasks. In this section, we study the stability and consistency of the performance of each score and aggregation method in different settings.

\noindent\textbf{Stability of performance across tasks.} In Fig.~\ref{fig:per_nclasses_perfs}, we plot the average {\AUROC} across our models and datasets per number of classes of the IN dataset. It is, therefore, the number of classes output by the model. Our best post-aggregation methods (\textit{i.e.,} Maximum cosine similarity and Integrated Rank-Weigthed) produced more consistent results across all settings. It can maintain excellent performance for all types of datasets, whereas the performance of baselines and other aggregation methods tends to fluctuate from one setting to another. \textit{More generally, we observe that data-driven aggregation methods tend to perform consistently on all tasks, whereas previous baselines' performance tends to vary.}

\noindent\textbf{Features aggregation vs. OOD scores aggregation.} Interestingly, we show that while Power Means pre-aggregation of the features yields better results than single-layer scores, they still follow the same trend, and the gain is more minor and inconsistent.

\begin{figure}
     \centering
      \includegraphics[width=0.43\textwidth, trim={0 0.5cm 0cm 0cm}]{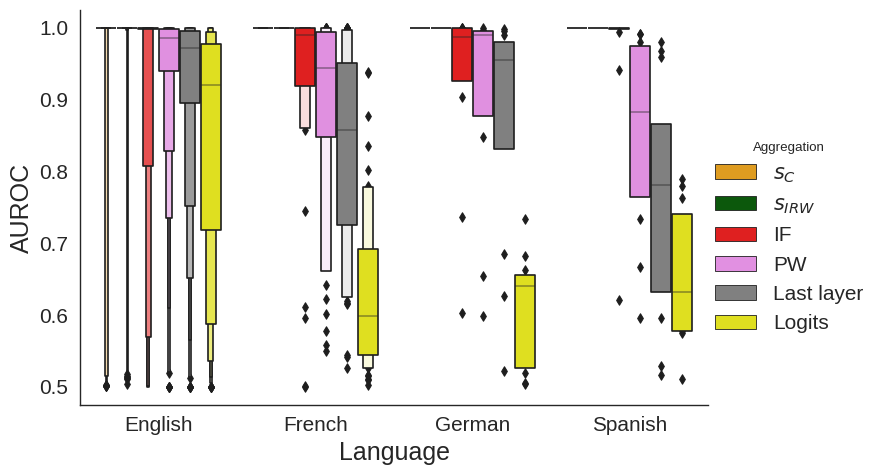}
        \caption{Stability and robustness comparison of the best-performing aggregation methods and underlying OOD scores with $S_C$ as underlying OOD score. Common baselines and SOTA display significant deviations in performance with the different languages, whereas score aggregation methods induce more consistent and better performance.}
        \label{fig:robustness_across_languages}
\end{figure}

\noindent \textbf{Stability of results across languages.} In Fig.~\ref{fig:robustness_across_languages} we present the deviations in performance of the different OOD detection methods and show that our methods are significantly more robust across languages and tasks than baselines and previous SOTA.

\begin{figure}
    \centering
    \includegraphics[width=0.43\textwidth, trim={0 1cm 0cm 0cm}]{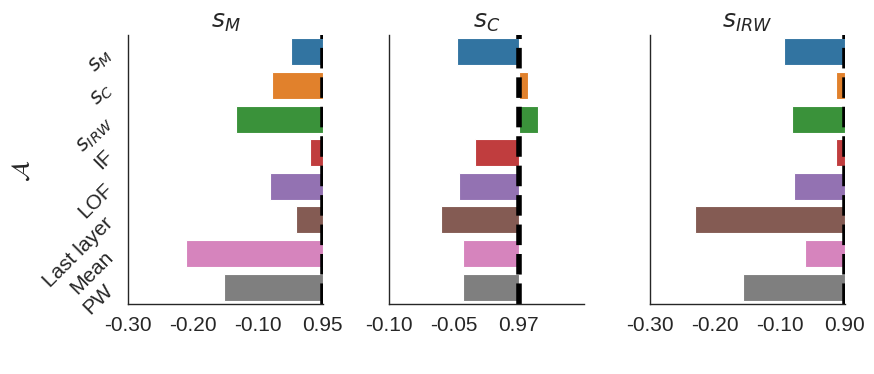}
    \caption{Average performance difference in terms of AUROC between aggregation methods and the oracle (best possible layer).\vspace{-0.5cm}}
    \label{fig:difference_with_best}
\end{figure}

\noindent \textbf{Comparison with the oracle.} As shown in Fig.~\ref{fig:perfs_per_layer}, there often exists a layer that yields very high OOD detection performance. An oracle that knows which layer to consider would perform better than all of the baselines and SOTA, as they know the best layer. In Fig.~\ref{fig:difference_with_best}, we show that our aggregation methods can outperform that oracle (Green bar, $S_C$ scores aggregated with $S_{\text{IRW}}$), whereas SOTA and baselines yield significantly worse results.

\section{Conclusions}

{
We proposed aggregating OOD scores across all the layers of the encoder of a text classifier instead of relying on scores computed on a single hand-picked layer (logits) to improve OOD detection. We confirmed that all the layers are not equal regarding OOD detection and, more importantly, that the common choices for OOD detection (logits) are often not the best choice. We validated our methods on an extended text OOD classification benchmark {\BENCHMARK} we introduced. We showed that our aggregation methods are not only able to outperform previous baselines and recent work, but they were also able to outperform an oracle that would be able to choose the best layer to perform OOD detection for a given task. This leads us to conclude that valuable information for OOD detection is scattered across all the encoder layers. While this tool shows promising results, it should not be trusted blindly. It relies on anomaly detection with respect to the training set and thus can remove otherwise well-handled samples. It is also not infallible and can miss OOD samples, which can cause harm to the model's performance.
}

\clearpage
\section{Acknowledgements}
We thank our reviewers for their insights and useful advice during the different reviewing processes. This work was performed using HPC resources from GENCI–IDRIS (Grant 2022-AD011013945). This work was supprted by HPC resources of CINES and GENCI. The authors would like to thank the staff
of CINES for technical support in managing the Adastra GPU cluster, in particular; Jean-Christophe Penalva,
Johanne Charpentier, Mathieu Cloirec, Jerome Castaings, Gérard Vernou, Bertrand Cirou and José Ricardo
Kouakou.
\bibliography{tacl2021}

\newpage
\appendix
\onecolumn

\section{Benchmark details}

\subsection{Datasets references}

\begin{table}[htbp]
\centering
\caption{List of Datasets}
\label{tab:datasets}
\resizebox{\textwidth}{!}{
\begin{tabular}{|l|l|}
\hline
\textbf{Dataset Name} & \textbf{URL} \\ \hline
SetFit/emotion & https://huggingface.co/datasets/SetFit/emotion \\ \hline
banking77 \cite{Casanueva2020} & https://huggingface.co/datasets/banking77 \\ \hline
SetFit/20-newsgroups\cite{newsgroup} & https://huggingface.co/datasets/SetFit/20\_newsgroups \\ \hline
imdb \cite{maas-etal-2011-learning} & https://huggingface.co/datasets/imdb \\ \hline
AmazonScience/massive \cite{fitzgerald2022massive} & https://huggingface.co/datasets/AmazonScience/massive \\ \hline
glue \cite{wang2018glue} & https://huggingface.co/datasets/glue \\ \hline
super-glue \cite{wang2019superglue} & https://huggingface.co/datasets/super\_glue \\ \hline
sst2 \cite{socher-etal-2013-recursive} & https://huggingface.co/datasets/sst2 \\ \hline
trec \cite{hovy-etal-2001-toward} & https://huggingface.co/datasets/trec \\ \hline
go-emotions \cite{goemotions} & https://huggingface.co/datasets/go\_emotions \\ \hline
clinc-oos \cite{larson2019evaluation} & https://huggingface.co/datasets/clinc\_oos \\ \hline
HINT3 \cite{hint3} & https://github.com/hellohaptik/HINT3 \\ \hline
xquad \cite{Artetxe:etal:2019} & https://huggingface.co/datasets/xquad \\ \hline
catalonia-independence \cite{} & https://huggingface.co/datasets/catalonia\_independence \\ \hline
muchocine \cite{} & https://huggingface.co/datasets/muchocine \\ \hline
allocine \cite{allocinedataset} & https://huggingface.co/datasets/allocine \\ \hline
paws-x \cite{pawsx2019emnlp} & https://huggingface.co/datasets/paws-x \\ \hline
swiss-judgment-prediction \cite{niklaus2022empirical} & https://huggingface.co/datasets/rcds/swiss\_judgment\_prediction \\ \hline
xnli \cite{conneau2018xnli} & https://huggingface.co/datasets/xnli/ \\ \hline
tweet-sentiment-multilingual \cite{barbieri-etal-2022-xlm} & https://huggingface.co/datasets/cardiffnlp/tweet\_sentiment\_multilingual \\ \hline
xstance\cite{vamvas2020x} & https://huggingface.co/datasets/x\_stance \\
\hline
\end{tabular}}
\end{table}

\subsection{OOD pairs}
\label{sec:ood_pairs}

\begin{table}[H]
    \centering

    \resizebox{0.8\textwidth}{!}{

\begin{tabular}{lll}
\toprule
 &  & OUT-DS \\
Language & IN-DS & \\
\midrule
\multirow[c]{10}{*}{English} & 20ng & go-emotions,sst2,imdb,trec,mnli,snli,rte,b77,massive,trec-fine,emotion \\
 & b77 & go-emotions,sst2,imdb,20ng,trec,mnli,snli,rte,massive,trec-fine,emotion \\
 & emotion & go-emotions,sst2,imdb,20ng,trec,mnli,snli,rte,b77,massive,trec-fine \\
 & go-emotions & sst2,imdb,20ng,trec,mnli,snli,rte,b77,massive,trec-fine,emotion \\
 & imdb & go-emotions,sst2,20ng,trec,mnli,snli,rte,b77,massive,trec-fine,emotion \\
 & massive & go-emotions,sst2,imdb,20ng,trec,mnli,snli,rte,b77,trec-fine,emotion \\
 & rte & go-emotions,sst2,imdb,20ng,trec,mnli,snli,b77,massive,trec-fine,emotion \\
 & sst2 & go-emotions,imdb,20ng,trec,mnli,snli,rte,b77,massive,trec-fine,emotion \\
 & trec & go-emotions,sst2,imdb,20ng,mnli,snli,rte,b77,massive,trec-fine,emotion \\
 & trec-fine & go-emotions,sst2,imdb,20ng,trec,mnli,snli,rte,b77,massive,emotion \\
\cline{1-3}
\multirow[c]{7}{*}{French} & fr-allocine & fr-cls,fr-xnli,fr-pawsx,fr-xstance,fr-swiss-judgement,fr-tweet-sentiment \\
 & fr-cls & fr-xnli,fr-pawsx,fr-allocine,fr-xstance,fr-swiss-judgement,fr-tweet-sentiment \\
 & fr-pawsx & fr-cls,fr-xnli,fr-allocine,fr-xstance,fr-swiss-judgement,fr-tweet-sentiment \\
 & fr-swiss-judgement & fr-cls,fr-xnli,fr-pawsx,fr-allocine,fr-xstance,fr-tweet-sentiment \\
 & fr-tweet-sentiment & fr-cls,fr-xnli,fr-pawsx,fr-allocine,fr-xstance,fr-swiss-judgement \\
 & fr-xnli & fr-cls,fr-pawsx,fr-allocine,fr-xstance,fr-swiss-judgement,fr-tweet-sentiment \\
 & fr-xstance & fr-cls,fr-xnli,fr-pawsx,fr-allocine,fr-swiss-judgement,fr-tweet-sentiment \\
\cline{1-3}
\multirow[c]{4}{*}{German} & de-pawsx & de-xstance,de-swiss-judgement,de-tweet-sentiment \\
 & de-swiss-judgement & de-xstance,de-tweet-sentiment,de-pawsx \\
 & de-tweet-sentiment & de-xstance,de-swiss-judgement,de-pawsx \\
 & de-xstance & de-swiss-judgement,de-tweet-sentiment,de-pawsx \\
\cline{1-3}
\multirow[c]{4}{*}{Spanish} & es-cine & es-tweet-sentiment,es-pawsx,es-tweet-inde \\
 & es-pawsx & es-tweet-sentiment,es-cine,es-tweet-inde \\
 & es-tweet-inde & es-tweet-sentiment,es-pawsx,es-cine \\
 & es-tweet-sentiment & es-pawsx,es-cine,es-tweet-inde \\
\cline{1-3}
\bottomrule
\end{tabular}

    }

    \caption{List of OOD pairs considered in our benchmark for each language., the name reported here in this table are the names of the datasets on the Huggingface datasets hub.}
    \label{tab:pair_list}
\end{table}
\begin{table}[H]
    \centering
    \resizebox{0.3\textwidth}{!}{
\begin{tabular}{llr}
\toprule
 & Dataset & Number of classes \\
Language &  &  \\
\midrule
\multirow[c]{13}{*}{English} & go-emotions & 28 \\
 & sst2 & 2 \\
 & imdb & 2 \\
 & 20ng & 20 \\
 & b77 & 77 \\
 & massive & 60 \\
 & trec-fine & 50 \\
 & emotion & 6 \\
 & trec & 6 \\
 & rte & 2 \\
 & mnli & 3 \\
 & snli & 3 \\
\cline{1-3}
\multirow[c]{7}{*}{French} & cls & 2 \\
 & xnli & 3 \\
 & pawsx & 2 \\
 & allocine & 2 \\
 & xstance & 2 \\
 & swiss-judgement & 2 \\
 & tweet-sentiment & 3 \\
\cline{1-3}
\multirow[c]{4}{*}{German} & xstance & 2 \\
 & swiss-judgement & 2 \\
 & tweet-sentiment & 3 \\
 & pawsx & 2 \\
\cline{1-3}
\multirow[c]{4}{*}{Spanish} & tweet-sentiment & 3 \\
 & pawsx & 2 \\
 & cine & 5 \\
 & tweet-inde & 3 \\
\bottomrule
\end{tabular}}

    \caption{Details of the datasets composing {\BENCHMARK}.}
    \label{tab:classes_per_datasets}
\end{table}

\subsection{OOD detection performance metrics}
\label{sec:perfs_metrics_app}
For evaluation we follow previous work in anomaly detection and use {\AUROC}, {\FPR}, {\AUPRIN}/{\AUPROUT} and {\ERR}. 

\noindent \textbf{Area Under the Receiver Operating Characteristic curve ({\AUROC};).} The Receiver Operating Characteristic curve is curve obtained by plotting the True positive rate against the False positive rate. The area under this curve is the probability that an in-distribution example $\mathbf{X}_{in}$ has a anomaly score higher than an OOD sample $\xbold_{out}$: {\AUROC}$ = \Pr(s(\xbold_{in}) > s(\xbold_{out}))$. It is given by $\gamma \mapsto (\Pr\big( s(\xbold) > \gamma \, | \, Z=0\big) , \Pr\big( s(\xbold) > \gamma \, | \, Z=1\big))$.

\noindent  \textbf{False Positive Rate at $95 \%$ True Positive Rate ({\FPR})}. We accept to allow only a given false positive rate $r$ corresponding to a defined level of safety and we want to know what share of positive samples we actually catch under this constraint. It leads to select a threshold $\gamma_r$ such that the corresponding TPR equals $r$. At this threshold, one then computes:
        $\Pr( s(\xbold) > \gamma_r \, | \, Z=0)$
    with $\gamma_r$ s.t. $\textrm{TPR}(\gamma_r)=r$.
    $r$ is chosen depending on the difficulty of the task at hand and the required level of safety. 

\noindent \textbf{Area Under the Precision-Recall curve ({\AUPRIN}/{\AUPROUT})}. The Precision-Recall curve plots the recall (true detection rate) against the precision (actual proportion of OOD amongst the predicted OOD). The area under this curve captures the trade-off between precision and recall made by the model. A high value represents a high precision and a high recall \textit{i.e.} the detector captures most of the positive samples while having few false positives.

\noindent \textbf{Detection error ({\ERR})}. It is simply the probability of miss-classification for the best threshold.

\section{Layer importance in OOD detection}

Our extensive experiments consistently show that there almost always exists a layer that is excellent at separating OOD data. 

\begin{table}[]
    \centering
    \resizebox{0.30\textwidth}{!}{
\begin{tabular}{llrr}
\toprule
 &  & \multicolumn{2}{c}{\AUROC} \\
Ours & Aggregation &  &  \\
\midrule
\multirow[c]{5}{*}{$\operatorname{Agg}$} & $s_M$ & 0.88 & ±0.16 \\
 & $s_{C}$ & 0.91 & ±0.16 \\
 & $s_{IRW}$ & 0.85 & ±0.19 \\
 & IF & 0.92 & ±0.13 \\
 & LOF & 0.87 & ±0.15 \\
\cline{1-4}
\multirow[c]{2}{*}{$\operatorname{Agg}_{\emptyset}$} & Mean & 0.84 & ±0.18 \\
 & PW & 0.82 & ±0.17 \\
\cline{1-4}
Oracle & Oracle & 0.94 & ±0.12 \\
\cline{1-4}
\bottomrule
\end{tabular}
}

    \caption{Average performance of aggregation scores along with the performance of the oracle that is able to select the best layer to perform OOD detection.}
    \label{tab:oracle_comparison}
\end{table}

In Table~\ref{tab:overal_average_with_oracle} we present the {\AUROC} performance of all aggregation with all metrics in comparison with the oracle's performance. We see that there is much room for improvement and that we are able to extract these improvements and even go further.

\begin{table}[H] \centering
\resizebox{0.3\textwidth}{!}{

\begin{tabular}{lllrr}
\toprule
 &  &  & \multicolumn{2}{c}{\AUROC} \\
Metric & Ours & Aggregation &  &  \\
\midrule
\multirow[c]{8}{*}{$s_M$} & \multirow[c]{5}{*}{$\operatorname{Agg}$} & $s_{IRW}$ & 0.82 & ±0.20 \\
 &  & LOF & 0.87 & ±0.15 \\
 &  & $s_{C}$ & 0.88 & ±0.17 \\
 &  & $s_M$ & 0.91 & ±0.14 \\
 &  & IF & 0.94 & ±0.10 \\
\cline{2-5}
 & \multirow[c]{2}{*}{$\operatorname{Agg}_{\emptyset}$} & Mean & 0.74 & ±0.18 \\
 &  & PW & 0.80 & ±0.17 \\
\cline{2-5}
 & Oracle & Oracle & 0.95 & ±0.09 \\
\cline{1-5} \cline{2-5}
\multirow[c]{8}{*}{$s_{C}$} & \multirow[c]{5}{*}{$\operatorname{Agg}$} & $s_M$ & 0.93 & ±0.11 \\
 &  & LOF & 0.93 & ±0.11 \\
 &  & IF & 0.94 & ±0.14 \\
 &  & $s_{C}$ & 0.98 & ±0.09 \\
 &  & $s_{IRW}$ & 0.99 & ±0.07 \\
\cline{2-5}
 & \multirow[c]{2}{*}{$\operatorname{Agg}_{\emptyset}$} & PW & 0.93 & ±0.11 \\
 &  & Mean & 0.93 & ±0.11 \\
\cline{2-5}
 & Oracle & Oracle & 0.97 & ±0.07 \\
\cline{1-5} \cline{2-5}
\multirow[c]{8}{*}{$s_{IRW}$} & \multirow[c]{5}{*}{$\operatorname{Agg}$} & $s_M$ & 0.81 & ±0.18 \\
 &  & $s_{IRW}$ & 0.81 & ±0.19 \\
 &  & LOF & 0.82 & ±0.15 \\
 &  & $s_{C}$ & 0.88 & ±0.18 \\
 &  & IF & 0.89 & ±0.15 \\
\cline{2-5}
 & \multirow[c]{2}{*}{$\operatorname{Agg}_{\emptyset}$} & PW & 0.74 & ±0.17 \\
 &  & Mean & 0.84 & ±0.18 \\
\cline{2-5}
 & Oracle & Oracle & 0.90 & ±0.16 \\
\cline{1-5} \cline{2-5}
\bottomrule
\end{tabular}

}
\caption{Performance per metrics and aggregation with the oracle aggregation for comparison}
\label{tab:overal_average_with_oracle}
\end{table}

\subsection{Explainability and layer significance}

\noindent \textbf{Best Layer Selection (Oracle).} In Fig~\ref{fig:perfs_per_layer}, we showed that high OOD detection performance could be reached, provided that we know which is the best layer to perform the OOD detection on. We compare our aggregation methods to an oracle method that always uses the best layer. We show in Fig.~\ref{fig:difference_with_best} that our aggregation's methods outperform baselines and, in some cases, the performance of the oracle. This means that \textbf{our aggregation methods reach and even outperform oracle performance.} 

\begin{figure}
    \centering
    \includegraphics[width=0.48\textwidth]{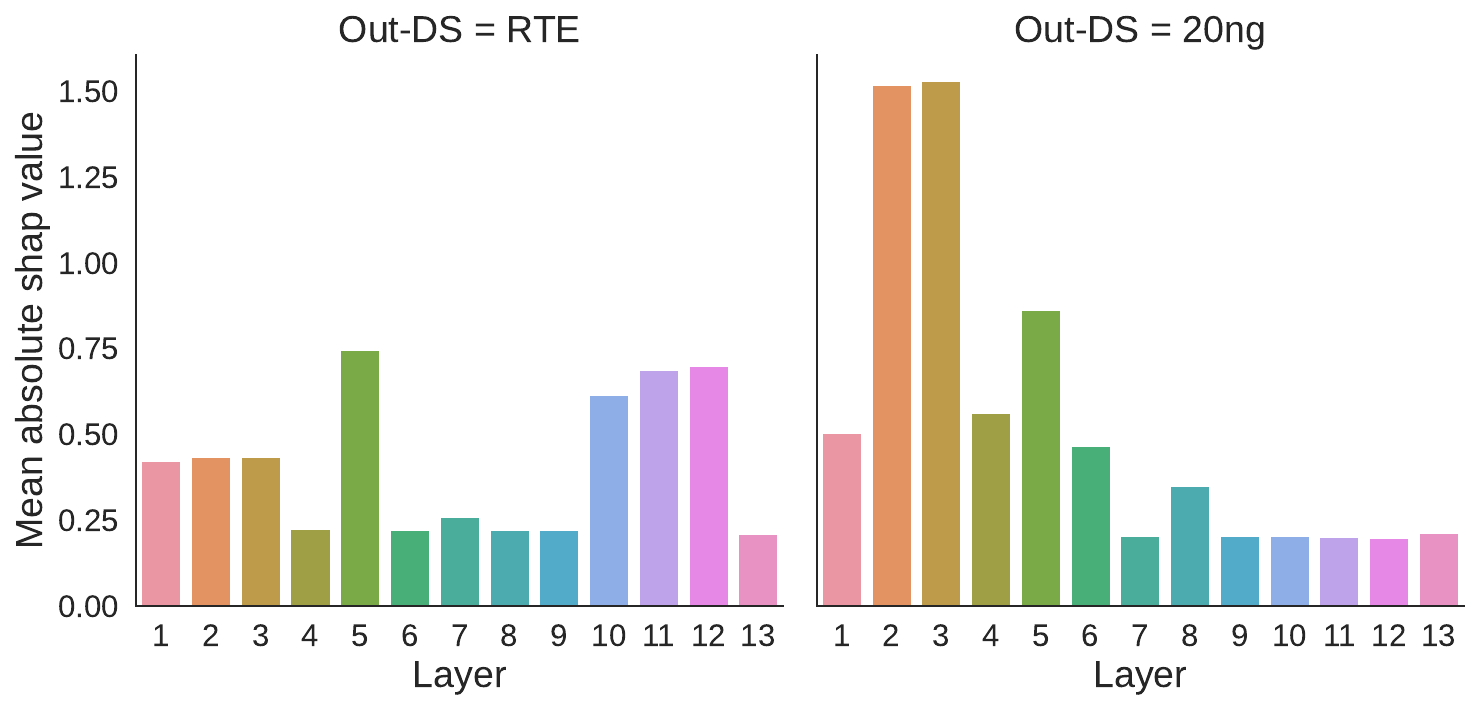}
         \caption{Importance (in terms of Shap score of each layer in OOD detection for Go-Emotions as IN-DS with BERT for two different OOD datasets. It represents the importance of each feature (Mahalanobis distance computed at a given layer) for the isolation forest anomaly detection algorithms. We can see that the significant layers are different from one OOD dataset to another and that in both presented cases, the last layer ($12$) nor the logits ($13$) are not the most useful ones.}
        \label{fig:feature_importance}
\end{figure}

\noindent \textbf{Retro-engineering and explainability.} We propose an explainability analysis of the learned aggregation algorithms to gain more insights into the layer selection retained by our data-driven detectors. We report in Fig.~\ref{fig:feature_importance} the SHAP scores of one of them to distinguish Go-Emotions samples from RTE samples and 20 news-group samples. It outlines the different importance of the layers for different tasks. Not surprisingly, we found that different layers better separate different classes and tasks. We also confirm that the last layer is not always the best suited for OOD data separation.

\section{Additional aggregation algorithms}
\label{sec:additional_algorithms}

We proposed in our work a framework that accommodates a wide range of aggregation algorithms. We focused on unsupervised anomaly detection algorithms and common statistics. However, many others options are available in different flavors. For instance, we focused on per-class anomaly detection but we can redo all our work for all classes at once or implement common OOD scores as aggregation mechanisms. In this section, we propose additional aggregation algorithms that are worth exploring. We provide their formalization and results when available and propose direction for future works.

\subsection{Details of anomaly detection algorithms}

\noindent \textbf{Common OOD scores for score aggregation} We propose to first aggregate the OOD scores using common detectors at a meta level. We apply the Mahalanobis distance ($S_M$), the Maximum Cosine Similarity ($S_C$) and the data depth ($S_{\text{IRW}}$) to the scores trajectory through the model. We gather the scores at each layer for a reference set and then compute these scores with the trajectory scores as features instead of the internal representation of the model.

\noindent\textit{Local Outlier Factor} It measures the density of objects around a sample to decide whether it is an inlier or an outlier. It relies on the $k$-distance of a sample, \textit{e.g,} its distance to its $k$th closest neighbour and considers $N_k(\xbold)$ the set of the $k$ nearest neighbours. 

For stability reasons, the usual LOF method uses the reachability distance\footnote{which is not a proper mathematical distance since it is not symmetric} which is defined as $\operatorname{Rd}(\xbold, \xbold') = \max (k-\operatorname{distance}(\xbold'), d(\xbold, \xbold'))$. Intuitively, it is the distance between $\xbold$ and $\xbold'$ if they are far enough from each other otherwise they are in the same $k$ nearest neighbors set and in that case the diameter of the set is used as minimal distance. From these definitions, we can define the local density of a sample $\xbold$, $\operatorname{dens}(\xbold) = \left( \frac{\sum_{\xbold' \in N_k(\xbold)} \operatorname{Rd}_k(\xbold, \xbold')}{\card{N_k(\xbold)}} \right)$. The LOF score compare the density of a sample to the densities of its neighbor: $\LOF(\xbold)= \frac{\sum_{\xbold' \in N_k(\xbold)} \operatorname{dens}_k(\xbold')}{\card{N_k(\xbold)}\operatorname{dens}_k(\xbold)}$. If $\LOF(\xbold) = 1$ the sample has the same density as his neighbor, if it's lower than $1$ it has a higher density than its neighbor and thus is likely to be an inlier whereas if the score is higher than $1$ it has a smaller density than its neighbor and should be an outlier.

\noindent\textit{Isolation Forest} This popular algorithm is built on the idea that abnormal instances should be easier to distinguish from normal instances. It leads to a score that measures the complexity of separating samples based on the number of necessary decision trees required to isolate a data point. In addition to its computational ease, it benefits from stable hyper-parameters and is well suited to the unsupervised setting.
Formally speaking we consider a recursive partitioning problem which is formalized as an isolation tree structure. The number of trees required to isolate a point is the length of a path through the tree from the root to a leaf.

\subsection{Per-class scoring vs. Global aggregation}

We reproduced the common strategy of relying on a per-class OOD score and then using the minimum score over the classes as the OOD score. This strategy relies on the intuition that an IN sample, belonging to a given class will at least have a small anomaly score regarding this class whereas an OOD sample would have only high scores.

However, using our aggregation tools we can imagine relying on per-class OOD scores for the underlying scores but including them in the aggregation mechanism. For comprehensiveness' sake, we report here the results under this setting but we found that in most cases per-class scores remain the better solution.

Following our notations and framework, it means formally that the aggregation algorithms are now not indexed by the classes but take as input a vector containing all the scores per layer and per class:

\begin{align}
    \mathcal{A}: &(\Rrond^L \times \Rrond^C) \times (\Rrond^L \times \Rrond^C)^{N}  \rightarrow \mathbb{R} \nonumber\\
&\mathbf{s} \times \mathcal{R} \mapsto  \mathcal{A}(\mathbf{s}, \mathcal{R}),
\end{align} 

\begin{table}[H] \centering
\resizebox{0.5\textwidth}{!}{
\begin{tabular}{llrrrrrrrrrr}
\toprule
 &  & \multicolumn{2}{c}{\AUROC} & \multicolumn{2}{c}{\FPR} & \multicolumn{2}{c}{\ERR} &   \\
Ours & Aggregation &  &  &  &  &  &  &    \\
\midrule
\multirow[c]{5}{*}{$\operatorname{Agg}$} & $s_C*$ & 0.55 & \color[HTML]{A0A1A3} ±0.09 & 0.92 & \color[HTML]{A0A1A3} ±0.14 & 0.48 & \color[HTML]{A0A1A3} ±0.27 \\
 & $s_M*$ & 0.81 & \color[HTML]{A0A1A3} ±0.18 & 0.45 & \color[HTML]{A0A1A3} ±0.40 & \underline{\bfseries 0.22} & \color[HTML]{A0A1A3} ±0.23 \\
 & $s_{IRW}*$ & 0.79 & \color[HTML]{A0A1A3} ±0.19 & 0.55 & \color[HTML]{A0A1A3} ±0.39 & 0.29 & \color[HTML]{A0A1A3} ±0.26 \\
 & \texttt{IF}* & \underline{\bfseries 0.83} & \color[HTML]{A0A1A3} ±0.16 & \underline{\bfseries 0.36} & \color[HTML]{A0A1A3} ±0.32 & 0.22 & \color[HTML]{A0A1A3} ±0.22 \\
 & \texttt{LoF}* & 0.50 & \color[HTML]{A0A1A3} ±0.00 & 1.00 & \color[HTML]{A0A1A3} ±0.02 & 0.45 & \color[HTML]{A0A1A3} ±0.30 \\
\cline{1-8}
\bottomrule
\end{tabular}

}
\caption{Overall average performance of each aggregation method overall architectures, tasks and underlying OOD scores when aggregating all layers and classes at once.}
\label{tab:average_per_aggr_no_classes}
\end{table}

\begin{table}
\centering \resizebox{0.45\textwidth}{!}{
\begin{tabular}{lllrrrrrr}
\toprule
 &  & & \multicolumn{2}{c}{\AUROC} & \multicolumn{2}{c}{\FPR} & \multicolumn{2}{c}{\ERR} \\
Metric & Ours & Aggregation &  &  &  &  &  &  \\
\midrule
\multirow[c]{5}{*}{$s_M$} & \multirow[c]{5}{*}{$\operatorname{Agg}$} & $s_C*$ & 0.57 & \color[HTML]{A0A1A3} ±0.12 & 0.89 & \color[HTML]{A0A1A3} ±0.19 & 0.48 & \color[HTML]{A0A1A3} ±0.27 \\
 &  & $s_M*$ & 0.75 & \color[HTML]{A0A1A3} ±0.16 & 0.67 & \color[HTML]{A0A1A3} ±0.35 & \underline{0.33} & \color[HTML]{A0A1A3} ±0.25 \\
 &  & $s_{IRW}*$ & 0.67 & \color[HTML]{A0A1A3} ±0.14 & 0.77 & \color[HTML]{A0A1A3} ±0.27 & 0.38 & \color[HTML]{A0A1A3} ±0.23 \\
 &  & \texttt{IF}* & \underline{0.75} & \color[HTML]{A0A1A3} ±0.15 & \underline{0.55} & \color[HTML]{A0A1A3} ±0.29 & 0.34 & \color[HTML]{A0A1A3} ±0.21 \\
 &  & \texttt{LoF}* & 0.50 & \color[HTML]{A0A1A3} ±0.01 & 0.99 & \color[HTML]{A0A1A3} ±0.03 & 0.42 & \color[HTML]{A0A1A3} ±0.28 \\
\cline{1-9} \cline{2-9}
\multirow[c]{5}{*}{$s_{C}$} & \multirow[c]{5}{*}{$\operatorname{Agg}$} & $s_C*$ & 0.51 & \color[HTML]{A0A1A3} ±0.01 & 0.97 & \color[HTML]{A0A1A3} ±0.02 & 0.50 & \color[HTML]{A0A1A3} ±0.27 \\
 &  & $s_M*$ & 0.75 & \color[HTML]{A0A1A3} ±0.16 & 0.53 & \color[HTML]{A0A1A3} ±0.34 & 0.28 & \color[HTML]{A0A1A3} ±0.21 \\
 &  & $s_{IRW}*$ & 0.82 & \color[HTML]{A0A1A3} ±0.19 & 0.47 & \color[HTML]{A0A1A3} ±0.38 & 0.24 & \color[HTML]{A0A1A3} ±0.24 \\
 &  & \texttt{IF}* & \underline{0.84} & \color[HTML]{A0A1A3} ±0.15 & \underline{0.34} & \color[HTML]{A0A1A3} ±0.32 & \underline{0.23} & \color[HTML]{A0A1A3} ±0.24 \\
 &  & \texttt{LoF}* & 0.50 & \color[HTML]{A0A1A3} ±0.00 & 1.00 & \color[HTML]{A0A1A3} ±0.00 & 0.44 & \color[HTML]{A0A1A3} ±0.29 \\
\cline{1-9} \cline{2-9}
\multirow[c]{5}{*}{$s_{\text{IRW}}$} & \multirow[c]{5}{*}{$\operatorname{Agg}$} & $s_C*$ & 0.58 & \color[HTML]{A0A1A3} ±0.10 & 0.91 & \color[HTML]{A0A1A3} ±0.14 & 0.45 & \color[HTML]{A0A1A3} ±0.26 \\
 &  & $s_M*$ & \underline{\bfseries 0.93} & \color[HTML]{A0A1A3} ±0.15 & \underline{\bfseries 0.16} & \color[HTML]{A0A1A3} ±0.31 & \underline{\bfseries 0.07} & \color[HTML]{A0A1A3} ±0.11 \\
 &  & $s_{IRW}*$ & 0.86 & \color[HTML]{A0A1A3} ±0.18 & 0.43 & \color[HTML]{A0A1A3} ±0.40 & 0.26 & \color[HTML]{A0A1A3} ±0.29 \\
 &  & \texttt{IF}* & 0.91 & \color[HTML]{A0A1A3} ±0.13 & 0.19 & \color[HTML]{A0A1A3} ±0.25 & 0.11 & \color[HTML]{A0A1A3} ±0.15 \\
 &  & \texttt{LoF}* & 0.50 & \color[HTML]{A0A1A3} ±0.00 & 1.00 & \color[HTML]{A0A1A3} ±0.00 & 0.50 & \color[HTML]{A0A1A3} ±0.30 \\
\cline{1-9} \cline{2-9}
\bottomrule
\end{tabular}

} 
\caption{Average performance of the aggregation methods (\textit{i.e.}, for $\operatorname{Agg}$) performed on the whole score matrix at once (per-class and per-layer scores concatenated into a single vector) instead of aggregation each trajectory separately.} \label{tab:average_perf_all_main} \end{table}

\noindent \textbf{Overall aggregation vs per-class aggregation.} Consistently with previous work, our aggregation methods do not perform as well when used to produce directly a single overall score instead of being used class-wise and then taking the minimum score over the classes. In Table~\ref{tab:average_perf_all_main} we report the OOD detection performance of this setting.

\subsection{Without reference statistical baselines}

The simplest way to aggregate OOD scores is to consider statistical aggregation over the layers and the classes. We showed that even basic aggregations such as taking the median score enable significant gains with respect to the last layer baselines.

\section{Computational cost}
\label{sec:computational_costs}

\noindent \textbf{Time complexity.} While the addition of an aggregation method induces obvious additional computational costs they are actually quite limited (they do not slow down the process significantly. They require only to compute the usual OOD scores on each layer (which does not change the asymptotic complexity) and then to perform the inference of common anomaly detection algorithms. For example, Isolation forests are known to have a linear complexity in the number of samples and to be able to perform well and fast with numerous and very high dimensional data.

\begin{table}[h]
\centering

\begin{tabular}{|l|c|}
\hline
Aggregation Method & Time (s) \\
\hline
$s_C$ & $1.40 \times 10^{-4}$ s \\
\hline
IF & $1.14 \times 10^{-3}$ s \\
\hline
$s_{IRW}$ & $5.69 \times 10^{-4}$ s \\
\hline
LOF & $4.95 \times 10^{-4}$ s \\
\hline
$s_M$ & $5.82 \times 10^{-3}$ s \\
\hline
max & $3.84 \times 10^{-5}$ s \\
\hline
mean & $3.32 \times 10^{-5}$ s \\
\hline
median & $6.01 \times 10^{-5}$ s \\
\hline
min & $3.32 \times 10^{-5}$ s \\
\hline
\end{tabular}
\caption{Computational cost overhead induced by the OOD detection methods on top of a classifier for a single sample.}
\end{table}

\noindent \textbf{Memory footprint.} Perhaps most of the overhead is a memory overhead: for underlying OOD scores relying on a reference set we have to store one trained score for each layer. In the case of the Mahalanobis distance, it means storing $L$ covariance matrices instead of one in addition of the trained aggregation algorithm.

\section{Explainability and variability}

Isolation forests are constructed by choosing at random separating plans and thus each run might give different importance to features. We benchmarked the methods over $10$ seeds to alleviate variability and validate our results. It showed that while some features could be permuted the overall trend were consistent: features that are not relevant for a run do not significantly gain in importance. 
\begin{figure*}
     \centering
     \begin{subfigure}[b]{0.45\textwidth}
         \centering
    \includegraphics[width=\textwidth]{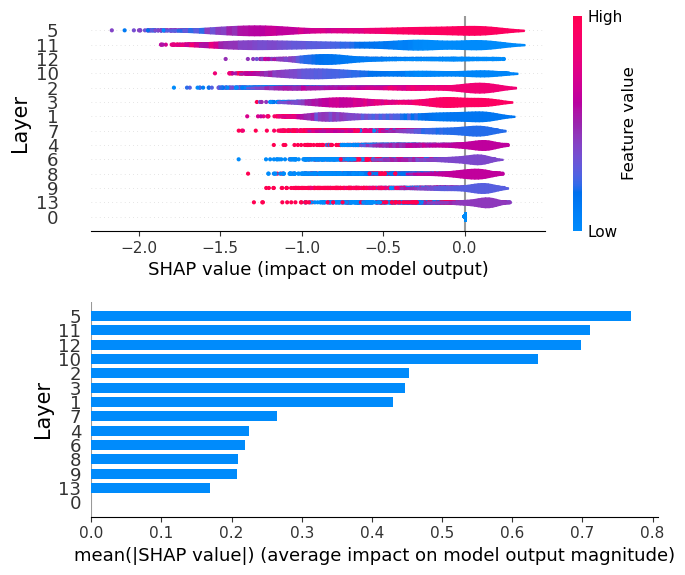}
         \caption{RTE as Out-DS}
     \end{subfigure}
     \begin{subfigure}[b]{0.45\textwidth}
         \centering
         \includegraphics[width=\textwidth]{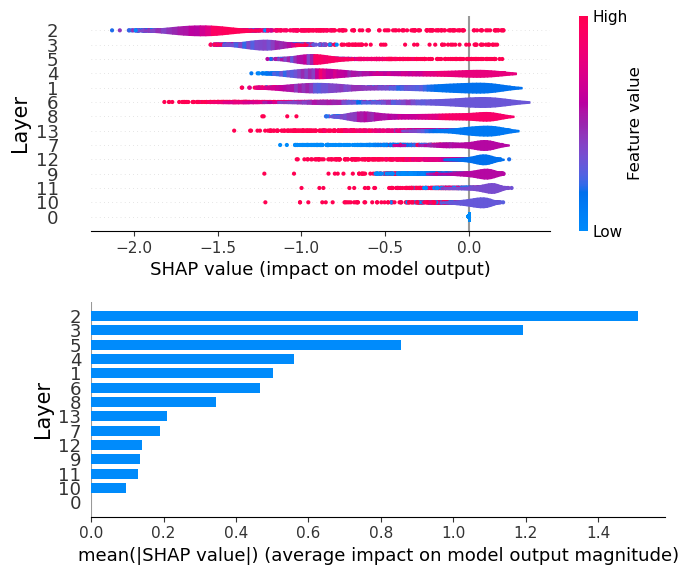}
         \caption{20ng as Out-DS}
     \end{subfigure}
     \caption{Importance (in terms of shap score of each layer in OOD detection for Go-Emotions as IN-DS with a bert architecture for two different OOD datasets. The first row represents the samples from a dataset containing IN and OUT samples. Each line corresponds to a feature and shows how a given feature's value influences the model to rule the sample as an anomaly (negative) or as a normal (positive) sample. The features are ranked by importance in the decision. The second row represents the importance of each feature (OOD score computed at a given layer) for the isolation forest anomaly detection algorithms. We can see that the significant layers are different from one OOD dataset to another and that in both presented cases, the last layer (Layer 12) is not the most useful one.}
        \label{fig:feature_importance_app}
\end{figure*}

\begin{figure}
     \centering
      \includegraphics[width=0.43\textwidth, trim={0 0.5cm 0cm 0cm}]{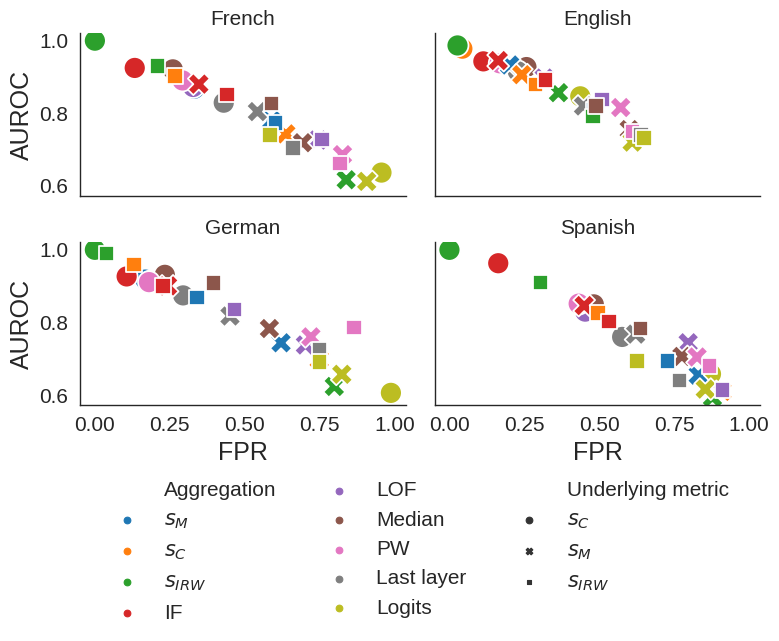}
        \caption{Comparison of the best-performing aggregation methods and underlying OOD scores in terms of {\AUROC} and {\FPR}. We can see that most OOD detectors' performance varies greatly from language to language. The maximum cosine similarity scores aggregated with IRW (green circle) can maintain its performance in all four studied languages and the isolation forests.}
        \label{fig:per_lang_paretto}
\end{figure}

In Fig.~\ref{fig:per_lang_paretto}, we show the relation between {\AUROC} and {\FPR} for all our aggregation methods and underlying metrics for each language we studied. In contrast, the performance of most combinations varies with the language. \textbf{We especially notice that $s_C$ scores aggregated using either the Integrated rank-weighted or $s_C$ consistently achieve excellent performance across all languages.} 

\section{Experimental results}

\subsection{Performance per tasks}

\subsubsection{Language specific results}

In Tables~\ref{tab:average_perf_english}, \ref{tab:average_perf_french}, \ref{tab:average_perf_german}, \ref{tab:average_perf_spanish} we present the average performance of each aggregation methods on each language.

\begin{table}[!ht] \centering \resizebox{0.4\textwidth}{!}{\begin{tabular}{lllrrrrrrrrrr}
\toprule
 &  &  & \multicolumn{2}{c}{\AUROC} & \multicolumn{2}{c}{\FPR} & \multicolumn{2}{c}{\ERR} & \multicolumn{2}{c}{\AUPRIN} & \multicolumn{2}{c}{\AUPROUT} \\
Metric & Ours & Agg. &  &  &  &  &  &  &  &  &  &  \\
\midrule
$E$ & Bas. & $E$ & \bfseries 0.86 & \color[HTML]{A0A1A3} ±0.16 & \bfseries 0.35 & \color[HTML]{A0A1A3} ±0.29 & \bfseries 0.17 & \color[HTML]{A0A1A3} ±0.17 & \bfseries 0.82 & \color[HTML]{A0A1A3} ±0.25 & \bfseries 0.81 & \color[HTML]{A0A1A3} ±0.22 \\
\cline{1-13} \cline{2-13}
\multirow[c]{15}{*}{$s_M$} & \multirow[c]{2}{*}{Bas.} & Last layer & 0.92 & \color[HTML]{A0A1A3} ±0.11 & 0.24 & \color[HTML]{A0A1A3} ±0.30 & 0.12 & \color[HTML]{A0A1A3} ±0.17 & 0.92 & \color[HTML]{A0A1A3} ±0.16 & 0.86 & \color[HTML]{A0A1A3} ±0.20 \\
 &  & Logits & 0.72 & \color[HTML]{A0A1A3} ±0.14 & 0.61 & \color[HTML]{A0A1A3} ±0.26 & 0.30 & \color[HTML]{A0A1A3} ±0.22 & 0.66 & \color[HTML]{A0A1A3} ±0.29 & 0.67 & \color[HTML]{A0A1A3} ±0.27 \\
\cline{2-13}
 & \multirow[c]{5}{*}{$\operatorname{Agg}_{\emptyset}$} & \texttt{PW} & 0.82 & \color[HTML]{A0A1A3} ±0.17 & 0.57 & \color[HTML]{A0A1A3} ±0.40 & 0.26 & \color[HTML]{A0A1A3} ±0.24 & 0.84 & \color[HTML]{A0A1A3} ±0.22 & 0.67 & \color[HTML]{A0A1A3} ±0.30 \\
 &  & Mean & 0.77 & \color[HTML]{A0A1A3} ±0.18 & 0.54 & \color[HTML]{A0A1A3} ±0.40 & 0.30 & \color[HTML]{A0A1A3} ±0.27 & 0.68 & \color[HTML]{A0A1A3} ±0.30 & 0.76 & \color[HTML]{A0A1A3} ±0.28 \\
 &  & Median & 0.76 & \color[HTML]{A0A1A3} ±0.17 & 0.60 & \color[HTML]{A0A1A3} ±0.37 & 0.31 & \color[HTML]{A0A1A3} ±0.25 & 0.70 & \color[HTML]{A0A1A3} ±0.29 & 0.70 & \color[HTML]{A0A1A3} ±0.29 \\
 &  & Max & 0.75 & \color[HTML]{A0A1A3} ±0.19 & 0.55 & \color[HTML]{A0A1A3} ±0.40 & 0.31 & \color[HTML]{A0A1A3} ±0.27 & 0.67 & \color[HTML]{A0A1A3} ±0.30 & 0.75 & \color[HTML]{A0A1A3} ±0.29 \\
 &  & Min & 0.50 & \color[HTML]{A0A1A3} ±0.00 & 1.00 & \color[HTML]{A0A1A3} ±0.00 & 0.41 & \color[HTML]{A0A1A3} ±0.29 & 0.80 & \color[HTML]{A0A1A3} ±0.15 & 0.70 & \color[HTML]{A0A1A3} ±0.15 \\
\cline{2-13}
 & \multirow[c]{8}{*}{$\operatorname{Agg}$} & \texttt{IF} & \bfseries 0.95 & \color[HTML]{A0A1A3} ±0.09 & \bfseries 0.16 & \color[HTML]{A0A1A3} ±0.23 & \bfseries 0.09 & \color[HTML]{A0A1A3} ±0.12 & \bfseries 0.93 & \color[HTML]{A0A1A3} ±0.15 & 0.91 & \color[HTML]{A0A1A3} ±0.16 \\
 &  & $s_M$ & 0.93 & \color[HTML]{A0A1A3} ±0.11 & 0.20 & \color[HTML]{A0A1A3} ±0.29 & 0.12 & \color[HTML]{A0A1A3} ±0.17 & 0.91 & \color[HTML]{A0A1A3} ±0.19 & 0.90 & \color[HTML]{A0A1A3} ±0.17 \\
 &  & $s_{C}$ & 0.91 & \color[HTML]{A0A1A3} ±0.16 & 0.24 & \color[HTML]{A0A1A3} ±0.36 & 0.15 & \color[HTML]{A0A1A3} ±0.23 & 0.89 & \color[HTML]{A0A1A3} ±0.20 & \bfseries 0.93 & \color[HTML]{A0A1A3} ±0.15 \\
 &  & \texttt{LOF} & 0.90 & \color[HTML]{A0A1A3} ±0.14 & 0.32 & \color[HTML]{A0A1A3} ±0.36 & 0.17 & \color[HTML]{A0A1A3} ±0.23 & 0.88 & \color[HTML]{A0A1A3} ±0.20 & 0.85 & \color[HTML]{A0A1A3} ±0.21 \\
 &  & $s_M*$ & 0.78 & \color[HTML]{A0A1A3} ±0.15 & 0.63 & \color[HTML]{A0A1A3} ±0.37 & 0.30 & \color[HTML]{A0A1A3} ±0.25 & 0.75 & \color[HTML]{A0A1A3} ±0.27 & 0.68 & \color[HTML]{A0A1A3} ±0.31 \\
 &  & \texttt{IF}* & 0.78 & \color[HTML]{A0A1A3} ±0.14 & 0.49 & \color[HTML]{A0A1A3} ±0.27 & 0.31 & \color[HTML]{A0A1A3} ±0.20 & 0.55 & \color[HTML]{A0A1A3} ±0.28 & 0.84 & \color[HTML]{A0A1A3} ±0.22 \\
 &  & $s_C*$ & 0.56 & \color[HTML]{A0A1A3} ±0.12 & 0.89 & \color[HTML]{A0A1A3} ±0.19 & 0.49 & \color[HTML]{A0A1A3} ±0.28 & 0.51 & \color[HTML]{A0A1A3} ±0.31 & 0.59 & \color[HTML]{A0A1A3} ±0.29 \\
 &  & \texttt{LoF}* & 0.50 & \color[HTML]{A0A1A3} ±0.01 & 0.99 & \color[HTML]{A0A1A3} ±0.03 & 0.43 & \color[HTML]{A0A1A3} ±0.30 & 0.71 & \color[HTML]{A0A1A3} ±0.23 & 0.69 & \color[HTML]{A0A1A3} ±0.19 \\
\cline{1-13} \cline{2-13}
\multirow[c]{15}{*}{$s_{C}$} & \multirow[c]{2}{*}{Bas.} & Last layer & 0.92 & \color[HTML]{A0A1A3} ±0.11 & 0.22 & \color[HTML]{A0A1A3} ±0.26 & 0.11 & \color[HTML]{A0A1A3} ±0.13 & 0.90 & \color[HTML]{A0A1A3} ±0.18 & 0.88 & \color[HTML]{A0A1A3} ±0.18 \\
 &  & Logits & 0.85 & \color[HTML]{A0A1A3} ±0.16 & 0.44 & \color[HTML]{A0A1A3} ±0.41 & 0.24 & \color[HTML]{A0A1A3} ±0.28 & 0.83 & \color[HTML]{A0A1A3} ±0.24 & 0.81 & \color[HTML]{A0A1A3} ±0.23 \\
\cline{2-13}
 & \multirow[c]{5}{*}{$\operatorname{Agg}_{\emptyset}$} & Mean & 0.94 & \color[HTML]{A0A1A3} ±0.11 & 0.22 & \color[HTML]{A0A1A3} ±0.31 & 0.15 & \color[HTML]{A0A1A3} ±0.20 & 0.91 & \color[HTML]{A0A1A3} ±0.17 & 0.93 & \color[HTML]{A0A1A3} ±0.16 \\
 &  & \texttt{PW} & 0.94 & \color[HTML]{A0A1A3} ±0.11 & 0.17 & \color[HTML]{A0A1A3} ±0.26 & 0.09 & \color[HTML]{A0A1A3} ±0.13 & 0.92 & \color[HTML]{A0A1A3} ±0.16 & 0.91 & \color[HTML]{A0A1A3} ±0.17 \\
 &  & Min & 0.93 & \color[HTML]{A0A1A3} ±0.11 & 0.24 & \color[HTML]{A0A1A3} ±0.31 & 0.15 & \color[HTML]{A0A1A3} ±0.18 & 0.90 & \color[HTML]{A0A1A3} ±0.17 & 0.91 & \color[HTML]{A0A1A3} ±0.18 \\
 &  & Median & 0.93 & \color[HTML]{A0A1A3} ±0.12 & 0.26 & \color[HTML]{A0A1A3} ±0.34 & 0.17 & \color[HTML]{A0A1A3} ±0.23 & 0.88 & \color[HTML]{A0A1A3} ±0.19 & 0.92 & \color[HTML]{A0A1A3} ±0.16 \\
 &  & Max & 0.57 & \color[HTML]{A0A1A3} ±0.15 & 0.90 & \color[HTML]{A0A1A3} ±0.29 & 0.35 & \color[HTML]{A0A1A3} ±0.29 & 0.82 & \color[HTML]{A0A1A3} ±0.18 & 0.71 & \color[HTML]{A0A1A3} ±0.18 \\
\cline{2-13}
 & \multirow[c]{8}{*}{$\operatorname{Agg}$} & $s_{C}$ & \bfseries 0.98 & \color[HTML]{A0A1A3} ±0.10 & \bfseries 0.04 & \color[HTML]{A0A1A3} ±0.20 & 0.03 & \color[HTML]{A0A1A3} ±0.11 & \bfseries 0.98 & \color[HTML]{A0A1A3} ±0.12 & \bfseries 0.98 & \color[HTML]{A0A1A3} ±0.12 \\
 &  & \texttt{IF} & 0.94 & \color[HTML]{A0A1A3} ±0.14 & 0.11 & \color[HTML]{A0A1A3} ±0.29 & \bfseries 0.03 & \color[HTML]{A0A1A3} ±0.11 & 0.95 & \color[HTML]{A0A1A3} ±0.16 & 0.85 & \color[HTML]{A0A1A3} ±0.23 \\
 &  & \texttt{LOF} & 0.94 & \color[HTML]{A0A1A3} ±0.10 & 0.17 & \color[HTML]{A0A1A3} ±0.25 & 0.09 & \color[HTML]{A0A1A3} ±0.12 & 0.92 & \color[HTML]{A0A1A3} ±0.16 & 0.91 & \color[HTML]{A0A1A3} ±0.17 \\
 &  & $s_M$ & 0.94 & \color[HTML]{A0A1A3} ±0.10 & 0.18 & \color[HTML]{A0A1A3} ±0.25 & 0.10 & \color[HTML]{A0A1A3} ±0.13 & 0.92 & \color[HTML]{A0A1A3} ±0.16 & 0.90 & \color[HTML]{A0A1A3} ±0.17 \\
 &  & \texttt{IF}* & 0.88 & \color[HTML]{A0A1A3} ±0.12 & 0.25 & \color[HTML]{A0A1A3} ±0.26 & 0.18 & \color[HTML]{A0A1A3} ±0.21 & 0.67 & \color[HTML]{A0A1A3} ±0.27 & 0.93 & \color[HTML]{A0A1A3} ±0.11 \\
 &  & $s_M*$ & 0.78 & \color[HTML]{A0A1A3} ±0.16 & 0.47 & \color[HTML]{A0A1A3} ±0.33 & 0.25 & \color[HTML]{A0A1A3} ±0.20 & 0.64 & \color[HTML]{A0A1A3} ±0.28 & 0.78 & \color[HTML]{A0A1A3} ±0.27 \\
 &  & $s_C*$ & 0.51 & \color[HTML]{A0A1A3} ±0.01 & 0.97 & \color[HTML]{A0A1A3} ±0.02 & 0.48 & \color[HTML]{A0A1A3} ±0.28 & 0.53 & \color[HTML]{A0A1A3} ±0.30 & 0.49 & \color[HTML]{A0A1A3} ±0.30 \\
 &  & \texttt{LoF}* & 0.50 & \color[HTML]{A0A1A3} ±0.00 & 1.00 & \color[HTML]{A0A1A3} ±0.00 & 0.40 & \color[HTML]{A0A1A3} ±0.29 & 0.80 & \color[HTML]{A0A1A3} ±0.15 & 0.70 & \color[HTML]{A0A1A3} ±0.15 \\
\cline{1-13} \cline{2-13}
\multirow[c]{15}{*}{$s_{\text{IRW}}$} & \multirow[c]{2}{*}{Bas.} & Logits & 0.73 & \color[HTML]{A0A1A3} ±0.17 & 0.65 & \color[HTML]{A0A1A3} ±0.30 & 0.28 & \color[HTML]{A0A1A3} ±0.21 & 0.75 & \color[HTML]{A0A1A3} ±0.25 & 0.68 & \color[HTML]{A0A1A3} ±0.26 \\
 &  & Last layer & 0.66 & \color[HTML]{A0A1A3} ±0.13 & 0.79 & \color[HTML]{A0A1A3} ±0.21 & 0.38 & \color[HTML]{A0A1A3} ±0.24 & 0.65 & \color[HTML]{A0A1A3} ±0.29 & 0.64 & \color[HTML]{A0A1A3} ±0.26 \\
\cline{2-13}
 & \multirow[c]{5}{*}{$\operatorname{Agg}_{\emptyset}$} & Max & 0.84 & \color[HTML]{A0A1A3} ±0.18 & 0.45 & \color[HTML]{A0A1A3} ±0.40 & 0.22 & \color[HTML]{A0A1A3} ±0.22 & 0.85 & \color[HTML]{A0A1A3} ±0.19 & 0.83 & \color[HTML]{A0A1A3} ±0.22 \\
 &  & Mean & 0.84 & \color[HTML]{A0A1A3} ±0.18 & 0.45 & \color[HTML]{A0A1A3} ±0.41 & 0.22 & \color[HTML]{A0A1A3} ±0.23 & 0.85 & \color[HTML]{A0A1A3} ±0.20 & 0.83 & \color[HTML]{A0A1A3} ±0.21 \\
 &  & Median & 0.82 & \color[HTML]{A0A1A3} ±0.18 & 0.49 & \color[HTML]{A0A1A3} ±0.39 & 0.25 & \color[HTML]{A0A1A3} ±0.24 & 0.82 & \color[HTML]{A0A1A3} ±0.22 & 0.82 & \color[HTML]{A0A1A3} ±0.21 \\
 &  & \texttt{PW} & 0.75 & \color[HTML]{A0A1A3} ±0.17 & 0.61 & \color[HTML]{A0A1A3} ±0.35 & 0.29 & \color[HTML]{A0A1A3} ±0.23 & 0.75 & \color[HTML]{A0A1A3} ±0.26 & 0.73 & \color[HTML]{A0A1A3} ±0.25 \\
 &  & Min & 0.50 & \color[HTML]{A0A1A3} ±0.00 & 1.00 & \color[HTML]{A0A1A3} ±0.00 & 0.41 & \color[HTML]{A0A1A3} ±0.29 & 0.80 & \color[HTML]{A0A1A3} ±0.15 & 0.70 & \color[HTML]{A0A1A3} ±0.15 \\
\cline{2-13}
 & \multirow[c]{8}{*}{$\operatorname{Agg}$} & $s_M*$ & \bfseries 0.99 & \color[HTML]{A0A1A3} ±0.02 & \bfseries 0.03 & \color[HTML]{A0A1A3} ±0.05 & \bfseries 0.03 & \color[HTML]{A0A1A3} ±0.03 & \bfseries 0.96 & \color[HTML]{A0A1A3} ±0.09 & \bfseries 0.99 & \color[HTML]{A0A1A3} ±0.04 \\
 &  & \texttt{IF}* & 0.96 & \color[HTML]{A0A1A3} ±0.04 & 0.10 & \color[HTML]{A0A1A3} ±0.10 & 0.06 & \color[HTML]{A0A1A3} ±0.05 & 0.89 & \color[HTML]{A0A1A3} ±0.16 & 0.96 & \color[HTML]{A0A1A3} ±0.07 \\
 &  & \texttt{IF} & 0.89 & \color[HTML]{A0A1A3} ±0.14 & 0.32 & \color[HTML]{A0A1A3} ±0.36 & 0.15 & \color[HTML]{A0A1A3} ±0.18 & 0.89 & \color[HTML]{A0A1A3} ±0.18 & 0.80 & \color[HTML]{A0A1A3} ±0.25 \\
 &  & $s_{C}$ & 0.88 & \color[HTML]{A0A1A3} ±0.18 & 0.29 & \color[HTML]{A0A1A3} ±0.38 & 0.18 & \color[HTML]{A0A1A3} ±0.22 & 0.88 & \color[HTML]{A0A1A3} ±0.19 & 0.93 & \color[HTML]{A0A1A3} ±0.11 \\
 &  & \texttt{LOF} & 0.84 & \color[HTML]{A0A1A3} ±0.15 & 0.51 & \color[HTML]{A0A1A3} ±0.35 & 0.24 & \color[HTML]{A0A1A3} ±0.23 & 0.84 & \color[HTML]{A0A1A3} ±0.23 & 0.71 & \color[HTML]{A0A1A3} ±0.27 \\
 &  & $s_M$ & 0.81 & \color[HTML]{A0A1A3} ±0.18 & 0.48 & \color[HTML]{A0A1A3} ±0.39 & 0.19 & \color[HTML]{A0A1A3} ±0.19 & 0.84 & \color[HTML]{A0A1A3} ±0.21 & 0.75 & \color[HTML]{A0A1A3} ±0.24 \\
 &  & $s_C*$ & 0.60 & \color[HTML]{A0A1A3} ±0.10 & 0.90 & \color[HTML]{A0A1A3} ±0.15 & 0.43 & \color[HTML]{A0A1A3} ±0.26 & 0.60 & \color[HTML]{A0A1A3} ±0.29 & 0.53 & \color[HTML]{A0A1A3} ±0.31 \\
 &  & \texttt{LoF}* & 0.50 & \color[HTML]{A0A1A3} ±0.00 & 1.00 & \color[HTML]{A0A1A3} ±0.00 & 0.51 & \color[HTML]{A0A1A3} ±0.31 & 0.75 & \color[HTML]{A0A1A3} ±0.16 & 0.62 & \color[HTML]{A0A1A3} ±0.28 \\
\cline{1-13} \cline{2-13}
MSP & Bas. & MSP & \bfseries 0.86 & \color[HTML]{A0A1A3} ±0.15 & \bfseries 0.35 & \color[HTML]{A0A1A3} ±0.26 & \bfseries 0.17 & \color[HTML]{A0A1A3} ±0.16 & \bfseries 0.81 & \color[HTML]{A0A1A3} ±0.24 & \bfseries 0.81 & \color[HTML]{A0A1A3} ±0.20 \\
\cline{1-13} \cline{2-13}
\bottomrule
\end{tabular}
} \caption{Average performance on the english benchamrk.} \label{tab:average_perf_english} \end{table}

\begin{table}[!ht] \centering \resizebox{0.4\textwidth}{!}{\begin{tabular}{lllrrrrrrrrrr}
\toprule
 &  &  & \multicolumn{2}{c}{\AUROC} & \multicolumn{2}{c}{\FPR} & \multicolumn{2}{c}{\ERR} & \multicolumn{2}{c}{\AUPRIN} & \multicolumn{2}{c}{\AUPROUT} \\
Metric & Ours & Agg. &  &  &  &  &  &  &  &  &  &  \\
\midrule
$E$ & Bas. & $E$ & \bfseries 0.69 & \color[HTML]{A0A1A3} ±0.16 & \bfseries 0.62 & \color[HTML]{A0A1A3} ±0.25 & \bfseries 0.31 & \color[HTML]{A0A1A3} ±0.19 & \bfseries 0.62 & \color[HTML]{A0A1A3} ±0.26 & \bfseries 0.69 & \color[HTML]{A0A1A3} ±0.27 \\
\cline{1-13} \cline{2-13}
\multirow[c]{14}{*}{$s_M$} & Bas. & Logits & 0.61 & \color[HTML]{A0A1A3} ±0.08 & 0.91 & \color[HTML]{A0A1A3} ±0.07 & 0.52 & \color[HTML]{A0A1A3} ±0.24 & 0.55 & \color[HTML]{A0A1A3} ±0.26 & 0.61 & \color[HTML]{A0A1A3} ±0.28 \\
\cline{2-13}
 & \multirow[c]{5}{*}{$\operatorname{Agg}_{\emptyset}$} & Median & 0.72 & \color[HTML]{A0A1A3} ±0.15 & 0.69 & \color[HTML]{A0A1A3} ±0.25 & 0.41 & \color[HTML]{A0A1A3} ±0.21 & 0.61 & \color[HTML]{A0A1A3} ±0.27 & 0.73 & \color[HTML]{A0A1A3} ±0.26 \\
 &  & \texttt{PW} & 0.69 & \color[HTML]{A0A1A3} ±0.12 & 0.83 & \color[HTML]{A0A1A3} ±0.18 & 0.43 & \color[HTML]{A0A1A3} ±0.24 & 0.70 & \color[HTML]{A0A1A3} ±0.24 & 0.61 & \color[HTML]{A0A1A3} ±0.29 \\
 &  & Mean & 0.62 & \color[HTML]{A0A1A3} ±0.10 & 0.80 & \color[HTML]{A0A1A3} ±0.16 & 0.43 & \color[HTML]{A0A1A3} ±0.23 & 0.55 & \color[HTML]{A0A1A3} ±0.28 & 0.63 & \color[HTML]{A0A1A3} ±0.26 \\
 &  & Max & 0.60 & \color[HTML]{A0A1A3} ±0.08 & 0.84 & \color[HTML]{A0A1A3} ±0.15 & 0.45 & \color[HTML]{A0A1A3} ±0.25 & 0.56 & \color[HTML]{A0A1A3} ±0.29 & 0.60 & \color[HTML]{A0A1A3} ±0.26 \\
 &  & Min & 0.50 & \color[HTML]{A0A1A3} ±0.00 & 1.00 & \color[HTML]{A0A1A3} ±0.00 & 0.50 & \color[HTML]{A0A1A3} ±0.28 & 0.75 & \color[HTML]{A0A1A3} ±0.14 & 0.75 & \color[HTML]{A0A1A3} ±0.14 \\
\cline{2-13}
 & \multirow[c]{8}{*}{$\operatorname{Agg}$} & \texttt{IF} & \bfseries 0.88 & \color[HTML]{A0A1A3} ±0.14 & \bfseries 0.35 & \color[HTML]{A0A1A3} ±0.31 & \bfseries 0.18 & \color[HTML]{A0A1A3} ±0.19 & \bfseries 0.85 & \color[HTML]{A0A1A3} ±0.19 & \bfseries 0.84 & \color[HTML]{A0A1A3} ±0.24 \\
 &  & $s_M$ & 0.78 & \color[HTML]{A0A1A3} ±0.16 & 0.59 & \color[HTML]{A0A1A3} ±0.34 & 0.31 & \color[HTML]{A0A1A3} ±0.25 & 0.75 & \color[HTML]{A0A1A3} ±0.24 & 0.73 & \color[HTML]{A0A1A3} ±0.29 \\
 &  & $s_{C}$ & 0.74 & \color[HTML]{A0A1A3} ±0.19 & 0.64 & \color[HTML]{A0A1A3} ±0.41 & 0.33 & \color[HTML]{A0A1A3} ±0.20 & 0.79 & \color[HTML]{A0A1A3} ±0.15 & 0.79 & \color[HTML]{A0A1A3} ±0.15 \\
 &  & \texttt{LOF} & 0.73 & \color[HTML]{A0A1A3} ±0.14 & 0.74 & \color[HTML]{A0A1A3} ±0.23 & 0.39 & \color[HTML]{A0A1A3} ±0.24 & 0.67 & \color[HTML]{A0A1A3} ±0.29 & 0.66 & \color[HTML]{A0A1A3} ±0.27 \\
 &  & $s_C*$ & 0.59 & \color[HTML]{A0A1A3} ±0.12 & 0.89 & \color[HTML]{A0A1A3} ±0.15 & 0.45 & \color[HTML]{A0A1A3} ±0.27 & 0.59 & \color[HTML]{A0A1A3} ±0.31 & 0.59 & \color[HTML]{A0A1A3} ±0.24 \\
 &  & $s_M*$ & 0.57 & \color[HTML]{A0A1A3} ±0.05 & 0.89 & \color[HTML]{A0A1A3} ±0.06 & 0.37 & \color[HTML]{A0A1A3} ±0.22 & 0.66 & \color[HTML]{A0A1A3} ±0.25 & 0.44 & \color[HTML]{A0A1A3} ±0.25 \\
 &  & \texttt{IF}* & 0.56 & \color[HTML]{A0A1A3} ±0.05 & 0.89 & \color[HTML]{A0A1A3} ±0.05 & 0.47 & \color[HTML]{A0A1A3} ±0.23 & 0.52 & \color[HTML]{A0A1A3} ±0.28 & 0.56 & \color[HTML]{A0A1A3} ±0.28 \\
 &  & \texttt{LoF}* & 0.50 & \color[HTML]{A0A1A3} ±0.01 & 1.00 & \color[HTML]{A0A1A3} ±0.01 & 0.41 & \color[HTML]{A0A1A3} ±0.27 & 0.70 & \color[HTML]{A0A1A3} ±0.20 & 0.61 & \color[HTML]{A0A1A3} ±0.21 \\
\cline{1-13} \cline{2-13}
\multirow[c]{14}{*}{$s_{C}$} & Bas. & Logits & 0.64 & \color[HTML]{A0A1A3} ±0.12 & 0.96 & \color[HTML]{A0A1A3} ±0.10 & 0.50 & \color[HTML]{A0A1A3} ±0.26 & 0.62 & \color[HTML]{A0A1A3} ±0.26 & 0.61 & \color[HTML]{A0A1A3} ±0.27 \\
\cline{2-13}
 & \multirow[c]{5}{*}{$\operatorname{Agg}_{\emptyset}$} & Median & 0.92 & \color[HTML]{A0A1A3} ±0.13 & 0.26 & \color[HTML]{A0A1A3} ±0.32 & 0.17 & \color[HTML]{A0A1A3} ±0.23 & 0.90 & \color[HTML]{A0A1A3} ±0.21 & 0.91 & \color[HTML]{A0A1A3} ±0.16 \\
 &  & \texttt{PW} & 0.89 & \color[HTML]{A0A1A3} ±0.14 & 0.29 & \color[HTML]{A0A1A3} ±0.30 & 0.17 & \color[HTML]{A0A1A3} ±0.20 & 0.85 & \color[HTML]{A0A1A3} ±0.20 & 0.88 & \color[HTML]{A0A1A3} ±0.20 \\
 &  & Mean & 0.89 & \color[HTML]{A0A1A3} ±0.15 & 0.36 & \color[HTML]{A0A1A3} ±0.35 & 0.21 & \color[HTML]{A0A1A3} ±0.23 & 0.89 & \color[HTML]{A0A1A3} ±0.20 & 0.85 & \color[HTML]{A0A1A3} ±0.21 \\
 &  & Min & 0.84 & \color[HTML]{A0A1A3} ±0.15 & 0.56 & \color[HTML]{A0A1A3} ±0.33 & 0.29 & \color[HTML]{A0A1A3} ±0.24 & 0.85 & \color[HTML]{A0A1A3} ±0.19 & 0.76 & \color[HTML]{A0A1A3} ±0.24 \\
 &  & Max & 0.61 & \color[HTML]{A0A1A3} ±0.16 & 0.92 & \color[HTML]{A0A1A3} ±0.23 & 0.48 & \color[HTML]{A0A1A3} ±0.29 & 0.70 & \color[HTML]{A0A1A3} ±0.26 & 0.74 & \color[HTML]{A0A1A3} ±0.19 \\
\cline{2-13}
 & \multirow[c]{8}{*}{$\operatorname{Agg}$} & $s_{C}$ & \bfseries 1.00 & \color[HTML]{A0A1A3} ±0.00 & \bfseries 0.00 & \color[HTML]{A0A1A3} ±0.00 & \bfseries 0.02 & \color[HTML]{A0A1A3} ±0.01 & \bfseries 1.00 & \color[HTML]{A0A1A3} ±0.00 & \bfseries 1.00 & \color[HTML]{A0A1A3} ±0.00 \\
 &  & \texttt{IF} & 0.92 & \color[HTML]{A0A1A3} ±0.13 & 0.13 & \color[HTML]{A0A1A3} ±0.22 & 0.07 & \color[HTML]{A0A1A3} ±0.14 & 0.86 & \color[HTML]{A0A1A3} ±0.25 & 0.85 & \color[HTML]{A0A1A3} ±0.19 \\
 &  & \texttt{LOF} & 0.87 & \color[HTML]{A0A1A3} ±0.14 & 0.33 & \color[HTML]{A0A1A3} ±0.30 & 0.19 & \color[HTML]{A0A1A3} ±0.20 & 0.82 & \color[HTML]{A0A1A3} ±0.21 & 0.87 & \color[HTML]{A0A1A3} ±0.21 \\
 &  & $s_M$ & 0.87 & \color[HTML]{A0A1A3} ±0.15 & 0.33 & \color[HTML]{A0A1A3} ±0.31 & 0.20 & \color[HTML]{A0A1A3} ±0.22 & 0.81 & \color[HTML]{A0A1A3} ±0.23 & 0.87 & \color[HTML]{A0A1A3} ±0.20 \\
 &  & \texttt{IF}* & 0.61 & \color[HTML]{A0A1A3} ±0.06 & 0.81 & \color[HTML]{A0A1A3} ±0.11 & 0.47 & \color[HTML]{A0A1A3} ±0.21 & 0.49 & \color[HTML]{A0A1A3} ±0.27 & 0.65 & \color[HTML]{A0A1A3} ±0.25 \\
 &  & $s_M*$ & 0.59 & \color[HTML]{A0A1A3} ±0.08 & 0.86 & \color[HTML]{A0A1A3} ±0.12 & 0.45 & \color[HTML]{A0A1A3} ±0.23 & 0.54 & \color[HTML]{A0A1A3} ±0.28 & 0.57 & \color[HTML]{A0A1A3} ±0.29 \\
 &  & $s_C*$ & 0.51 & \color[HTML]{A0A1A3} ±0.01 & 0.99 & \color[HTML]{A0A1A3} ±0.00 & 0.59 & \color[HTML]{A0A1A3} ±0.26 & 0.42 & \color[HTML]{A0A1A3} ±0.27 & 0.60 & \color[HTML]{A0A1A3} ±0.26 \\
 &  & \texttt{LoF}* & 0.50 & \color[HTML]{A0A1A3} ±0.00 & 1.00 & \color[HTML]{A0A1A3} ±0.00 & 0.50 & \color[HTML]{A0A1A3} ±0.28 & 0.75 & \color[HTML]{A0A1A3} ±0.14 & 0.75 & \color[HTML]{A0A1A3} ±0.14 \\
\cline{1-13} \cline{2-13}
\multirow[c]{14}{*}{$s_{\text{IRW}}$} & Bas. & Logits & 0.74 & \color[HTML]{A0A1A3} ±0.10 & 0.59 & \color[HTML]{A0A1A3} ±0.21 & 0.31 & \color[HTML]{A0A1A3} ±0.19 & 0.64 & \color[HTML]{A0A1A3} ±0.25 & 0.74 & \color[HTML]{A0A1A3} ±0.21 \\
\cline{2-13}
 & \multirow[c]{5}{*}{$\operatorname{Agg}_{\emptyset}$} & Median & 0.83 & \color[HTML]{A0A1A3} ±0.14 & 0.59 & \color[HTML]{A0A1A3} ±0.35 & 0.34 & \color[HTML]{A0A1A3} ±0.26 & 0.80 & \color[HTML]{A0A1A3} ±0.24 & 0.77 & \color[HTML]{A0A1A3} ±0.24 \\
 &  & Mean & 0.82 & \color[HTML]{A0A1A3} ±0.15 & 0.60 & \color[HTML]{A0A1A3} ±0.35 & 0.37 & \color[HTML]{A0A1A3} ±0.28 & 0.76 & \color[HTML]{A0A1A3} ±0.27 & 0.79 & \color[HTML]{A0A1A3} ±0.22 \\
 &  & Max & 0.80 & \color[HTML]{A0A1A3} ±0.15 & 0.63 & \color[HTML]{A0A1A3} ±0.30 & 0.38 & \color[HTML]{A0A1A3} ±0.25 & 0.73 & \color[HTML]{A0A1A3} ±0.27 & 0.77 & \color[HTML]{A0A1A3} ±0.22 \\
 &  & \texttt{PW} & 0.66 & \color[HTML]{A0A1A3} ±0.09 & 0.82 & \color[HTML]{A0A1A3} ±0.18 & 0.42 & \color[HTML]{A0A1A3} ±0.24 & 0.64 & \color[HTML]{A0A1A3} ±0.28 & 0.60 & \color[HTML]{A0A1A3} ±0.24 \\
 &  & Min & 0.50 & \color[HTML]{A0A1A3} ±0.00 & 1.00 & \color[HTML]{A0A1A3} ±0.00 & 0.50 & \color[HTML]{A0A1A3} ±0.28 & 0.75 & \color[HTML]{A0A1A3} ±0.14 & 0.75 & \color[HTML]{A0A1A3} ±0.14 \\
\cline{2-13}
 & \multirow[c]{8}{*}{$\operatorname{Agg}$} & $s_{C}$ & \bfseries 0.90 & \color[HTML]{A0A1A3} ±0.13 & \bfseries 0.27 & \color[HTML]{A0A1A3} ±0.31 & \bfseries 0.16 & \color[HTML]{A0A1A3} ±0.15 & \bfseries 0.89 & \color[HTML]{A0A1A3} ±0.15 & \bfseries 0.91 & \color[HTML]{A0A1A3} ±0.12 \\
 &  & \texttt{IF} & 0.85 & \color[HTML]{A0A1A3} ±0.15 & 0.44 & \color[HTML]{A0A1A3} ±0.36 & 0.22 & \color[HTML]{A0A1A3} ±0.21 & 0.83 & \color[HTML]{A0A1A3} ±0.24 & 0.79 & \color[HTML]{A0A1A3} ±0.27 \\
 &  & $s_M$ & 0.77 & \color[HTML]{A0A1A3} ±0.17 & 0.60 & \color[HTML]{A0A1A3} ±0.34 & 0.28 & \color[HTML]{A0A1A3} ±0.23 & 0.78 & \color[HTML]{A0A1A3} ±0.25 & 0.69 & \color[HTML]{A0A1A3} ±0.30 \\
 &  & \texttt{LOF} & 0.73 & \color[HTML]{A0A1A3} ±0.14 & 0.76 & \color[HTML]{A0A1A3} ±0.27 & 0.40 & \color[HTML]{A0A1A3} ±0.25 & 0.70 & \color[HTML]{A0A1A3} ±0.26 & 0.66 & \color[HTML]{A0A1A3} ±0.29 \\
 &  & \texttt{IF}* & 0.65 & \color[HTML]{A0A1A3} ±0.08 & 0.69 & \color[HTML]{A0A1A3} ±0.16 & 0.38 & \color[HTML]{A0A1A3} ±0.20 & 0.54 & \color[HTML]{A0A1A3} ±0.28 & 0.69 & \color[HTML]{A0A1A3} ±0.22 \\
 &  & $s_M*$ & 0.60 & \color[HTML]{A0A1A3} ±0.07 & 0.85 & \color[HTML]{A0A1A3} ±0.12 & 0.28 & \color[HTML]{A0A1A3} ±0.15 & 0.74 & \color[HTML]{A0A1A3} ±0.19 & 0.40 & \color[HTML]{A0A1A3} ±0.21 \\
 &  & $s_C*$ & 0.52 & \color[HTML]{A0A1A3} ±0.01 & 0.96 & \color[HTML]{A0A1A3} ±0.01 & 0.55 & \color[HTML]{A0A1A3} ±0.26 & 0.46 & \color[HTML]{A0A1A3} ±0.28 & 0.57 & \color[HTML]{A0A1A3} ±0.28 \\
 &  & \texttt{LoF}* & 0.50 & \color[HTML]{A0A1A3} ±0.00 & 1.00 & \color[HTML]{A0A1A3} ±0.00 & 0.53 & \color[HTML]{A0A1A3} ±0.28 & 0.73 & \color[HTML]{A0A1A3} ±0.14 & 0.64 & \color[HTML]{A0A1A3} ±0.26 \\
\cline{1-13} \cline{2-13}
MSP & Bas. & MSP & \bfseries 0.69 & \color[HTML]{A0A1A3} ±0.16 & \bfseries 0.62 & \color[HTML]{A0A1A3} ±0.25 & \bfseries 0.31 & \color[HTML]{A0A1A3} ±0.19 & \bfseries 0.63 & \color[HTML]{A0A1A3} ±0.25 & \bfseries 0.69 & \color[HTML]{A0A1A3} ±0.27 \\
\cline{1-13} \cline{2-13}
\bottomrule
\end{tabular}
} \caption{Average performance on the french benchamrk.} \label{tab:average_perf_french} \end{table}

\begin{table}[!ht] \centering \resizebox{0.4\textwidth}{!}{\begin{tabular}{lllrrrrrrrrrr}
\toprule
 &  &  & \multicolumn{2}{c}{\AUROC} & \multicolumn{2}{c}{\FPR} & \multicolumn{2}{c}{\ERR} & \multicolumn{2}{c}{\AUPRIN} & \multicolumn{2}{c}{\AUPROUT} \\
Metric & Ours & Agg. &  &  &  &  &  &  &  &  &  &  \\
\midrule
$E$ & Bas. & $E$ & \bfseries 0.69 & \color[HTML]{A0A1A3} ±0.15 & \bfseries 0.67 & \color[HTML]{A0A1A3} ±0.29 & \bfseries 0.35 & \color[HTML]{A0A1A3} ±0.18 & \bfseries 0.67 & \color[HTML]{A0A1A3} ±0.16 & \bfseries 0.70 & \color[HTML]{A0A1A3} ±0.17 \\
\cline{1-13} \cline{2-13}
\multirow[c]{14}{*}{$s_M$} & Bas. & Logits & 0.62 & \color[HTML]{A0A1A3} ±0.08 & 0.85 & \color[HTML]{A0A1A3} ±0.07 & 0.48 & \color[HTML]{A0A1A3} ±0.12 & 0.55 & \color[HTML]{A0A1A3} ±0.16 & 0.65 & \color[HTML]{A0A1A3} ±0.12 \\
\cline{2-13}
 & \multirow[c]{5}{*}{$\operatorname{Agg}_{\emptyset}$} & Median & 0.71 & \color[HTML]{A0A1A3} ±0.15 & 0.78 & \color[HTML]{A0A1A3} ±0.22 & 0.43 & \color[HTML]{A0A1A3} ±0.17 & 0.69 & \color[HTML]{A0A1A3} ±0.20 & 0.69 & \color[HTML]{A0A1A3} ±0.20 \\
 &  & \texttt{PW} & 0.70 & \color[HTML]{A0A1A3} ±0.13 & 0.83 & \color[HTML]{A0A1A3} ±0.15 & 0.46 & \color[HTML]{A0A1A3} ±0.16 & 0.69 & \color[HTML]{A0A1A3} ±0.19 & 0.69 & \color[HTML]{A0A1A3} ±0.13 \\
 &  & Mean & 0.58 & \color[HTML]{A0A1A3} ±0.07 & 0.90 & \color[HTML]{A0A1A3} ±0.07 & 0.47 & \color[HTML]{A0A1A3} ±0.11 & 0.56 & \color[HTML]{A0A1A3} ±0.16 & 0.57 & \color[HTML]{A0A1A3} ±0.15 \\
 &  & Max & 0.57 & \color[HTML]{A0A1A3} ±0.06 & 0.91 & \color[HTML]{A0A1A3} ±0.07 & 0.48 & \color[HTML]{A0A1A3} ±0.12 & 0.55 & \color[HTML]{A0A1A3} ±0.17 & 0.56 & \color[HTML]{A0A1A3} ±0.14 \\
 &  & Min & 0.50 & \color[HTML]{A0A1A3} ±0.00 & 1.00 & \color[HTML]{A0A1A3} ±0.00 & 0.50 & \color[HTML]{A0A1A3} ±0.14 & 0.75 & \color[HTML]{A0A1A3} ±0.07 & 0.75 & \color[HTML]{A0A1A3} ±0.07 \\
\cline{2-13}
 & \multirow[c]{8}{*}{$\operatorname{Agg}$} & \texttt{IF} & \bfseries 0.85 & \color[HTML]{A0A1A3} ±0.13 & \bfseries 0.45 & \color[HTML]{A0A1A3} ±0.27 & \bfseries 0.24 & \color[HTML]{A0A1A3} ±0.13 & \bfseries 0.83 & \color[HTML]{A0A1A3} ±0.15 & \bfseries 0.83 & \color[HTML]{A0A1A3} ±0.18 \\
 &  & \texttt{LOF} & 0.74 & \color[HTML]{A0A1A3} ±0.07 & 0.80 & \color[HTML]{A0A1A3} ±0.11 & 0.42 & \color[HTML]{A0A1A3} ±0.11 & 0.72 & \color[HTML]{A0A1A3} ±0.13 & 0.70 & \color[HTML]{A0A1A3} ±0.13 \\
 &  & $s_M$ & 0.65 & \color[HTML]{A0A1A3} ±0.09 & 0.83 & \color[HTML]{A0A1A3} ±0.10 & 0.45 & \color[HTML]{A0A1A3} ±0.14 & 0.62 & \color[HTML]{A0A1A3} ±0.18 & 0.65 & \color[HTML]{A0A1A3} ±0.09 \\
 &  & $s_{C}$ & 0.61 & \color[HTML]{A0A1A3} ±0.10 & 0.91 & \color[HTML]{A0A1A3} ±0.12 & 0.46 & \color[HTML]{A0A1A3} ±0.05 & 0.64 & \color[HTML]{A0A1A3} ±0.10 & 0.65 & \color[HTML]{A0A1A3} ±0.10 \\
 &  & \texttt{IF}* & 0.56 & \color[HTML]{A0A1A3} ±0.04 & 0.89 & \color[HTML]{A0A1A3} ±0.05 & 0.42 & \color[HTML]{A0A1A3} ±0.11 & 0.60 & \color[HTML]{A0A1A3} ±0.13 & 0.51 & \color[HTML]{A0A1A3} ±0.11 \\
 &  & $s_C*$ & 0.56 & \color[HTML]{A0A1A3} ±0.08 & 0.94 & \color[HTML]{A0A1A3} ±0.09 & 0.45 & \color[HTML]{A0A1A3} ±0.12 & 0.63 & \color[HTML]{A0A1A3} ±0.14 & 0.54 & \color[HTML]{A0A1A3} ±0.17 \\
 &  & $s_M*$ & 0.53 & \color[HTML]{A0A1A3} ±0.02 & 0.93 & \color[HTML]{A0A1A3} ±0.02 & 0.49 & \color[HTML]{A0A1A3} ±0.13 & 0.52 & \color[HTML]{A0A1A3} ±0.14 & 0.53 & \color[HTML]{A0A1A3} ±0.14 \\
 &  & \texttt{LoF}* & 0.50 & \color[HTML]{A0A1A3} ±0.00 & 1.00 & \color[HTML]{A0A1A3} ±0.00 & 0.47 & \color[HTML]{A0A1A3} ±0.14 & 0.68 & \color[HTML]{A0A1A3} ±0.17 & 0.65 & \color[HTML]{A0A1A3} ±0.14 \\
\cline{1-13} \cline{2-13}
\multirow[c]{14}{*}{$s_{C}$} & Bas. & Logits & 0.66 & \color[HTML]{A0A1A3} ±0.10 & 0.87 & \color[HTML]{A0A1A3} ±0.09 & 0.46 & \color[HTML]{A0A1A3} ±0.12 & 0.64 & \color[HTML]{A0A1A3} ±0.13 & 0.64 & \color[HTML]{A0A1A3} ±0.17 \\
\cline{2-13}
 & \multirow[c]{5}{*}{$\operatorname{Agg}_{\emptyset}$} & \texttt{PW} & 0.85 & \color[HTML]{A0A1A3} ±0.13 & 0.43 & \color[HTML]{A0A1A3} ±0.32 & 0.23 & \color[HTML]{A0A1A3} ±0.17 & 0.84 & \color[HTML]{A0A1A3} ±0.16 & 0.84 & \color[HTML]{A0A1A3} ±0.17 \\
 &  & Median & 0.85 & \color[HTML]{A0A1A3} ±0.13 & 0.48 & \color[HTML]{A0A1A3} ±0.37 & 0.28 & \color[HTML]{A0A1A3} ±0.21 & 0.84 & \color[HTML]{A0A1A3} ±0.16 & 0.84 & \color[HTML]{A0A1A3} ±0.16 \\
 &  & Mean & 0.85 & \color[HTML]{A0A1A3} ±0.14 & 0.49 & \color[HTML]{A0A1A3} ±0.36 & 0.28 & \color[HTML]{A0A1A3} ±0.21 & 0.84 & \color[HTML]{A0A1A3} ±0.16 & 0.83 & \color[HTML]{A0A1A3} ±0.18 \\
 &  & Min & 0.82 & \color[HTML]{A0A1A3} ±0.14 & 0.58 & \color[HTML]{A0A1A3} ±0.30 & 0.33 & \color[HTML]{A0A1A3} ±0.18 & 0.82 & \color[HTML]{A0A1A3} ±0.16 & 0.80 & \color[HTML]{A0A1A3} ±0.18 \\
 &  & Max & 0.52 & \color[HTML]{A0A1A3} ±0.04 & 1.00 & \color[HTML]{A0A1A3} ±0.00 & 0.49 & \color[HTML]{A0A1A3} ±0.14 & 0.68 & \color[HTML]{A0A1A3} ±0.17 & 0.74 & \color[HTML]{A0A1A3} ±0.07 \\
\cline{2-13}
 & \multirow[c]{8}{*}{$\operatorname{Agg}$} & $s_{C}$ & \bfseries 1.00 & \color[HTML]{A0A1A3} ±0.00 & \bfseries 0.00 & \color[HTML]{A0A1A3} ±0.00 & \bfseries 0.01 & \color[HTML]{A0A1A3} ±0.01 & \bfseries 1.00 & \color[HTML]{A0A1A3} ±0.00 & \bfseries 1.00 & \color[HTML]{A0A1A3} ±0.00 \\
 &  & \texttt{IF} & 0.96 & \color[HTML]{A0A1A3} ±0.11 & 0.16 & \color[HTML]{A0A1A3} ±0.38 & 0.09 & \color[HTML]{A0A1A3} ±0.21 & 0.95 & \color[HTML]{A0A1A3} ±0.18 & 0.92 & \color[HTML]{A0A1A3} ±0.16 \\
 &  & $s_M$ & 0.84 & \color[HTML]{A0A1A3} ±0.14 & 0.45 & \color[HTML]{A0A1A3} ±0.32 & 0.24 & \color[HTML]{A0A1A3} ±0.18 & 0.82 & \color[HTML]{A0A1A3} ±0.18 & 0.83 & \color[HTML]{A0A1A3} ±0.16 \\
 &  & \texttt{LOF} & 0.83 & \color[HTML]{A0A1A3} ±0.14 & 0.45 & \color[HTML]{A0A1A3} ±0.31 & 0.24 & \color[HTML]{A0A1A3} ±0.17 & 0.81 & \color[HTML]{A0A1A3} ±0.18 & 0.83 & \color[HTML]{A0A1A3} ±0.16 \\
 &  & \texttt{IF}* & 0.58 & \color[HTML]{A0A1A3} ±0.07 & 0.84 & \color[HTML]{A0A1A3} ±0.10 & 0.44 & \color[HTML]{A0A1A3} ±0.14 & 0.55 & \color[HTML]{A0A1A3} ±0.15 & 0.59 & \color[HTML]{A0A1A3} ±0.15 \\
 &  & $s_M*$ & 0.55 & \color[HTML]{A0A1A3} ±0.02 & 0.91 & \color[HTML]{A0A1A3} ±0.03 & 0.48 & \color[HTML]{A0A1A3} ±0.12 & 0.52 & \color[HTML]{A0A1A3} ±0.14 & 0.56 & \color[HTML]{A0A1A3} ±0.14 \\
 &  & $s_C*$ & 0.51 & \color[HTML]{A0A1A3} ±0.01 & 0.99 & \color[HTML]{A0A1A3} ±0.00 & 0.47 & \color[HTML]{A0A1A3} ±0.13 & 0.54 & \color[HTML]{A0A1A3} ±0.15 & 0.47 & \color[HTML]{A0A1A3} ±0.14 \\
 &  & \texttt{LoF}* & 0.50 & \color[HTML]{A0A1A3} ±0.00 & 1.00 & \color[HTML]{A0A1A3} ±0.00 & 0.50 & \color[HTML]{A0A1A3} ±0.14 & 0.75 & \color[HTML]{A0A1A3} ±0.07 & 0.75 & \color[HTML]{A0A1A3} ±0.07 \\
\cline{1-13} \cline{2-13}
\multirow[c]{14}{*}{$s_{\text{IRW}}$} & Bas. & Logits & 0.69 & \color[HTML]{A0A1A3} ±0.08 & 0.63 & \color[HTML]{A0A1A3} ±0.17 & 0.35 & \color[HTML]{A0A1A3} ±0.13 & 0.59 & \color[HTML]{A0A1A3} ±0.15 & 0.75 & \color[HTML]{A0A1A3} ±0.14 \\
\cline{2-13}
 & \multirow[c]{5}{*}{$\operatorname{Agg}_{\emptyset}$} & Mean & 0.79 & \color[HTML]{A0A1A3} ±0.16 & 0.67 & \color[HTML]{A0A1A3} ±0.32 & 0.37 & \color[HTML]{A0A1A3} ±0.21 & 0.80 & \color[HTML]{A0A1A3} ±0.18 & 0.75 & \color[HTML]{A0A1A3} ±0.17 \\
 &  & Median & 0.78 & \color[HTML]{A0A1A3} ±0.17 & 0.64 & \color[HTML]{A0A1A3} ±0.41 & 0.37 & \color[HTML]{A0A1A3} ±0.26 & 0.77 & \color[HTML]{A0A1A3} ±0.22 & 0.77 & \color[HTML]{A0A1A3} ±0.16 \\
 &  & Max & 0.76 & \color[HTML]{A0A1A3} ±0.15 & 0.74 & \color[HTML]{A0A1A3} ±0.25 & 0.41 & \color[HTML]{A0A1A3} ±0.18 & 0.77 & \color[HTML]{A0A1A3} ±0.18 & 0.74 & \color[HTML]{A0A1A3} ±0.16 \\
 &  & \texttt{PW} & 0.68 & \color[HTML]{A0A1A3} ±0.13 & 0.87 & \color[HTML]{A0A1A3} ±0.11 & 0.45 & \color[HTML]{A0A1A3} ±0.14 & 0.68 & \color[HTML]{A0A1A3} ±0.21 & 0.63 & \color[HTML]{A0A1A3} ±0.13 \\
 &  & Min & 0.50 & \color[HTML]{A0A1A3} ±0.00 & 1.00 & \color[HTML]{A0A1A3} ±0.00 & 0.50 & \color[HTML]{A0A1A3} ±0.14 & 0.75 & \color[HTML]{A0A1A3} ±0.07 & 0.75 & \color[HTML]{A0A1A3} ±0.07 \\
\cline{2-13}
 & \multirow[c]{8}{*}{$\operatorname{Agg}$} & $s_{C}$ & \bfseries 0.83 & \color[HTML]{A0A1A3} ±0.14 & \bfseries 0.50 & \color[HTML]{A0A1A3} ±0.33 & \bfseries 0.28 & \color[HTML]{A0A1A3} ±0.17 & \bfseries 0.80 & \color[HTML]{A0A1A3} ±0.15 & \bfseries 0.83 & \color[HTML]{A0A1A3} ±0.14 \\
 &  & \texttt{IF} & 0.80 & \color[HTML]{A0A1A3} ±0.15 & 0.53 & \color[HTML]{A0A1A3} ±0.31 & 0.29 & \color[HTML]{A0A1A3} ±0.17 & 0.77 & \color[HTML]{A0A1A3} ±0.16 & 0.79 & \color[HTML]{A0A1A3} ±0.19 \\
 &  & $s_M$ & 0.69 & \color[HTML]{A0A1A3} ±0.16 & 0.73 & \color[HTML]{A0A1A3} ±0.28 & 0.35 & \color[HTML]{A0A1A3} ±0.12 & 0.69 & \color[HTML]{A0A1A3} ±0.11 & 0.65 & \color[HTML]{A0A1A3} ±0.24 \\
 &  & \texttt{IF}* & 0.63 & \color[HTML]{A0A1A3} ±0.08 & 0.74 & \color[HTML]{A0A1A3} ±0.15 & 0.41 & \color[HTML]{A0A1A3} ±0.13 & 0.55 & \color[HTML]{A0A1A3} ±0.15 & 0.69 & \color[HTML]{A0A1A3} ±0.14 \\
 &  & \texttt{LOF} & 0.61 & \color[HTML]{A0A1A3} ±0.09 & 0.91 & \color[HTML]{A0A1A3} ±0.04 & 0.47 & \color[HTML]{A0A1A3} ±0.12 & 0.63 & \color[HTML]{A0A1A3} ±0.12 & 0.57 & \color[HTML]{A0A1A3} ±0.15 \\
 &  & $s_M*$ & 0.57 & \color[HTML]{A0A1A3} ±0.05 & 0.89 & \color[HTML]{A0A1A3} ±0.08 & 0.37 & \color[HTML]{A0A1A3} ±0.07 & 0.66 & \color[HTML]{A0A1A3} ±0.09 & 0.46 & \color[HTML]{A0A1A3} ±0.10 \\
 &  & $s_C*$ & 0.51 & \color[HTML]{A0A1A3} ±0.01 & 0.96 & \color[HTML]{A0A1A3} ±0.02 & 0.53 & \color[HTML]{A0A1A3} ±0.13 & 0.48 & \color[HTML]{A0A1A3} ±0.13 & 0.55 & \color[HTML]{A0A1A3} ±0.14 \\
 &  & \texttt{LoF}* & 0.50 & \color[HTML]{A0A1A3} ±0.00 & 0.99 & \color[HTML]{A0A1A3} ±0.00 & 0.46 & \color[HTML]{A0A1A3} ±0.13 & 0.77 & \color[HTML]{A0A1A3} ±0.07 & 0.62 & \color[HTML]{A0A1A3} ±0.18 \\
\cline{1-13} \cline{2-13}
MSP & Bas. & MSP & \bfseries 0.68 & \color[HTML]{A0A1A3} ±0.15 & \bfseries 0.69 & \color[HTML]{A0A1A3} ±0.28 & \bfseries 0.36 & \color[HTML]{A0A1A3} ±0.17 & \bfseries 0.65 & \color[HTML]{A0A1A3} ±0.16 & \bfseries 0.69 & \color[HTML]{A0A1A3} ±0.17 \\
\cline{1-13} \cline{2-13}
\bottomrule
\end{tabular}
} \caption{Average performance on the spanish benchamrk.} \label{tab:average_perf_spanish} \end{table}

\begin{table}[!ht] \centering \resizebox{0.4\textwidth}{!}{\begin{tabular}{lllrrrrrrrrrr}
\toprule
 &  &  & \multicolumn{2}{c}{\AUROC} & \multicolumn{2}{c}{\FPR} & \multicolumn{2}{c}{\ERR} & \multicolumn{2}{c}{\AUPRIN} & \multicolumn{2}{c}{\AUPROUT} \\
Metric & Ours & Agg. &  &  &  &  &  &  &  &  &  &  \\
\midrule
$E$ & Bas. & $E$ & \bfseries 0.55 & \color[HTML]{A0A1A3} ±0.29 & \bfseries 0.67 & \color[HTML]{A0A1A3} ±0.28 & \bfseries 0.36 & \color[HTML]{A0A1A3} ±0.27 & \bfseries 0.54 & \color[HTML]{A0A1A3} ±0.34 & \bfseries 0.65 & \color[HTML]{A0A1A3} ±0.33 \\
\cline{1-13} \cline{2-13}
\multirow[c]{14}{*}{$s_M$} & Bas. & Logits & 0.66 & \color[HTML]{A0A1A3} ±0.11 & 0.82 & \color[HTML]{A0A1A3} ±0.20 & 0.54 & \color[HTML]{A0A1A3} ±0.31 & 0.51 & \color[HTML]{A0A1A3} ±0.35 & 0.71 & \color[HTML]{A0A1A3} ±0.29 \\
\cline{2-13}
 & \multirow[c]{5}{*}{$\operatorname{Agg}_{\emptyset}$} & Median & 0.78 & \color[HTML]{A0A1A3} ±0.15 & 0.58 & \color[HTML]{A0A1A3} ±0.32 & 0.38 & \color[HTML]{A0A1A3} ±0.27 & 0.64 & \color[HTML]{A0A1A3} ±0.27 & \bfseries 0.82 & \color[HTML]{A0A1A3} ±0.27 \\
 &  & \texttt{PW} & 0.76 & \color[HTML]{A0A1A3} ±0.13 & 0.72 & \color[HTML]{A0A1A3} ±0.16 & 0.43 & \color[HTML]{A0A1A3} ±0.26 & 0.72 & \color[HTML]{A0A1A3} ±0.24 & 0.73 & \color[HTML]{A0A1A3} ±0.28 \\
 &  & Max & 0.65 & \color[HTML]{A0A1A3} ±0.10 & 0.82 & \color[HTML]{A0A1A3} ±0.17 & 0.24 & \color[HTML]{A0A1A3} ±0.18 & 0.83 & \color[HTML]{A0A1A3} ±0.16 & 0.39 & \color[HTML]{A0A1A3} ±0.26 \\
 &  & Mean & 0.63 & \color[HTML]{A0A1A3} ±0.10 & 0.83 & \color[HTML]{A0A1A3} ±0.14 & 0.38 & \color[HTML]{A0A1A3} ±0.27 & 0.66 & \color[HTML]{A0A1A3} ±0.31 & 0.53 & \color[HTML]{A0A1A3} ±0.30 \\
 &  & Min & 0.50 & \color[HTML]{A0A1A3} ±0.00 & 1.00 & \color[HTML]{A0A1A3} ±0.00 & 0.50 & \color[HTML]{A0A1A3} ±0.34 & 0.75 & \color[HTML]{A0A1A3} ±0.17 & 0.75 & \color[HTML]{A0A1A3} ±0.17 \\
\cline{2-13}
 & \multirow[c]{8}{*}{$\operatorname{Agg}$} & \texttt{IF} & \bfseries 0.90 & \color[HTML]{A0A1A3} ±0.15 & \bfseries 0.24 & \color[HTML]{A0A1A3} ±0.33 & \bfseries 0.07 & \color[HTML]{A0A1A3} ±0.06 & \bfseries 0.96 & \color[HTML]{A0A1A3} ±0.04 & 0.81 & \color[HTML]{A0A1A3} ±0.32 \\
 &  & $s_M$ & 0.74 & \color[HTML]{A0A1A3} ±0.15 & 0.62 & \color[HTML]{A0A1A3} ±0.30 & 0.35 & \color[HTML]{A0A1A3} ±0.28 & 0.71 & \color[HTML]{A0A1A3} ±0.29 & 0.73 & \color[HTML]{A0A1A3} ±0.30 \\
 &  & \texttt{LOF} & 0.74 & \color[HTML]{A0A1A3} ±0.11 & 0.70 & \color[HTML]{A0A1A3} ±0.27 & 0.41 & \color[HTML]{A0A1A3} ±0.28 & 0.63 & \color[HTML]{A0A1A3} ±0.34 & 0.69 & \color[HTML]{A0A1A3} ±0.28 \\
 &  & $s_{C}$ & 0.70 & \color[HTML]{A0A1A3} ±0.15 & 0.75 & \color[HTML]{A0A1A3} ±0.32 & 0.39 & \color[HTML]{A0A1A3} ±0.15 & 0.73 & \color[HTML]{A0A1A3} ±0.13 & 0.73 & \color[HTML]{A0A1A3} ±0.15 \\
 &  & $s_C*$ & 0.63 & \color[HTML]{A0A1A3} ±0.11 & 0.96 & \color[HTML]{A0A1A3} ±0.09 & 0.25 & \color[HTML]{A0A1A3} ±0.21 & 0.84 & \color[HTML]{A0A1A3} ±0.15 & 0.40 & \color[HTML]{A0A1A3} ±0.19 \\
 &  & \texttt{IF}* & 0.57 & \color[HTML]{A0A1A3} ±0.04 & 0.87 & \color[HTML]{A0A1A3} ±0.05 & 0.53 & \color[HTML]{A0A1A3} ±0.28 & 0.44 & \color[HTML]{A0A1A3} ±0.33 & 0.64 & \color[HTML]{A0A1A3} ±0.30 \\
 &  & $s_M*$ & 0.57 & \color[HTML]{A0A1A3} ±0.06 & 0.90 & \color[HTML]{A0A1A3} ±0.06 & 0.26 & \color[HTML]{A0A1A3} ±0.18 & 0.78 & \color[HTML]{A0A1A3} ±0.20 & 0.30 & \color[HTML]{A0A1A3} ±0.22 \\
 &  & \texttt{LoF}* & 0.50 & \color[HTML]{A0A1A3} ±0.00 & 1.00 & \color[HTML]{A0A1A3} ±0.01 & 0.31 & \color[HTML]{A0A1A3} ±0.27 & 0.76 & \color[HTML]{A0A1A3} ±0.20 & 0.50 & \color[HTML]{A0A1A3} ±0.26 \\
\cline{1-13} \cline{2-13}
\multirow[c]{14}{*}{$s_{C}$} & Bas. & Logits & 0.61 & \color[HTML]{A0A1A3} ±0.08 & 0.99 & \color[HTML]{A0A1A3} ±0.02 & 0.58 & \color[HTML]{A0A1A3} ±0.32 & 0.51 & \color[HTML]{A0A1A3} ±0.31 & 0.65 & \color[HTML]{A0A1A3} ±0.31 \\
\cline{2-13}
 & \multirow[c]{5}{*}{$\operatorname{Agg}_{\emptyset}$} & Median & 0.93 & \color[HTML]{A0A1A3} ±0.12 & 0.23 & \color[HTML]{A0A1A3} ±0.35 & 0.22 & \color[HTML]{A0A1A3} ±0.32 & 0.89 & \color[HTML]{A0A1A3} ±0.22 & 0.97 & \color[HTML]{A0A1A3} ±0.02 \\
 &  & Mean & 0.93 & \color[HTML]{A0A1A3} ±0.11 & 0.28 & \color[HTML]{A0A1A3} ±0.40 & 0.26 & \color[HTML]{A0A1A3} ±0.35 & 0.90 & \color[HTML]{A0A1A3} ±0.18 & 0.97 & \color[HTML]{A0A1A3} ±0.03 \\
 &  & \texttt{PW} & 0.91 & \color[HTML]{A0A1A3} ±0.14 & 0.18 & \color[HTML]{A0A1A3} ±0.24 & 0.06 & \color[HTML]{A0A1A3} ±0.04 & 0.96 & \color[HTML]{A0A1A3} ±0.03 & 0.86 & \color[HTML]{A0A1A3} ±0.23 \\
 &  & Min & 0.90 & \color[HTML]{A0A1A3} ±0.12 & 0.39 & \color[HTML]{A0A1A3} ±0.37 & 0.32 & \color[HTML]{A0A1A3} ±0.34 & 0.86 & \color[HTML]{A0A1A3} ±0.21 & 0.93 & \color[HTML]{A0A1A3} ±0.06 \\
 &  & Max & 0.56 & \color[HTML]{A0A1A3} ±0.07 & 0.98 & \color[HTML]{A0A1A3} ±0.04 & 0.38 & \color[HTML]{A0A1A3} ±0.31 & 0.69 & \color[HTML]{A0A1A3} ±0.31 & 0.60 & \color[HTML]{A0A1A3} ±0.25 \\
\cline{2-13}
 & \multirow[c]{8}{*}{$\operatorname{Agg}$} & $s_{C}$ & \bfseries 1.00 & \color[HTML]{A0A1A3} ±0.00 & \bfseries 0.00 & \color[HTML]{A0A1A3} ±0.00 & \bfseries 0.02 & \color[HTML]{A0A1A3} ±0.01 & \bfseries 1.00 & \color[HTML]{A0A1A3} ±0.00 & \bfseries 1.00 & \color[HTML]{A0A1A3} ±0.00 \\
 &  & \texttt{IF} & 0.93 & \color[HTML]{A0A1A3} ±0.13 & 0.11 & \color[HTML]{A0A1A3} ±0.18 & 0.02 & \color[HTML]{A0A1A3} ±0.03 & 0.95 & \color[HTML]{A0A1A3} ±0.07 & 0.84 & \color[HTML]{A0A1A3} ±0.16 \\
 &  & $s_M$ & 0.92 & \color[HTML]{A0A1A3} ±0.12 & 0.17 & \color[HTML]{A0A1A3} ±0.20 & 0.06 & \color[HTML]{A0A1A3} ±0.04 & 0.96 & \color[HTML]{A0A1A3} ±0.03 & 0.88 & \color[HTML]{A0A1A3} ±0.19 \\
 &  & \texttt{LOF} & 0.91 & \color[HTML]{A0A1A3} ±0.12 & 0.19 & \color[HTML]{A0A1A3} ±0.19 & 0.06 & \color[HTML]{A0A1A3} ±0.04 & 0.96 & \color[HTML]{A0A1A3} ±0.03 & 0.87 & \color[HTML]{A0A1A3} ±0.19 \\
 &  & \texttt{IF}* & 0.61 & \color[HTML]{A0A1A3} ±0.06 & 0.78 & \color[HTML]{A0A1A3} ±0.10 & 0.51 & \color[HTML]{A0A1A3} ±0.23 & 0.40 & \color[HTML]{A0A1A3} ±0.31 & 0.73 & \color[HTML]{A0A1A3} ±0.27 \\
 &  & $s_M*$ & 0.56 & \color[HTML]{A0A1A3} ±0.03 & 0.90 & \color[HTML]{A0A1A3} ±0.03 & 0.43 & \color[HTML]{A0A1A3} ±0.29 & 0.58 & \color[HTML]{A0A1A3} ±0.34 & 0.49 & \color[HTML]{A0A1A3} ±0.33 \\
 &  & $s_C*$ & 0.51 & \color[HTML]{A0A1A3} ±0.00 & 0.99 & \color[HTML]{A0A1A3} ±0.00 & 0.75 & \color[HTML]{A0A1A3} ±0.21 & 0.24 & \color[HTML]{A0A1A3} ±0.21 & 0.76 & \color[HTML]{A0A1A3} ±0.20 \\
 &  & \texttt{LoF}* & 0.50 & \color[HTML]{A0A1A3} ±0.00 & 1.00 & \color[HTML]{A0A1A3} ±0.00 & 0.50 & \color[HTML]{A0A1A3} ±0.34 & 0.75 & \color[HTML]{A0A1A3} ±0.17 & 0.75 & \color[HTML]{A0A1A3} ±0.17 \\
\cline{1-13} \cline{2-13}
\multirow[c]{14}{*}{$s_{\text{IRW}}$} & Bas. & Logits & 0.69 & \color[HTML]{A0A1A3} ±0.09 & 0.75 & \color[HTML]{A0A1A3} ±0.21 & 0.48 & \color[HTML]{A0A1A3} ±0.29 & 0.54 & \color[HTML]{A0A1A3} ±0.30 & 0.75 & \color[HTML]{A0A1A3} ±0.25 \\
\cline{2-13}
 & \multirow[c]{5}{*}{$\operatorname{Agg}_{\emptyset}$} & Mean & 0.92 & \color[HTML]{A0A1A3} ±0.09 & 0.35 & \color[HTML]{A0A1A3} ±0.39 & 0.24 & \color[HTML]{A0A1A3} ±0.32 & 0.89 & \color[HTML]{A0A1A3} ±0.19 & 0.87 & \color[HTML]{A0A1A3} ±0.15 \\
 &  & Median & 0.91 & \color[HTML]{A0A1A3} ±0.13 & 0.40 & \color[HTML]{A0A1A3} ±0.43 & 0.24 & \color[HTML]{A0A1A3} ±0.31 & 0.88 & \color[HTML]{A0A1A3} ±0.23 & 0.83 & \color[HTML]{A0A1A3} ±0.21 \\
 &  & Max & 0.88 & \color[HTML]{A0A1A3} ±0.13 & 0.40 & \color[HTML]{A0A1A3} ±0.33 & 0.28 & \color[HTML]{A0A1A3} ±0.31 & 0.84 & \color[HTML]{A0A1A3} ±0.24 & 0.84 & \color[HTML]{A0A1A3} ±0.19 \\
 &  & \texttt{PW} & 0.79 & \color[HTML]{A0A1A3} ±0.14 & 0.86 & \color[HTML]{A0A1A3} ±0.25 & 0.47 & \color[HTML]{A0A1A3} ±0.35 & 0.77 & \color[HTML]{A0A1A3} ±0.25 & 0.65 & \color[HTML]{A0A1A3} ±0.26 \\
 &  & Min & 0.50 & \color[HTML]{A0A1A3} ±0.00 & 1.00 & \color[HTML]{A0A1A3} ±0.00 & 0.50 & \color[HTML]{A0A1A3} ±0.34 & 0.75 & \color[HTML]{A0A1A3} ±0.17 & 0.75 & \color[HTML]{A0A1A3} ±0.17 \\
\cline{2-13}
 & \multirow[c]{8}{*}{$\operatorname{Agg}$} & $s_{C}$ & \bfseries 0.96 & \color[HTML]{A0A1A3} ±0.07 & \bfseries 0.13 & \color[HTML]{A0A1A3} ±0.19 & \bfseries 0.09 & \color[HTML]{A0A1A3} ±0.09 & \bfseries 0.94 & \color[HTML]{A0A1A3} ±0.10 & \bfseries 0.97 & \color[HTML]{A0A1A3} ±0.06 \\
 &  & \texttt{IF} & 0.90 & \color[HTML]{A0A1A3} ±0.18 & 0.23 & \color[HTML]{A0A1A3} ±0.34 & 0.16 & \color[HTML]{A0A1A3} ±0.24 & 0.85 & \color[HTML]{A0A1A3} ±0.26 & 0.91 & \color[HTML]{A0A1A3} ±0.19 \\
 &  & $s_M$ & 0.87 & \color[HTML]{A0A1A3} ±0.17 & 0.34 & \color[HTML]{A0A1A3} ±0.40 & 0.25 & \color[HTML]{A0A1A3} ±0.31 & 0.77 & \color[HTML]{A0A1A3} ±0.30 & 0.90 & \color[HTML]{A0A1A3} ±0.20 \\
 &  & \texttt{LOF} & 0.84 & \color[HTML]{A0A1A3} ±0.17 & 0.47 & \color[HTML]{A0A1A3} ±0.40 & 0.33 & \color[HTML]{A0A1A3} ±0.33 & 0.75 & \color[HTML]{A0A1A3} ±0.29 & 0.86 & \color[HTML]{A0A1A3} ±0.23 \\
 &  & \texttt{IF}* & 0.67 & \color[HTML]{A0A1A3} ±0.09 & 0.66 & \color[HTML]{A0A1A3} ±0.19 & 0.41 & \color[HTML]{A0A1A3} ±0.21 & 0.44 & \color[HTML]{A0A1A3} ±0.31 & 0.75 & \color[HTML]{A0A1A3} ±0.30 \\
 &  & $s_M*$ & 0.65 & \color[HTML]{A0A1A3} ±0.09 & 0.80 & \color[HTML]{A0A1A3} ±0.14 & 0.21 & \color[HTML]{A0A1A3} ±0.12 & 0.83 & \color[HTML]{A0A1A3} ±0.15 & 0.36 & \color[HTML]{A0A1A3} ±0.20 \\
 &  & $s_C*$ & 0.51 & \color[HTML]{A0A1A3} ±0.01 & 0.96 & \color[HTML]{A0A1A3} ±0.01 & 0.65 & \color[HTML]{A0A1A3} ±0.28 & 0.34 & \color[HTML]{A0A1A3} ±0.29 & 0.68 & \color[HTML]{A0A1A3} ±0.28 \\
 &  & \texttt{LoF}* & 0.50 & \color[HTML]{A0A1A3} ±0.00 & 1.00 & \color[HTML]{A0A1A3} ±0.00 & 0.59 & \color[HTML]{A0A1A3} ±0.33 & 0.70 & \color[HTML]{A0A1A3} ±0.16 & 0.72 & \color[HTML]{A0A1A3} ±0.26 \\
\cline{1-13} \cline{2-13}
MSP & Bas. & MSP & \bfseries 0.56 & \color[HTML]{A0A1A3} ±0.28 & \bfseries 0.67 & \color[HTML]{A0A1A3} ±0.27 & \bfseries 0.36 & \color[HTML]{A0A1A3} ±0.27 & \bfseries 0.55 & \color[HTML]{A0A1A3} ±0.34 & \bfseries 0.65 & \color[HTML]{A0A1A3} ±0.33 \\
\cline{1-13} \cline{2-13}
\bottomrule
\end{tabular}
} \caption{Average performance on the german benchamrk.} \label{tab:average_perf_german} \end{table}

In addition to the global performance, we provide the average OOD detection performance per IN-DS in Tables~\ref{tab:perfs_per_in_english} \ref{tab:perfs_per_in_french} \ref{tab:perfs_per_in_german} and \ref{tab:perfs_per_in_spanish} and per OUT-DS in Tables~\ref{tab:perfs_per_out_english} \ref{tab:perfs_per_out_french} \ref{tab:perfs_per_out_german} and \ref{tab:perfs_per_out_spanish}.

\begin{table*}[!ht] \centering \resizebox{0.5\textwidth}{!}{\begin{tabular}{lllrrrrrrrrr}
\toprule
 &  & in & 20ng & emotion & go-emotions & imdb & massive & rte & sst2 & trec & trec-fine \\
Metric & Ours & Agg. &  &  &  &  &  &  &  &  &  \\
\midrule
\multirow[c]{7}{*}{$s_{C}$} & \multirow[c]{2}{*}{$\operatorname{Agg}$} & \texttt{IF} & 0.95 & 0.95 & \bfseries 0.95 & 0.91 & \bfseries 0.95 & 0.82 & 0.91 & \bfseries 0.95 & \bfseries 0.95 \\
 &  & \texttt{LOF} & 0.99 & 0.99 & 0.81 & 0.97 & 0.92 & 0.81 & 0.94 & 0.93 & 0.90 \\
\cline{2-12}
 & \multirow[c]{3}{*}{$\operatorname{Agg}_{\emptyset}$} & Mean & 0.99 & \bfseries 0.99 & 0.82 & 0.92 & 0.92 & 0.84 & 0.96 & 0.93 & 0.90 \\
 &  & Median & 0.89 & 0.98 & 0.81 & 0.96 & 0.92 & 0.91 & 0.90 & 0.91 & 0.88 \\
 &  & \texttt{PW} & 0.98 & 0.99 & 0.79 & 0.96 & 0.92 & 0.84 & \bfseries 0.96 & 0.93 & 0.90 \\
\cline{2-12}
 & \multirow[c]{2}{*}{Bas.} & Last layer & 0.96 & 0.96 & 0.83 & 0.95 & 0.89 & 0.83 & 0.87 & 0.93 & 0.88 \\
 &  & Logits & 0.97 & 0.96 & 0.69 & 0.78 & 0.88 & 0.57 & 0.67 & 0.91 & 0.86 \\
\cline{1-12} \cline{2-12}
\multirow[c]{7}{*}{$s_{\text{IRW}}$} & \multirow[c]{2}{*}{$\operatorname{Agg}$} & \texttt{IF} & 0.99 & 0.95 & 0.81 & 0.98 & 0.85 & 0.91 & 0.77 & 0.90 & 0.88 \\
 &  & \texttt{LOF} & 0.92 & 0.90 & 0.66 & 0.78 & 0.75 & 0.84 & 0.78 & 0.89 & 0.85 \\
\cline{2-12}
 & \multirow[c]{3}{*}{$\operatorname{Agg}_{\emptyset}$} & Mean & 0.98 & 0.91 & 0.84 & 0.99 & 0.80 & \bfseries 0.95 & 0.82 & 0.90 & 0.50 \\
 &  & Median & 0.95 & 0.89 & 0.82 & 0.99 & 0.78 & 0.91 & 0.85 & 0.86 & 0.50 \\
 &  & \texttt{PW} & 0.87 & 0.84 & 0.73 & 0.91 & 0.67 & 0.79 & 0.77 & 0.72 & 0.50 \\
\cline{2-12}
 & \multirow[c]{2}{*}{Bas.} & Last layer & 0.85 & 0.79 & 0.72 & 0.87 & 0.71 & 0.80 & 0.81 & 0.77 & 0.50 \\
 &  & Logits & 0.88 & 0.90 & 0.61 & 0.88 & 0.62 & 0.57 & 0.84 & 0.60 & 0.50 \\
\cline{1-12} \cline{2-12}
\multirow[c]{7}{*}{$s_M$} & \multirow[c]{2}{*}{$\operatorname{Agg}$} & \texttt{IF} & \bfseries 1.00 & 0.98 & 0.80 & \bfseries 1.00 & 0.92 & 0.91 & 0.94 & 0.90 & 0.89 \\
 &  & \texttt{LOF} & 0.94 & 0.92 & 0.73 & 0.85 & 0.87 & 0.85 & 0.84 & 0.86 & 0.92 \\
\cline{2-12}
 & \multirow[c]{3}{*}{$\operatorname{Agg}_{\emptyset}$} & Mean & 0.91 & 0.83 & 0.65 & 0.72 & 0.72 & 0.68 & 0.72 & 0.74 & 0.84 \\
 &  & Median & 0.79 & 0.73 & 0.66 & 0.71 & 0.71 & 0.76 & 0.68 & 0.72 & 0.86 \\
 &  & \texttt{PW} & 0.85 & 0.85 & 0.65 & 0.87 & 0.77 & 0.76 & 0.72 & 0.83 & 0.90 \\
\cline{2-12}
 & \multirow[c]{2}{*}{Bas.} & Last layer & 0.93 & 0.85 & 0.64 & 0.77 & 0.79 & 0.75 & 0.72 & 0.83 & 0.88 \\
 &  & Logits & 0.85 & 0.71 & 0.65 & 0.73 & 0.62 & 0.61 & 0.66 & 0.65 & 0.76 \\
\cline{1-12} \cline{2-12}
\bottomrule
\end{tabular}
} \caption{Performance ({\AUROC}) of different aggregation per english IN-DS . } \label{tab:perfs_per_in_english} \end{table*}

\begin{table*}[!ht] \centering \resizebox{0.5\textwidth}{!}{\begin{tabular}{lllrrrrrrrrrrrr}
\toprule
 &  &  & \multicolumn{12}{c}{\AUROC} \\
 &  & out & 20ng & amazon-reviews-multi & b77 & emotion & go-emotions & imdb & massive & rte & snli & sst2 & trec & trec-fine \\
Metric & Ours & Agg. &  &  &  &  &  &  &  &  &  &  &  &  \\
\midrule
\multirow[c]{7}{*}{$s_{C}$} & \multirow[c]{2}{*}{$\operatorname{Agg}$} & \texttt{IF} & 0.54 & 0.41 & \bfseries 0.52 & \bfseries 0.53 & \bfseries 0.53 & \bfseries 0.54 & \bfseries 0.53 & 0.53 & \bfseries 0.54 & \bfseries 0.54 & 0.52 & \bfseries 0.54 \\
 &  & \texttt{LOF} & 0.52 & 0.52 & 0.50 & 0.51 & 0.51 & 0.52 & 0.51 & 0.50 & 0.53 & 0.51 & 0.52 & 0.52 \\
\cline{2-15}
 & \multirow[c]{3}{*}{$\operatorname{Agg}_{\emptyset}$} & Mean & 0.52 & 0.53 & 0.51 & 0.51 & 0.52 & 0.53 & 0.51 & 0.50 & 0.53 & 0.51 & 0.52 & 0.52 \\
 &  & Median & 0.51 & 0.53 & 0.50 & 0.51 & 0.52 & 0.52 & 0.53 & 0.51 & 0.52 & 0.52 & 0.52 & 0.52 \\
 &  & \texttt{PW} & 0.52 & \bfseries 0.53 & 0.51 & 0.51 & 0.52 & 0.52 & 0.52 & 0.51 & 0.53 & 0.52 & 0.52 & 0.52 \\
\cline{2-15}
 & \multirow[c]{2}{*}{Bas.} & Last layer & 0.53 & 0.52 & 0.49 & 0.51 & 0.51 & 0.53 & 0.52 & 0.50 & 0.52 & 0.51 & 0.51 & 0.52 \\
 &  & Logits & 0.50 & 0.51 & 0.48 & 0.50 & 0.51 & 0.51 & 0.51 & 0.50 & 0.52 & 0.51 & 0.49 & 0.50 \\
\cline{1-15} \cline{2-15}
\multirow[c]{7}{*}{$s_{\text{IRW}}$} & \multirow[c]{2}{*}{$\operatorname{Agg}$} & \texttt{IF} & 0.51 & 0.50 & 0.49 & 0.52 & 0.49 & 0.51 & 0.51 & 0.51 & 0.52 & 0.49 & 0.50 & 0.51 \\
 &  & \texttt{LOF} & 0.53 & 0.49 & 0.47 & 0.48 & 0.45 & 0.52 & 0.50 & 0.50 & 0.48 & 0.46 & 0.48 & 0.49 \\
\cline{2-15}
 & \multirow[c]{3}{*}{$\operatorname{Agg}_{\emptyset}$} & Mean & \bfseries 0.54 & 0.52 & 0.49 & 0.50 & 0.49 & 0.54 & 0.50 & \bfseries 0.53 & 0.51 & 0.50 & 0.49 & 0.50 \\
 &  & Median & 0.53 & 0.51 & 0.48 & 0.49 & 0.48 & 0.52 & 0.50 & 0.52 & 0.50 & 0.49 & 0.50 & 0.50 \\
 &  & \texttt{PW} & 0.48 & 0.44 & 0.45 & 0.46 & 0.43 & 0.47 & 0.46 & 0.47 & 0.46 & 0.46 & 0.47 & 0.48 \\
\cline{2-15}
 & \multirow[c]{2}{*}{Bas.} & Last layer & 0.49 & 0.47 & 0.44 & 0.45 & 0.44 & 0.48 & 0.46 & 0.47 & 0.47 & 0.45 & 0.46 & 0.45 \\
 &  & Logits & 0.44 & 0.47 & 0.45 & 0.42 & 0.44 & 0.44 & 0.47 & 0.45 & 0.45 & 0.44 & 0.46 & 0.47 \\
\cline{1-15} \cline{2-15}
\multirow[c]{7}{*}{$s_M$} & \multirow[c]{2}{*}{$\operatorname{Agg}$} & \texttt{IF} & 0.51 & 0.50 & 0.51 & 0.51 & 0.50 & 0.52 & 0.51 & 0.50 & 0.53 & 0.51 & \bfseries 0.53 & 0.52 \\
 &  & \texttt{LOF} & 0.52 & 0.51 & 0.50 & 0.51 & 0.50 & 0.52 & 0.51 & 0.53 & 0.52 & 0.50 & 0.51 & 0.51 \\
\cline{2-15}
 & \multirow[c]{3}{*}{$\operatorname{Agg}_{\emptyset}$} & Mean & 0.52 & 0.45 & 0.46 & 0.47 & 0.41 & 0.52 & 0.45 & 0.47 & 0.50 & 0.42 & 0.46 & 0.42 \\
 &  & Median & 0.52 & 0.47 & 0.42 & 0.45 & 0.41 & 0.51 & 0.47 & 0.44 & 0.49 & 0.42 & 0.47 & 0.43 \\
 &  & \texttt{PW} & 0.51 & 0.49 & 0.47 & 0.48 & 0.47 & 0.51 & 0.50 & 0.50 & 0.52 & 0.46 & 0.49 & 0.47 \\
\cline{2-15}
 & \multirow[c]{2}{*}{Bas.} & Last layer & 0.52 & 0.50 & 0.48 & 0.49 & 0.48 & 0.52 & 0.49 & 0.51 & 0.51 & 0.48 & 0.48 & 0.47 \\
 &  & Logits & 0.44 & 0.44 & 0.43 & 0.44 & 0.43 & 0.43 & 0.45 & 0.42 & 0.43 & 0.43 & 0.43 & 0.44 \\
\cline{1-15} \cline{2-15}
\bottomrule
\end{tabular}
} \caption{Performance ({\AUROC}) of different aggregation methods per english OUT-DS . } \label{tab:perfs_per_out_english} \end{table*}

\begin{table*}[!ht] \centering \resizebox{0.5\textwidth}{!}{\begin{tabular}{lllrrrrrrr}
\toprule
 &  & in & fr-allocine & fr-cls & fr-pawsx & fr-swiss-judgement & fr-tweet-sentiment & fr-xnli & fr-xstance \\
Metric & Ours & Agg. &  &  &  &  &  &  &  \\
\midrule
\multirow[c]{7}{*}{$s_{C}$} & \multirow[c]{2}{*}{$\operatorname{Agg}$} & \texttt{IF} & 0.87 & \bfseries 1.00 & 0.84 & 0.95 & \bfseries 1.00 & \bfseries 1.00 & 0.81 \\
 &  & \texttt{LOF} & 0.91 & \bfseries 1.00 & 0.95 & 0.87 & 0.66 & 0.82 & 0.89 \\
\cline{2-10}
 & \multirow[c]{3}{*}{$\operatorname{Agg}_{\emptyset}$} & Mean & \bfseries 0.92 & \bfseries 1.00 & 0.98 & 0.90 & 0.67 & 0.82 & 0.93 \\
 &  & Median & 0.89 & \bfseries 1.00 & \bfseries 0.99 & 0.98 & 0.73 & 0.92 & \bfseries 0.95 \\
 &  & \texttt{PW} & 0.91 & \bfseries 1.00 & 0.98 & 0.88 & 0.69 & 0.85 & 0.93 \\
\cline{2-10}
 & \multirow[c]{2}{*}{Bas.} & Last layer & 0.89 & \bfseries 1.00 & 0.91 & 0.86 & 0.63 & 0.75 & 0.76 \\
 &  & Logits & 0.82 & 0.67 & 0.64 & 0.56 & 0.57 & 0.58 & 0.62 \\
\cline{1-10} \cline{2-10}
\multirow[c]{7}{*}{$s_{\text{IRW}}$} & \multirow[c]{2}{*}{$\operatorname{Agg}$} & \texttt{IF} & 0.73 & 0.74 & 0.95 & \bfseries 0.99 & 0.86 & 0.89 & 0.80 \\
 &  & \texttt{LOF} & 0.63 & 0.77 & 0.68 & 0.85 & 0.72 & 0.77 & 0.67 \\
\cline{2-10}
 & \multirow[c]{3}{*}{$\operatorname{Agg}_{\emptyset}$} & Mean & 0.71 & 0.76 & 0.88 & 0.93 & 0.79 & 0.83 & 0.83 \\
 &  & Median & 0.71 & 0.75 & 0.96 & 0.86 & 0.84 & 0.83 & 0.84 \\
 &  & \texttt{PW} & 0.67 & 0.59 & 0.65 & 0.74 & 0.66 & 0.71 & 0.60 \\
\cline{2-10}
 & \multirow[c]{2}{*}{Bas.} & Last layer & 0.81 & 0.68 & 0.75 & 0.58 & 0.74 & 0.74 & 0.63 \\
 &  & Logits & 0.85 & 0.73 & 0.78 & 0.66 & 0.74 & 0.76 & 0.66 \\
\cline{1-10} \cline{2-10}
\multirow[c]{7}{*}{$s_M$} & \multirow[c]{2}{*}{$\operatorname{Agg}$} & \texttt{IF} & 0.89 & 0.94 & 0.95 & 0.93 & 0.67 & 0.86 & 0.92 \\
 &  & \texttt{LOF} & 0.70 & 0.87 & 0.84 & 0.60 & 0.78 & 0.56 & 0.73 \\
\cline{2-10}
 & \multirow[c]{3}{*}{$\operatorname{Agg}_{\emptyset}$} & Mean & 0.63 & 0.58 & 0.57 & 0.65 & 0.71 & 0.62 & 0.54 \\
 &  & Median & 0.75 & 0.82 & 0.65 & 0.63 & 0.71 & 0.67 & 0.80 \\
 &  & \texttt{PW} & 0.69 & 0.69 & 0.66 & 0.62 & 0.76 & 0.68 & 0.69 \\
\cline{2-10}
 & \multirow[c]{2}{*}{Bas.} & Last layer & 0.85 & 0.90 & 0.90 & 0.82 & 0.63 & 0.72 & 0.82 \\
 &  & Logits & 0.64 & 0.59 & 0.59 & 0.58 & 0.64 & 0.62 & 0.62 \\
\cline{1-10} \cline{2-10}
\bottomrule
\end{tabular}
} \caption{Performance ({\AUROC}) of different aggregation per french IN-DS . } \label{tab:perfs_per_in_french} \end{table*}

\begin{table*}[!ht] \centering \resizebox{0.5\textwidth}{!}{\begin{tabular}{lllrrrrrrr}
\toprule
 &  &  & \multicolumn{7}{c}{\AUROC} \\
 &  & out & fr-allocine & fr-cls & fr-pawsx & fr-swiss-judgement & fr-tweet-sentiment & fr-xnli & fr-xstance \\
Metric & Ours & Agg. &  &  &  &  &  &  &  \\
\midrule
\multirow[c]{7}{*}{$s_{C}$} & \multirow[c]{2}{*}{$\operatorname{Agg}$} & \texttt{IF} & 0.51 & \bfseries 0.54 & 0.50 & 0.51 & \bfseries 0.54 & \bfseries 0.53 & 0.50 \\
 &  & \texttt{LOF} & 0.50 & 0.46 & 0.51 & 0.51 & 0.50 & 0.49 & 0.49 \\
\cline{2-10}
 & \multirow[c]{3}{*}{$\operatorname{Agg}_{\emptyset}$} & Mean & 0.52 & 0.48 & \bfseries 0.52 & 0.51 & 0.50 & 0.51 & 0.51 \\
 &  & Median & \bfseries 0.52 & 0.51 & 0.50 & 0.50 & 0.50 & 0.52 & \bfseries 0.52 \\
 &  & \texttt{PW} & 0.51 & 0.48 & 0.50 & 0.51 & 0.50 & 0.50 & 0.50 \\
\cline{2-10}
 & \multirow[c]{2}{*}{Bas.} & Last layer & 0.48 & 0.43 & 0.50 & 0.50 & 0.51 & 0.48 & 0.49 \\
 &  & Logits & 0.28 & 0.29 & 0.43 & 0.41 & 0.40 & 0.37 & 0.37 \\
\cline{1-10} \cline{2-10}
\multirow[c]{7}{*}{$s_{\text{IRW}}$} & \multirow[c]{2}{*}{$\operatorname{Agg}$} & \texttt{IF} & 0.51 & 0.50 & 0.48 & \bfseries 0.52 & 0.50 & 0.45 & 0.47 \\
 &  & \texttt{LOF} & 0.38 & 0.38 & 0.44 & 0.50 & 0.46 & 0.40 & 0.37 \\
\cline{2-10}
 & \multirow[c]{3}{*}{$\operatorname{Agg}_{\emptyset}$} & Mean & 0.46 & 0.46 & 0.48 & 0.51 & 0.50 & 0.42 & 0.41 \\
 &  & Median & 0.47 & 0.47 & 0.46 & 0.51 & 0.50 & 0.43 & 0.43 \\
 &  & \texttt{PW} & 0.35 & 0.35 & 0.41 & 0.37 & 0.41 & 0.37 & 0.37 \\
\cline{2-10}
 & \multirow[c]{2}{*}{Bas.} & Last layer & 0.40 & 0.40 & 0.42 & 0.41 & 0.43 & 0.40 & 0.39 \\
 &  & Logits & 0.39 & 0.39 & 0.45 & 0.43 & 0.45 & 0.42 & 0.40 \\
\cline{1-10} \cline{2-10}
\multirow[c]{7}{*}{$s_M$} & \multirow[c]{2}{*}{$\operatorname{Agg}$} & \texttt{IF} & 0.48 & 0.48 & 0.51 & 0.51 & 0.50 & 0.50 & 0.49 \\
 &  & \texttt{LOF} & 0.41 & 0.40 & 0.44 & 0.50 & 0.46 & 0.38 & 0.40 \\
\cline{2-10}
 & \multirow[c]{3}{*}{$\operatorname{Agg}_{\emptyset}$} & Mean & 0.35 & 0.36 & 0.37 & 0.42 & 0.34 & 0.31 & 0.34 \\
 &  & Median & 0.38 & 0.39 & 0.46 & 0.48 & 0.51 & 0.33 & 0.37 \\
 &  & \texttt{PW} & 0.35 & 0.36 & 0.45 & 0.46 & 0.44 & 0.34 & 0.35 \\
\cline{2-10}
 & \multirow[c]{2}{*}{Bas.} & Last layer & 0.43 & 0.42 & 0.48 & 0.50 & 0.48 & 0.46 & 0.47 \\
 &  & Logits & 0.34 & 0.34 & 0.37 & 0.38 & 0.33 & 0.29 & 0.33 \\
\cline{1-10} \cline{2-10}
\bottomrule
\end{tabular}
} \caption{Performance ({\AUROC}) of different aggregation methods per french OUT-DS . } \label{tab:perfs_per_out_french} \end{table*}

\begin{table*}[!ht] \centering \resizebox{0.5\textwidth}{!}{\begin{tabular}{lllrrrr}
\toprule
 &  & in & es-cine & es-pawsx & es-tweet-inde & es-tweet-sentiment \\
Metric & Ours & Agg. &  &  &  &  \\
\midrule
\multirow[c]{7}{*}{$s_{C}$} & \multirow[c]{2}{*}{$\operatorname{Agg}$} & \texttt{IF} & \bfseries 1.00 & 0.87 & \bfseries 0.98 & \bfseries 1.00 \\
 &  & \texttt{LOF} & 0.77 & 0.97 & 0.84 & 0.74 \\
\cline{2-7}
 & \multirow[c]{3}{*}{$\operatorname{Agg}-{\emptyset}$} & Mean & 0.79 & 0.99 & 0.85 & 0.74 \\
 &  & Median & 0.86 & 0.99 & 0.82 & 0.72 \\
 &  & \texttt{PW} & 0.84 & 0.99 & 0.85 & 0.73 \\
\cline{2-7}
 & \multirow[c]{2}{*}{Bas.} & Last layer & 0.63 & 0.97 & 0.75 & 0.69 \\
 &  & Logits & 0.60 & 0.74 & 0.70 & 0.58 \\
\cline{1-7} \cline{2-7}
\multirow[c]{7}{*}{$s_{\text{IRW}}$} & \multirow[c]{2}{*}{$\operatorname{Agg}$} & \texttt{IF} & 0.84 & 0.98 & 0.77 & 0.62 \\
 &  & \texttt{LOF} & 0.60 & 0.67 & 0.61 & 0.58 \\
\cline{2-7}
 & \multirow[c]{3}{*}{$\operatorname{Agg}_{\emptyset}$} & Mean & 0.89 & 0.96 & 0.67 & 0.64 \\
 &  & Median & 0.86 & \bfseries 0.99 & 0.66 & 0.62 \\
 &  & \texttt{PW} & 0.72 & 0.83 & 0.57 & 0.59 \\
\cline{2-7}
 & \multirow[c]{2}{*}{Bas.} & Last layer & 0.66 & 0.69 & 0.61 & 0.59 \\
 &  & Logits & 0.72 & 0.73 & 0.74 & 0.58 \\
\cline{1-7} \cline{2-7}
\multirow[c]{7}{*}{$s_M$} & \multirow[c]{2}{*}{$\operatorname{Agg}$} & \texttt{IF} & 0.92 & 0.96 & 0.83 & 0.69 \\
 &  & \texttt{LOF} & 0.77 & 0.80 & 0.69 & 0.72 \\
\cline{2-7}
 & \multirow[c]{3}{*}{$\operatorname{Agg}_{\emptyset}$} & Mean & 0.53 & 0.57 & 0.55 & 0.66 \\
 &  & Median & 0.90 & 0.63 & 0.73 & 0.56 \\
 &  & \texttt{PW} & 0.67 & 0.71 & 0.74 & 0.70 \\
\cline{2-7}
 & \multirow[c]{2}{*}{Bas.} & Last layer & 0.69 & 0.85 & 0.83 & 0.70 \\
 &  & Logits & 0.58 & 0.67 & 0.55 & 0.66 \\
\cline{1-7} \cline{2-7}
\bottomrule
\end{tabular}
} \caption{Performance ({\AUROC}) of different aggregation per spanish IN-DS . } \label{tab:perfs_per_in_spanish} \end{table*}

\begin{table*}[!ht] \centering \resizebox{0.5\textwidth}{!}{\begin{tabular}{lllrrrr}
\toprule
 &  &  & \multicolumn{4}{c}{\AUROC} \\
 &  & out & es-cine & es-pawsx & es-tweet-inde & es-tweet-sentiment \\
Metric & Ours & Agg. &  &  &  &  \\
\midrule
\multirow[c]{7}{*}{$s_{C}$} & \multirow[c]{2}{*}{$\operatorname{Agg}$} & \texttt{IF} & 0.50 & \bfseries 0.50 & 0.50 & \bfseries 0.53 \\
 &  & \texttt{LOF} & 0.50 & 0.49 & 0.48 & 0.48 \\
\cline{2-7}
 & \multirow[c]{3}{*}{$\operatorname{Agg}-{\emptyset}$} & Mean & 0.51 & 0.49 & 0.49 & 0.49 \\
 &  & Median & 0.50 & 0.49 & 0.49 & 0.49 \\
 &  & \texttt{PW} & \bfseries 0.51 & 0.49 & 0.49 & 0.49 \\
\cline{2-7}
 & \multirow[c]{2}{*}{Bas.} & Last layer & 0.49 & 0.43 & 0.48 & 0.47 \\
 &  & Logits & 0.38 & 0.38 & 0.38 & 0.39 \\
\cline{1-7} \cline{2-7}
\multirow[c]{7}{*}{$s_{\text{IRW}}$} & \multirow[c]{2}{*}{$\operatorname{Agg}$} & \texttt{IF} & 0.47 & 0.43 & 0.49 & 0.49 \\
 &  & \texttt{LOF} & 0.26 & 0.35 & 0.37 & 0.37 \\
\cline{2-7}
 & \multirow[c]{3}{*}{$\operatorname{Agg}_{\emptyset}$} & Mean & 0.45 & 0.43 & 0.51 & 0.51 \\
 &  & Median & 0.48 & 0.41 & \bfseries 0.51 & 0.50 \\
 &  & \texttt{PW} & 0.41 & 0.31 & 0.43 & 0.46 \\
\cline{2-7}
 & \multirow[c]{2}{*}{Bas.} & Last layer & 0.32 & 0.32 & 0.34 & 0.37 \\
 &  & Logits & 0.37 & 0.42 & 0.37 & 0.39 \\
\cline{1-7} \cline{2-7}
\multirow[c]{7}{*}{$s_M$} & \multirow[c]{2}{*}{$\operatorname{Agg}$} & \texttt{IF} & 0.47 & 0.50 & 0.49 & 0.50 \\
 &  & \texttt{LOF} & 0.36 & 0.41 & 0.41 & 0.42 \\
\cline{2-7}
 & \multirow[c]{3}{*}{$\operatorname{Agg}_{\emptyset}$} & Mean & 0.35 & 0.31 & 0.33 & 0.32 \\
 &  & Median & 0.36 & 0.48 & 0.45 & 0.44 \\
 &  & \texttt{PW} & 0.38 & 0.44 & 0.38 & 0.38 \\
\cline{2-7}
 & \multirow[c]{2}{*}{Bas.} & Last layer & 0.45 & 0.47 & 0.43 & 0.44 \\
 &  & Logits & 0.40 & 0.35 & 0.30 & 0.32 \\
\cline{1-7} \cline{2-7}
\bottomrule
\end{tabular}
} \caption{Performance ({\AUROC}) of different aggregation methods per spanish OUT-DS . } \label{tab:perfs_per_out_spanish} \end{table*}

\begin{table*}[!ht] \centering \resizebox{0.5\textwidth}{!}{\begin{tabular}{lllrrrr}
\toprule
 &  & in & de-pawsx & de-swiss-judgement & de-tweet-sentiment & de-xstance \\
Metric & Ours & Agg. &  &  &  &  \\
\midrule
\multirow[c]{7}{*}{$s_{C}$} & \multirow[c]{2}{*}{$\operatorname{Agg}$} & \texttt{IF} & 0.88 & 0.99 & \bfseries 0.99 & 0.85 \\
 &  & \texttt{LOF} & 0.94 & 0.99 & 0.74 & 0.99 \\
\cline{2-7}
 & \multirow[c]{3}{*}{$\operatorname{Agg}-{\emptyset}$} & Mean & 0.97 & 1.00 & 0.76 & \bfseries 0.99 \\
 &  & Median & \bfseries 0.99 & 0.99 & 0.76 & 0.99 \\
 &  & \texttt{PW} & 0.96 & 0.99 & 0.70 & 0.99 \\
\cline{2-7}
 & \multirow[c]{2}{*}{Bas.} & Last layer & 0.93 & 0.99 & 0.61 & 0.97 \\
 &  & Logits & 0.67 & 0.53 & 0.55 & 0.67 \\
\cline{1-7} \cline{2-7}
\multirow[c]{7}{*}{$s_{\text{IRW}}$} & \multirow[c]{2}{*}{$\operatorname{Agg}$} & \texttt{IF} & 0.95 & \bfseries 1.00 & 0.68 & 0.97 \\
 &  & \texttt{LOF} & 0.84 & 0.98 & 0.68 & 0.84 \\
\cline{2-7}
 & \multirow[c]{3}{*}{$\operatorname{Agg}_{\emptyset}$} & Mean & 0.93 & 0.99 & 0.85 & 0.93 \\
 &  & Median & 0.97 & 0.99 & 0.77 & 0.90 \\
 &  & \texttt{PW} & 0.75 & 0.94 & 0.63 & 0.83 \\
\cline{2-7}
 & \multirow[c]{2}{*}{Bas.} & Last layer & 0.61 & 0.91 & 0.80 & 0.58 \\
 &  & Logits & 0.71 & 0.71 & 0.66 & 0.68 \\
\cline{1-7} \cline{2-7}
\multirow[c]{7}{*}{$s_M$} & \multirow[c]{2}{*}{$\operatorname{Agg}$} & \texttt{IF} & 0.95 & 1.00 & 0.66 & 0.99 \\
 &  & \texttt{LOF} & 0.85 & 0.67 & 0.74 & 0.69 \\
\cline{2-7}
 & \multirow[c]{3}{*}{$\operatorname{Agg}_{\emptyset}$} & Mean & 0.63 & 0.74 & 0.60 & 0.55 \\
 &  & Median & 0.73 & 0.79 & 0.66 & 0.95 \\
 &  & \texttt{PW} & 0.70 & 0.87 & 0.68 & 0.79 \\
\cline{2-7}
 & \multirow[c]{2}{*}{Bas.} & Last layer & 0.74 & 0.99 & 0.61 & 0.93 \\
 &  & Logits & 0.68 & 0.73 & 0.61 & 0.61 \\
\cline{1-7} \cline{2-7}
\bottomrule
\end{tabular}
} \caption{Performance ({\AUROC}) of different aggregation per german IN-DS . } \label{tab:perfs_per_in_german} \end{table*}

\begin{table*}[!ht] \centering \resizebox{0.5\textwidth}{!}{\begin{tabular}{lllrrrr}
\toprule
 &  &  & \multicolumn{4}{c}{\AUROC} \\
 &  & out & de-pawsx & de-swiss-judgement & de-tweet-sentiment & de-xstance \\
Metric & Ours & Agg. &  &  &  &  \\
\midrule
\multirow[c]{7}{*}{$s_{C}$} & \multirow[c]{2}{*}{$\operatorname{Agg}$} & \texttt{IF} & \bfseries 0.54 & 0.51 & 0.50 & 0.50 \\
 &  & \texttt{LOF} & 0.52 & 0.52 & 0.50 & 0.49 \\
\cline{2-7}
 & \multirow[c]{3}{*}{$\operatorname{Agg}-{\emptyset}$} & Mean & 0.51 & 0.53 & 0.50 & 0.51 \\
 &  & Median & 0.51 & 0.53 & 0.50 & \bfseries 0.53 \\
 &  & \texttt{PW} & 0.52 & 0.54 & 0.50 & 0.51 \\
\cline{2-7}
 & \multirow[c]{2}{*}{Bas.} & Last layer & 0.53 & 0.51 & 0.50 & 0.52 \\
 &  & Logits & 0.33 & 0.33 & 0.37 & 0.31 \\
\cline{1-7} \cline{2-7}
\multirow[c]{7}{*}{$s_{\text{IRW}}$} & \multirow[c]{2}{*}{$\operatorname{Agg}$} & \texttt{IF} & 0.53 & 0.50 & 0.50 & 0.52 \\
 &  & \texttt{LOF} & 0.49 & 0.49 & \bfseries 0.50 & 0.47 \\
\cline{2-7}
 & \multirow[c]{3}{*}{$\operatorname{Agg}_{\emptyset}$} & Mean & 0.50 & 0.50 & 0.50 & 0.49 \\
 &  & Median & 0.51 & 0.51 & 0.50 & 0.52 \\
 &  & \texttt{PW} & 0.48 & 0.39 & 0.47 & 0.47 \\
\cline{2-7}
 & \multirow[c]{2}{*}{Bas.} & Last layer & 0.45 & 0.41 & 0.42 & 0.50 \\
 &  & Logits & 0.35 & 0.40 & 0.39 & 0.42 \\
\cline{1-7} \cline{2-7}
\multirow[c]{7}{*}{$s_M$} & \multirow[c]{2}{*}{$\operatorname{Agg}$} & \texttt{IF} & 0.52 & \bfseries 0.54 & 0.50 & 0.52 \\
 &  & \texttt{LOF} & 0.36 & 0.49 & 0.44 & 0.35 \\
\cline{2-7}
 & \multirow[c]{3}{*}{$\operatorname{Agg}_{\emptyset}$} & Mean & 0.34 & 0.38 & 0.40 & 0.34 \\
 &  & Median & 0.46 & 0.48 & 0.49 & 0.35 \\
 &  & \texttt{PW} & 0.44 & 0.37 & 0.45 & 0.46 \\
\cline{2-7}
 & \multirow[c]{2}{*}{Bas.} & Last layer & 0.52 & 0.46 & 0.50 & 0.48 \\
 &  & Logits & 0.34 & 0.36 & 0.37 & 0.46 \\
\cline{1-7} \cline{2-7}
\bottomrule
\end{tabular}
} \caption{Performance ({\AUROC}) of different aggregation methods per german OUT-DS . } \label{tab:perfs_per_out_german} \end{table*}

\subsection{Performance on datasets with a very large number of classes.}
\begin{table}
\centering

\resizebox{0.5\textwidth}{!}{
\begin{tabular}{lllrr}
\toprule
 &  &  & clink & hint3 \\
Metric & Ours & Agg &  &  \\
\midrule
\multirow[c]{10}{*}{$s_{C}$} & \multirow[c]{4}{*}{$\operatorname{Agg}$} & $s_{C}$ & \bfseries 1.00 & \bfseries 0.97 \\
 &  & \texttt{IF} & 1.00 & 0.95 \\
 &  & \texttt{LOF} & 0.89 & 0.77 \\
 &  & $s_M$ & 0.89 & 0.76 \\
\cline{2-5}
 & \multirow[c]{3}{*}{$\operatorname{Agg}_{\emptyset}$} & Mean & 0.90 & 0.78 \\
 &  & Median & 0.90 & 0.76 \\
 &  & \texttt{PW} & 0.91 & 0.78 \\
\cline{2-5}
 & \multirow[c]{2}{*}{Bas.} & Last layer & 0.88 & 0.77 \\
 &  & Logits & 0.87 & 0.71 \\
\cline{1-5} \cline{2-5}
\multirow[c]{10}{*}{$s_{\text{IRW}}$} & \multirow[c]{4}{*}{$\operatorname{Agg}$} & $s_{C}$ & 0.96 & 0.50 \\
 &  & \texttt{IF} & 0.75 & 0.77 \\
 &  & \texttt{LOF} & 0.53 & 0.51 \\
 &  & $s_M$ & 0.72 & 0.50 \\
\cline{2-5}
 & \multirow[c]{3}{*}{$\operatorname{Agg}_{\emptyset}$} & Mean & 0.78 & 0.50 \\
 &  & Median & 0.65 & 0.50 \\
 &  & \texttt{PW} & 0.60 & 0.50 \\
\cline{2-5}
 & \multirow[c]{2}{*}{Bas.} & Last layer & 0.59 & 0.50 \\
 &  & Logits & 0.60 & 0.50 \\
\cline{1-5} \cline{2-5}
\multirow[c]{10}{*}{$s_M$} & \multirow[c]{4}{*}{$\operatorname{Agg}$} & $s_{C}$ & 0.96 & 0.98 \\
 &  & \texttt{IF} & 0.90 & 0.68 \\
 &  & \texttt{LOF} & 0.79 & 0.69 \\
 &  & $s_M$ & 0.90 & 0.80 \\
\cline{2-5}
 & \multirow[c]{3}{*}{$\operatorname{Agg}_{\emptyset}$} & Mean & 0.66 & 0.67 \\
 &  & Median & 0.66 & 0.61 \\
 &  & \texttt{PW} & 0.76 & 0.68 \\
\cline{2-5}
 & \multirow[c]{2}{*}{Bas.} & Last layer & 0.77 & 0.72 \\
 &  & Logits & 0.79 & 0.70 \\
\cline{1-5} \cline{2-5}
\bottomrule
\end{tabular}}
\caption{Performance of our aggregation methods on Clink and Hint3 datasets.}
\label{fig:hint3clink}
\end{table}

\section{Negative results}
\label{sec:negative_results}

\subsection{Computer vision experiments}

We ran additional experiments in computer vision on ViT-16/B (224x224) finetuned on CIFAR. Consistently with priori work we found that the last layer was most often than not the most useful layer for OOD detection.

\begin{table}[!ht]
    \centering
    \begin{tabular}{|l|l|l|l|l|l|l|l|}
    \hline
        ~ & ~ & $s_C$, $s_C$ & $s_C$, $s_M$ & $s_C$, Last layer & $s_M$, $s_M$ & $s_M$, \texttt{IF} & $s_M$, Last layer \\ \hline
        cifar10 & TNR & 78.1 & 39.5 & 78.0 & 34.7 & 11.7 & 79.3 \\ \hline
        ~ & ROC & 94.5 & 84.7 & 95.8 & 82.9 & 80.4 & 96.0 \\ \hline
        svhn & TNR & 74.4 & 63.7 & 76.1 & 83.8 & 97.6 & 82.9 \\ \hline
        ~ & ROC & 92.3 & 93.2 & 94.6 & 97.1 & 99.3 & 96.0 \\ \hline
        isun & TNR & 61.0 & 83.9 & 62.5 & 98.0 & 98.8 & 64.3 \\ \hline
        ~ & ROC & 90.7 & 96.8 & 92.7 & 99.5 & 99.6 & 93.6 \\ \hline
        lsun\_c & TNR & 73.7 & 66.7 & 74.9 & 96.5 & 99.1 & 80.7 \\ \hline
        ~ & ROC & 94.1 & 93.7 & 95.5 & 99.1 & 99.5 & 96.4 \\ \hline
        lsun\_r & TNR & 64.3 & 88.8 & 65.0 & 99.2 & 99.4 & 70.0 \\ \hline
        ~ & ROC & 91.5 & 97.6 & 93.3 & 99.8 & 99.7 & 94.4 \\ \hline
        tiny\_imagenet\_c & TNR & 71.9 & 84.0 & 73.0 & 99.6 & 99.9 & 76.5 \\ \hline
        ~ & ROC & 93.1 & 96.8 & 94.5 & 99.8 & 99.9 & 95.4 \\ \hline
        tiny\_imagenet\_r & TNR & 68.5 & 87.9 & 70.1 & 97.8 & 98.7 & 72.5 \\ \hline
        ~ & ROC & 92.7 & 97.5 & 93.9 & 99.5 & 99.6 & 94.6 \\ \hline
        textures & TNR & 92.3 & 99.6 & 93.0 & 100.0 & 100.0 & 97.4 \\ \hline
        ~ & ROC & 98.4 & 99.8 & 98.4 & 100.0 & 100.0 & 99.2 \\ \hline
        places365 & TNR & 68.3 & 90.8 & 70.4 & 100.0 & 100.0 & 81.2 \\ \hline
        ~ & ROC & 92.3 & 98.0 & 94.5 & 100.0 & 100.0 & 96.6 \\ \hline
        english\_chars & TNR & 82.1 & 79.6 & 82.9 & 92.8 & 98.7 & 90.0 \\ \hline
        ~ & ROC & 94.1 & 95.9 & 95.6 & 98.5 & 99.6 & 97.5 \\ \hline
    \end{tabular}
\end{table}

\begin{figure}
    \centering
    \includegraphics[width=\textwidth]{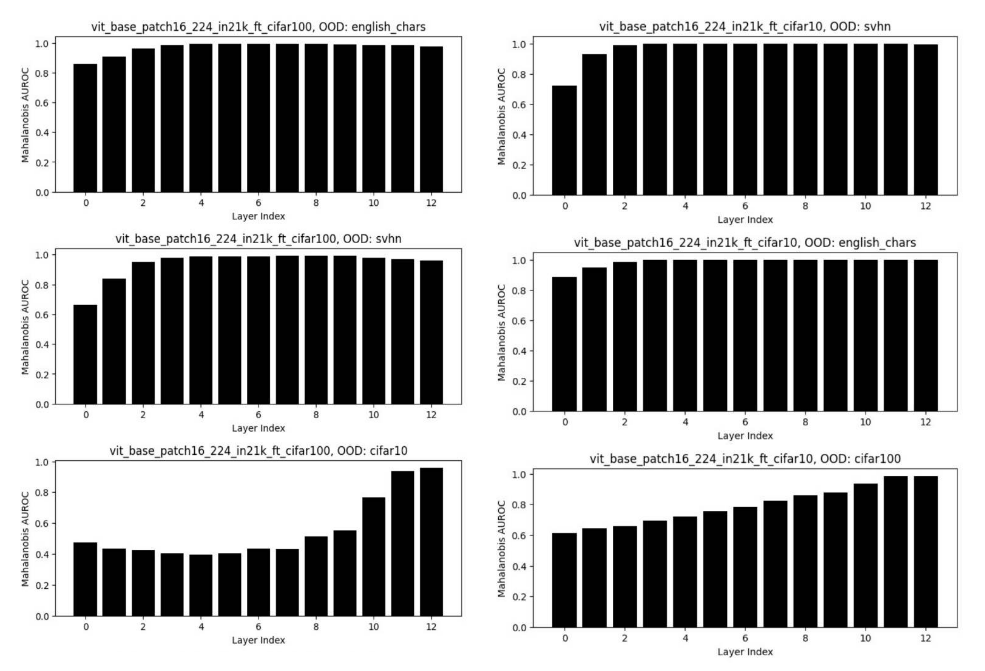}
    \caption{Performance in OOD detection for the different layers of the ViT models. We can observe that the last layer always yield very good OOD detection performance making aggregation methods (other than just selecting the last layer) somewhat irrelevant or at least not better than the usual heuristic.}
    \label{fig:cv_expe}
\end{figure}

\end{document}